\documentclass{article} 
\usepackage{arxiv}
\usepackage[utf8]{inputenc} 
\usepackage[T1]{fontenc}    
\usepackage{url}            
\usepackage{nicefrac}       
\usepackage{float}
\usepackage{fourier}

\usepackage{microtype}
\usepackage{hyperref}
\usepackage{url}
\usepackage{booktabs}

\usepackage{lineno}
\usepackage{enumitem}
\usepackage{latexsym}
\usepackage{booktabs}
\usepackage{graphicx}  
\usepackage{pifont}
\usepackage{subfigure}
\usepackage{multirow}
\usepackage{amsmath}
\usepackage{adjustbox}
\usepackage{array}
\usepackage[skins]{tcolorbox}
\usepackage{xcolor}
\usepackage{threeparttable}
\usepackage{colortbl} 
\usepackage{tabularx} 
\usepackage{xspace}
\usepackage{hyperref}
\usepackage{caption}
\usepackage{natbib}
\usepackage{amsmath}
\usepackage{amssymb}

\newcounter{finding}

\newcommand{\finding}[1]{%
  \refstepcounter{finding}%
  \begin{tcolorbox}[
        colback=white!90!gray,
        colframe=teal!60!black,
        arc=5pt,
        boxsep=5pt,
        left=1mm,
        right=1mm,
        top=1mm,
        bottom=1mm,
        boxrule=0.8pt,
        drop shadow=gray!50!white,
        enhanced jigsaw
    ]
    \vspace{-0.1cm}
        \paragraph{\textbf{\textit{Finding \thefinding:}}} #1
        \label{finding:\thefinding}
    \vspace{-0.1cm}
    \end{tcolorbox}
    \vspace{-0.1cm}
}

\newcommand\blfootnote[1]{%
  \begingroup
  \renewcommand\thefootnote{}\footnote{#1}%
  \addtocounter{footnote}{-1}%
  \endgroup
}

\newcolumntype{P}{>{\centering\arraybackslash}p{1.1cm}}
\definecolor{lightgray}{RGB}{230,230,230}
\definecolor{Periwinkle}{RGB}{204, 204, 255}
\definecolor{darkblue}{rgb}{0, 0, 0.5}
\hypersetup{colorlinks=true, citecolor=darkblue, linkcolor=darkblue, urlcolor=darkblue}

\newcommand{\task}{\textsc{TaskAnything}\xspace}
\newcommand{\judge}{\textsc{JudgeAnything}\xspace}
\newcommand{\arena}{\textsc{OmniArena}\xspace}
\newcommand{\method}{\textit{checklist-of-thought}\xspace}
\newcommand{\score}{\textit{Score Evaluation}\xspace}
\newcommand{\pair}{\textit{Pair Comparison}\xspace}
\newcommand{\overall}{\textit{Overall}\xspace}
\newcommand{\rubric}{\textit{Rubric}\xspace}
\newcommand{\checklist}{\textit{Checklist}\xspace}
\newcommand{\header}[1]{\smallskip \noindent\textbf{{#1}}\xspace}
\usepackage{wrapfig}

\newcommand{\na}{N/A}

\newcount\DraftStatus  
\definecolor{darkgreen}{rgb}{0,0.5,0} 
\definecolor{purple}{rgb}{1,0,1} 
\definecolor{todocolor}{rgb}{0.9,0.1,0.1} 
\definecolor{fixcolor}{rgb}{0.1,0.7,0.3} 
\definecolor{wycolor}{rgb}{0.9,0.1,0.1} 
\definecolor{hycolor}{rgb}{0.7,0.7,0.3} 

\newcommand{\nbc}[3]{\ifnum\DraftStatus=1
	{\colorbox{#3}{\bfseries\sffamily\scriptsize\textcolor{white}{#1}}}
	{\textcolor{#3}{\sf\small$\blacktriangleright$\emph{#2}$\blacktriangleleft$}}
\fi}

\newcommand{\draftnote}[2]{\ifnum\DraftStatus=1
	\marginpar{
		\tiny\raggedright
		\hbadness=10000
		\def\baselinestretch{0.8}
		\textcolor{#1}{\textsf{\hspace{0pt}#2}}}
\fi}

\title{Judge Anything: \\MLLM as a Judge Across Any Modality}

\renewcommand{\shorttitle}{Judge Anything: MLLM as a Judge Across Any Modality}

\author{Shu Pu\textsuperscript{1*}, Yaochen Wang\textsuperscript{1*}, Dongping Chen\textsuperscript{1$\ddagger$}, Yuhang Chen\textsuperscript{1\S}, Guohao Wang\textsuperscript{1\S}, Qi Qin\textsuperscript{1\S}, \\
\textbf{Zhongyi Zhang\textsuperscript{1\S}, Zhiyuan Zhang\textsuperscript{1\S}, Zetong Zhou\textsuperscript{1\S}, Shuang Gong\textsuperscript{1\S}, Yi Gui\textsuperscript{1},} \\ 
\textbf{Yao Wan\textsuperscript{1\textdagger}, Philip S. Yu\textsuperscript{2}}
\\~\\
\textsuperscript{1} Huazhong University of Science and Technology \\
\textsuperscript{2} University of Illinois Chicago
\\~\vspace{1em}\\}
\begin{document}


\maketitle

\blfootnote{* Co-first. \textsuperscript{$\ddagger$} Project Leader.}
\blfootnote{\textsuperscript{\textdagger} Correspondence to: Yao Wan~(\texttt{wanyao@hust.edu.cn}).}
\blfootnote{\textsuperscript{\S} Equal Contribution.}
\begin{abstract}
Evaluating generative foundation models on open-ended multimodal understanding (MMU) and generation (MMG) tasks across diverse modalities (\emph{e.g.}, images, audio, video) poses significant challenges due to the complexity of cross-modal interactions. 
To this end, the idea of utilizing Multimodal LLMs (MLLMs) as automated judges has emerged, with encouraging results in assessing vision-language understanding tasks.
Moving further, this paper extends MLLM-as-a-Judge across modalities to a unified manner by introducing two benchmarks, \task and \judge, to respectively evaluate the overall performance and judging capabilities of MLLMs across any-to-any modality tasks.
Specifically, \task evaluates the MMU and MMG capabilities across 15 any-to-any modality categories, employing 1,500 queries curated from well-established benchmarks.
Furthermore, \judge evaluates the judging capabilities of 5 advanced (\emph{e.g.}, GPT-4o and Gemini-2.0-Flash) from the perspectives of \pair and \score, providing a standardized testbed that incorporates human judgments and detailed rubrics.
Our extensive experiments reveal that while these MLLMs show promise in assessing MMU (\emph{i.e.}, achieving an average of 66.55\% in \pair setting and 42.79\% in \score setting), they encounter significant challenges with MMG tasks (\emph{i.e.}, averaging only 53.37\% in \pair setting and 30.05\% in \score setting), exposing cross-modality biases and hallucination issues.
To address this, we present \arena, an automated platform for evaluating omni-models and multimodal reward models. Our work highlights the need for fairer evaluation protocols and stronger alignment with human preferences.
The source code and dataset are publicly available at: \url{https://urrealhero.github.io/judgeanythingweb/}.
\end{abstract}

\section{Introduction}
The rapid advancement of generative models, particularly Large Language Models (LLMs)~\citep{2024gpt,liu2024deepseek} and diffusion-based visual generative models~\citep{rombach2022high,esser2024scaling}, has led to the widespread prevalence of AI-generated content (AIGC) across various modalities, including images~\citep{ghosh2023geneval}, video~\citep{yang2024cogvideox}, and audio~\citep{liu2024audioldm2}.
Recently, the omni-model is proposed to unify pre-training techniques across multiple modalities, aiming to integrate both multimodal understanding (MMU) and multimodal generation (MMG) capabilities~\citep{xie2024show,li2024omniflow,team2024chameleon}.


Despite this, evaluating the MMU and MMG capabilities of generative models typically relies on human judgment, given the inherently open-ended nature of related tasks. 
While human evaluations are commonly regarded as the gold standard \citep{huang2024vbench,jiang2025genai}, they tend to be time-consuming, expensive—particularly for high-dimensional modalities such as video and audio. Additionally, these evaluations are prone to inconsistency, as open-ended tasks often lack absolute ground truths or universally accepted evaluation criteria, further complicating reliable assessments.


To this end, researchers have explored automated evaluation methods, particularly by leveraging Multimodal LLMs (MLLMs) as assessment metrics - a concept referred to as MLLM-as-a-Judge~\citep{mllm-as-a-judge,LlavaCritic}. This approach introduces automated assessment of vision-and-language tasks, offering both qualitative insights and quantitative scores.
While inconsistency, biases, and hallucination remain, MLLM-as-a-Judge has demonstrated utility and promising results across a range of generative tasks—including text-to-image~\citep{chen2024mj}, text-to-video~\citep{luo2024videoautoarena}, and interleaved multimodal generation~\citep{chen2025interleaved,zhou2024gate}.
It has also been used as a reward model for vision-language alignment~\citep{li2024vlrewardbench, Yasunaga2025MultimodalRH}. These developments bring us to a key question:
\begin{center}
\textit{
Can MLLMs serve as a unified judge for assessing the \\understanding and generation ability of any-to-any modality tasks?
}
\end{center}

In other words, can MLLMs extend their human-aligned judgment capabilities—previously demonstrated in text-based~\citep{zhou2024gate} and image-based~\citep{mllm-as-a-judge,chen2024mj} tasks (see Table~\ref{tab: benchmark comparison})—to a broader range of modalities, such as images, video, and audio? Even if MLLMs cannot fully replicate human judgments, can they still provide meaningful, and reliable assessments that reduce dependence on human evaluation and guide the development of multimodal AI-generated rewards~\citep{lee2023rlaif,li2024vlrewardbench}?

\begin{table*}[!t]
    \centering
    \caption{Comparison to current works. \task uniquely incorporate diverse modalities and open-ended questions to evaluate omni-models using verified metrics that have been validated against human annotations for potential biases with an automated model arena. \judge pioneer in assessing MLLM-as-a-Judge across various modalities in \score and \pair settings. \ding{108} means that question types are mixed. See Appendix \ref{Appendix: Full Related Works} for detailed related works.}
    \label{tab: benchmark comparison}
    \resizebox{\textwidth}{!}{
    \begin{tabular}{lc|cccc|cccc|ccc}
    \toprule[1.5pt]
         \multirow{2}{*}{Benchmark} & \multirow{2}{*}{\#Size} & \multicolumn{4}{c|}{Input Modality} & \multicolumn{4}{c|}{Output Modality} & Open-ended & Verified &\multirow{2}{*}{Arena} \\
         & & Text & Image & Video & Audio & Text & Image & Video & Audio & Question & Metric & \\ \midrule
        \multicolumn{13}{c}{\textcolor{violet!80!white}{\cellcolor{gray!10!white} \textbf{\textit{Multimodal Understanding and Generation}}}}\\
        
        \midrule
        ISG \citep{chen2025interleaved} & 1,150 & \ding{52} & \ding{52} & \ding{56} & \ding{56} & \ding{52} & \ding{52} & \ding{56} & \ding{56} & \ding{52} & \ding{52} & \ding{56}\\
         MMIE \citep{xia2024mmie} & 20,103 & \ding{52} & \ding{52} & \ding{56} & \ding{56} & \ding{52} & \ding{52} & \ding{56} & \ding{56} & \ding{108} & \ding{52} & \ding{56}\\
         OmniBench \citep{li2024omnibench} & 1,142 & \ding{52} & \ding{52} & \ding{52} & \ding{52} & \ding{52} & \ding{56} & \ding{56} & \ding{56} & \ding{56} & \ding{56} & \ding{56}\\
         OmnixR \citep{chen2024omnixr} & 1,800 & \ding{52} & \ding{52} & \ding{52} & \ding{52} & \ding{52} & \ding{52} & \ding{52} & \ding{52} & \ding{108} & \ding{56} & \ding{56} \\
         Eval-Anything \citep{AlignAnything} & 264 & \ding{52} & \ding{52} & \ding{52} & \ding{52} & \ding{52} & \ding{52} & \ding{52} & \ding{52} & \ding{52} & \ding{56} & \ding{56} \\
         MixEval-X \citep{ni2024mixeval} & 8,300 & \ding{52} & \ding{52} & \ding{52} & \ding{52} & \ding{52} & \ding{52} & \ding{52} & \ding{52} & \ding{108} & \ding{56} & \ding{56} \\
         \task (ours) & 1,500 & \ding{52} & \ding{52} & \ding{52} & \ding{52} & \ding{52} & \ding{52} & \ding{52} & \ding{52} & \ding{52} & \ding{52} & \ding{52} \\
         \midrule
        \multicolumn{13}{c}{\textcolor{violet!80!white}{\cellcolor{gray!10!white} \textbf{\textit{Multimodal LLM-as-a-Judge}}}}\\
        
        \midrule
        MLLM-as-a-Judge \citep{mllm-as-a-judge} & 15,450 & \ding{52} & \ding{52} & \ding{56} & \ding{56} & \ding{52} & \ding{52} & \ding{56} & \ding{56} & \ding{56} & \ding{52} & \na \\
        VL-RewardBench \citep{li2024vlrewardbench} & 1,546 & \ding{52} & \ding{52} & \ding{56} & \ding{56} & \ding{52} & \ding{52} & \ding{56} & \ding{56} & \ding{56} & \ding{52} & \na \\
        MM-RewardBench \citep{Yasunaga2025MultimodalRH} & 5,211 & \ding{52} & \ding{52} & \ding{56} & \ding{56} & \ding{52} & \ding{52} & \ding{56} & \ding{56} & \ding{56} & \ding{52} & \na \\
        \judge (ours) & 9,000 &  \ding{52} & \ding{52} & \ding{52} & \ding{52} & \ding{52} & \ding{52} & \ding{52} & \ding{52} & \ding{52} & \ding{52} & \na \\
         \bottomrule[1.5pt]
    \end{tabular}
    }
\end{table*}

To address the aforementioned questions, we start by introducing a new benchmark, \task, to comprehensively evaluate the capabilities of MLLMs in both MMU and MMG across an \textit{any-to-any} framework. 
This benchmark consists of 15 open-ended tasks and 1,500 queries sourced from established datasets, providing an unconstrained yet categorically balanced testbed.
Next, we collect candidate responses to these queries using state-of-the-art generative models and compile query-specific checklists for evaluation.
Finally, we introduce \judge which incorporates these queries, response candidates, and checklists into \pair and \score settings, creating a standardized testbed for evaluating the effectiveness of MLLM-as-a-Judge in MMU and MMG against human-annotated judgments.
In our experiments, we specifically evaluate the judging capabilities of five advanced, widely used MLLMs on \judge, including GPT-4o, Gemini-1.5-Pro, LearnLM-1.5-Pro, and Gemini-2.0-Flash/Lite.
Experimental results reveal that MLLMs align more closely with human preferences on \pair than on \score, with both tasks benefiting from clearly fixed rubrics and the \checklist approach. Notably, while MLLMs demonstrate strong judging performance on MMU tasks, their alignment remains limited in MMG tasks, particularly in video and audio generation scenarios.
Among these models, Gemini-1.5-Pro stands out due to its robust multimodal perception, long-context reasoning, and instruction-following capabilities, achieving an average 70.6\% accuracy on \pair and  0.745 Pearson similarity on \score in MMU tasks.

To further advance \textit{any-to-any} omni-models and reward models in the multimodal domain, we present \arena, a standardized testbed for evaluating existing omni-models and multimodal reward models based on our \task and \judge benchmarks. Our experimental results, leveraging Gemini-1.5-Pro as an automated judge, demonstrate that Gemini-1.5-Pro excels among omni-models in MMU tasks, while ModaVerse achieves superior performance in MMG tasks. Once deployed, \arena will facilitate seamless participation from new models and judges in an \textit{any-to-any} fashion, while simultaneously integrating real-world votes from the broader community to collect diverse and representative judgments.

The main contributions of this paper are as follows:

\begin{itemize}[leftmargin=*]
    \item \textbf{Two Benchmarks.} We propose \task, a comprehensive benchmark for evaluating the MMU and MMG capabilities of MLLMs. Building on \task, we also introduce \judge to extensively assess the judging capabilities of MLLMs using human annotated judgments and fine-grained checklists for each sample in an \textit{any-to-any} manner from the perspectives of 
    \textit{Score Evaluation} and \textit{Pair Comparison}.
    \item \textbf{An Automated Arena for Omni Models.} We develop \arena, an automated evaluation platform for omni-models that supports diverse modality inputs and outputs, facilitating future research in multimodal generation and understanding.
    \item \textbf{Findings and Implications.} 
    Extensive experiments reveal that current MLLM-as-a-Judge partially align with human judgment while their reliability as judges for open-ended \textit{any-to-any} queries remains significantly limited. Furthermore, although MLLMs enhanced by well-constructed principle rubrics and sample-wise checklists show improvement, they still fall short due to a range of cross-modality biases and hallucinations, undermining their reliability when serving as judges.
\end{itemize}

\section{\task and \judge}
\begin{figure*}[!t]
    \centering
    \includegraphics[width=\textwidth, height=\textheight, keepaspectratio]{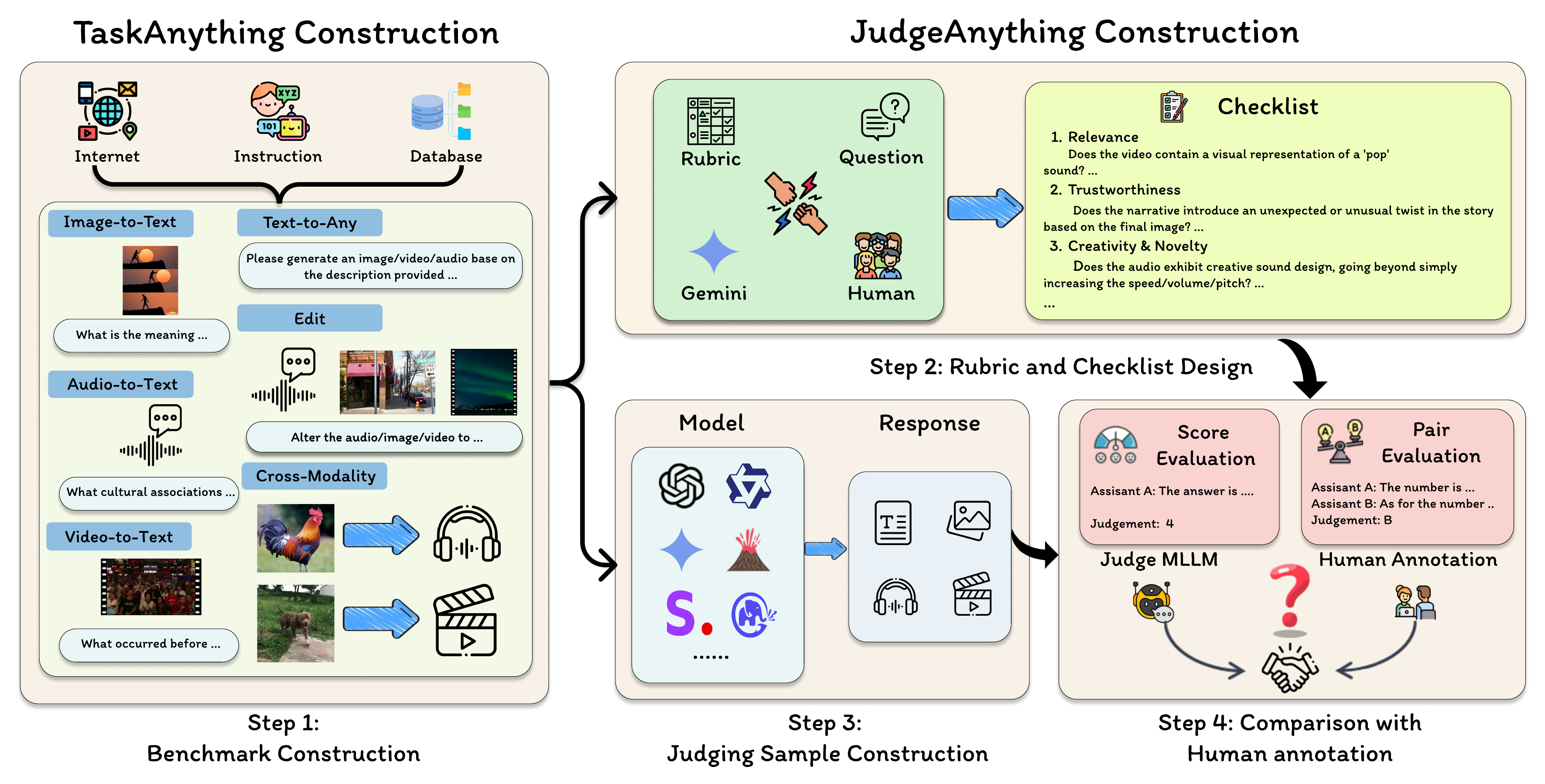}
    \vspace{-1.5em}
    \caption{The construction of \task and \judge follows a systematic four-step approach. First, we compile open-ended \textit{any-to-any} instructions from existing benchmarks and datasets, followed by rigorous human annotation to ensure sample diversity and quality in \task. Subsequently, we collect model responses and develop evaluation principles through an Human-MLLM collaborative approach, creating detailed assessment checklists for each sample. Finally, we curate instruction-responses pairs to evaluate the effectiveness of MLLM-as-a-Judge in \textit{any-to-any} generation tasks, benchmarking these automated assessments against expert human judgments.}
    \vspace{-1em}
    \label{fig: pipeline}
\end{figure*}

We introduce \task for open-ended \textit{any-to-any} generation evaluation. Based on \task, we propose \judge to evaluate whether MLLMs can serve as metrics for \textit{any-to-any} generation assessment. As shown in Figure~\ref{fig: pipeline}, we take a four-step approach to curate the entire benchmark. We provides benchmark construction details in Appendix~\ref{Appendix: Benchmark Construction}.

\subsection{\task Construction}
We collect samples from previous well-constructed and data-balanced benchmarks, as shown in Table~\ref{BenchmarkConstruction}, followed by manually selection to filter out similar and not open-source samples. For \textbf{MMU} tasks (\emph{e.g.}, Video-to-Text), we further incorporate human refinements to remove predefined constraints (\emph{e.g.}, output format) and ensure a more natural, free-form structure. For \textbf{MMG} tasks (\emph{e.g.}, text-to-video), we filter out NSFW content and low-quality queries to ensure the query can be answered. For some tasks where the field remains relatively underexplored, like visual-to-audio, we collect samples using a human-in-the-loop approach to curate diverse queries. These queries are sourced from video datasets scraped and filtered from YouTube\footnote{\url{https://youtube.com}}, including~\citep{AVSYNC15, VGGSOUND}, ensuring relevance and diversity. Finally, we successfully curate a high quality and comprehensive open-ended \textit{any-to-any} benchmark dataset $\mathbb{Q}$, comprising 1,500 queries, with each task containing 100 queries.

\subsection{Rubric and Checklist Design}
We adopt a standardized assessment framework in addition to directly prompting models to assign scores or choose for a more fine-grained evaluation. Building on recent studies~\citep{li2024generation2judgement,gu2024LLMJudgesurvey}, we define six evaluation principle rubrics for comprehensive assessment, detailed in  Appendix~\ref{Appendix: Experiment Setup Details}. To specialize these rubrics for each sample, we prompt Gemini-1.5-Pro~\citep{gemini} to generate task-specific checklists based on principle rubrics and open-ended queries. However, we observe that Gemini-1.5-Pro~\citep{gemini} demonstrates limited instruction-following capability in video and audio modalities. To mitigate this limitation, we employ a two-step process: first, generating captions for video or audio content, and then using these captions as context to refine the checklist for these tasks. Finally, we manually select 1 to 6 items from the synthetic checklist to construct the final checklist.

\begin{figure*}[!t]
    \centering
    \includegraphics[width=\textwidth, height=\textheight, keepaspectratio]{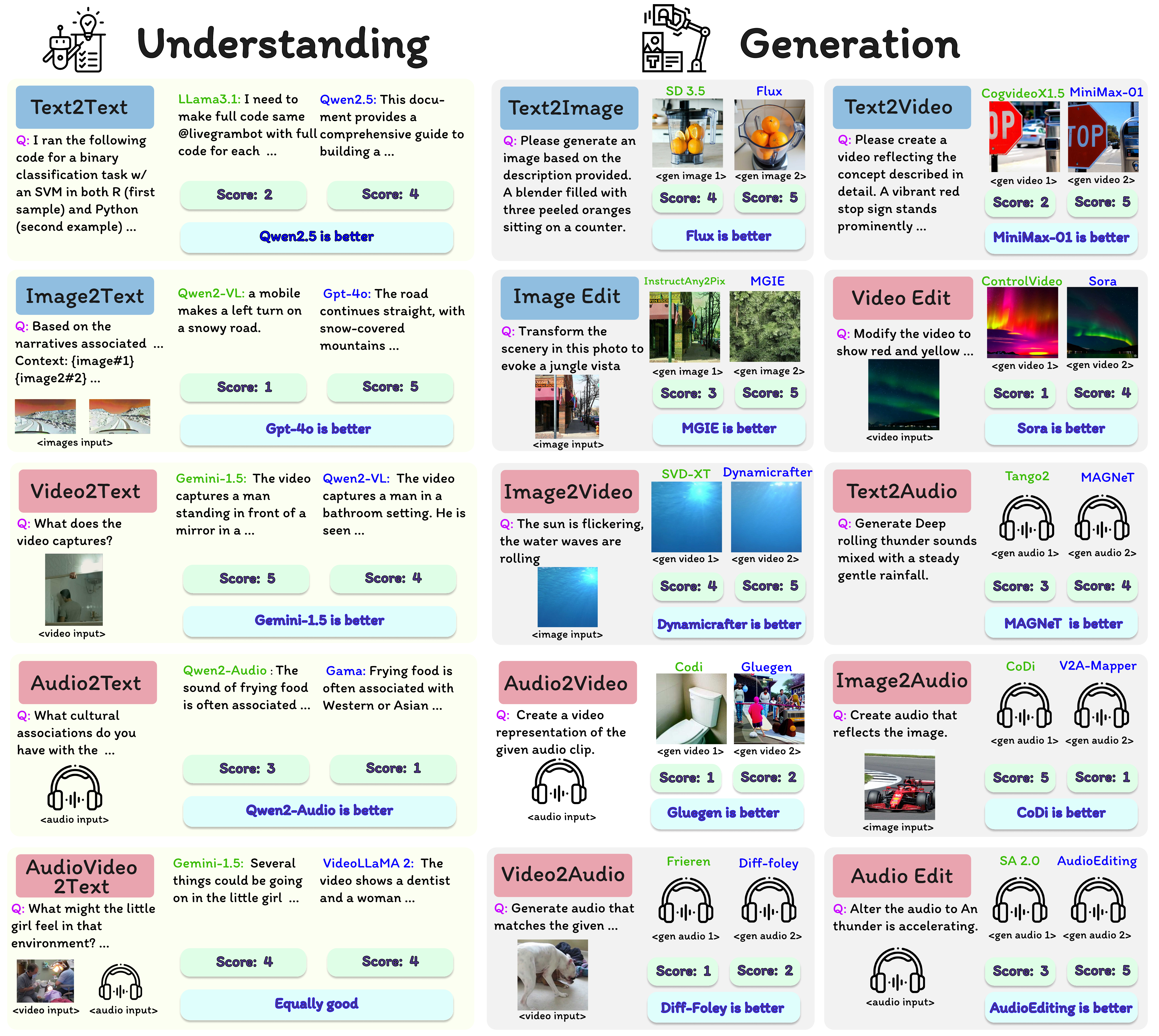}
    \vspace{-1em}
    \caption{\task and \judge comprise 15 \textit{any-to-any} combinations spanning text, image, video, and audio modalities. The \task samples are curated from established benchmarks, while responses to queries are generated using \emph{state-of-the-art} models to construct \judge in both \pair and \score settings. 
    }
    \vspace{-1em}
    \label{fig: overview}
\end{figure*}

\subsection{Judging Sample Construction}
For each \textit{any-to-any} task, we utilize four \textit{state-of-the-art} models (see Tables~\ref{tab: MMU models} and \ref{tab: MMG models} for details) to generate responses to the queries, resulting in a total response set $\mathbb{R}$ of 6,000 entries. These responses are then manually reviewed to ensure quality, with strict adherence to the NSFW guidelines. 
We use both \score and \pair to evaluate MLLM-as-a-Judge across various modalities. \score requires the model to provide an integer rating from 1 to 5, where 1 represents the worst performance and 5 represents the best. \pair, on the other hand, asks the model to select the better option or declare a tie between two candidate responses. At this stage, we construct judging samples for \score and \pair as follows:

\begin{itemize}[leftmargin=*]
    \item $\mathbb{D}_{\text{score}} = \{(Q_i, R_i) \mid Q_i \in \mathbb{Q},\ R_i \in \mathbb{R}\}$
    \item $\mathbb{D}_{\text{pair}} = \{(Q_i, R_i^1, R_i^2) \mid Q_i \in \mathbb{Q},\ R_i^1, R_i^2 \in \mathbb{R},\ R_i^1 \neq R_i^2\}$
\end{itemize}
The $\mathbb{D}_{\text{score}}$ dataset contains question-response pairs for absolute evaluation, while $\mathbb{D}_{\text{pair}}$ consists of triples for comparative assessment between two different responses. Responses $R_i^1$ and $R_i^2$ in each pair are systematically sampled from different models to ensure diverse comparisons.

\begin{wraptable}{r}{0.5\textwidth}  
    \centering
    \vspace{-1em}
    \caption{Data statistics for constructing \task and \judge. Each sample from human annotator are under cross-validation.}
    \label{tab:statistics}
    \resizebox{0.48\textwidth}{!}{  
    \begin{tabular}{clclcr}
        \toprule[1.5pt]
        \textbf{Step}& \textbf{Input} & \textbf{Num.} & \textbf{Output} & \textbf{Per Input} & \textbf{Total} \\
        \midrule
        \multirow{2}{*}{1} & Previous  & \multirow{2}{*}{/} & Open-ended  & \multirow{2}{*}{/} & \multirow{2}{*}{1500} \\
        & Benchmarks & & Instructions & & \\
        \midrule
        \multirow{2}{*}{2} & \multirow{2}{*}{Instructions} & \multirow{2}{*}{1500} & Human-annotated  & \multirow{2}{*}{19.11} & \multirow{2}{*}{28673} \\
        & & & checklist & & \\
        \midrule
        \multirow{3}{*}{3} & Instructions & 1500 & Model Responses & 4 & 6000 \\ \cmidrule{2-6}
        & Instructions + & \multirow{2}{*}{1500} & \pair & 2 & 3000 \\
        &  Responses &  & \score & 4 & 6000 \\
        \midrule
        \multirow{2}{*}{4} & \pair & 3000 & \multirow{2}{*}{Human Annotation} & $5*3$ & 45000 \\
        & \score & 6000 &  & $5*3$ & 90000 \\
        \bottomrule[1.5pt]
    \end{tabular}
    }
    \vspace{-1em}
\end{wraptable}

\subsection{Comparison with Human Annotations}
We collect the ground truth of these judging problems from 10 expert annotators. These annotators are proficient in AIGC content, with different genders, ages, and educational backgrounds to ensure data quality and diversity. 
They are require to give objective judgments that strictly following our rules and instructions without any bias that could undermine the fairness of the judgments~\citep{ye2024justice}.
We also conduct annotation on checklists to capture human preferences in a fine-grained manner. See Appendix \ref{Appendix: Human Annotation Details} for further details. \emph{\textbf{We implement cross-validation between different annotators for each sample and conduct continuous monitoring to ensure they maintain objectivity and fairness.}}

\begin{table*}[!t]
\centering
\caption{Model performance on \task. We \textbf{bold} the best and \underline{underline} the second best within each block (\overall, \textit{Rubrics}, and \textit{Checklist}).}
\resizebox{\textwidth}{!}{
\begin{tabular}{l|cc|cccc|cc|cccc}
\toprule[1.5pt]
\multirow{3}{*}{\textbf{Models}} & \multicolumn{6}{c|}{\textbf{Multimodal Understanding}} & \multicolumn{6}{c}{\textbf{Multimodal Generation}} \\
\cmidrule{2-13}
& \multicolumn{2}{c|}{Pair Comparison} & \multicolumn{4}{c|}{Score Evaluation} & \multicolumn{2}{c|}{Pair Comparison} & \multicolumn{4}{c}{Score Evaluation} \\
& w. Tie & w.o. Tie & Agreement & Pearson & Spearman & MAE & w. Tie & w.o. Tie & Agreement & Pearson & Spearman & MAE \\
\midrule
\multicolumn{13}{c}{\cellcolor{gray!10}\textbf{\textit{\overall}}}\\
\midrule
GPT-4o          
& \underline{61.20} & \underline{77.59} & \textbf{38.40} & \underline{0.461} & \underline{0.433} & \textbf{0.919} 
& 52.55 & 69.83 & 31.78 & \underline{0.444} & \underline{0.444} & \textbf{1.176} \\

Gemini-1.5-Pro          
& 60.50 & 77.14 & \underline{37.45} & 0.456 & 0.420 & 1.022 
& \underline{54.70} & \underline{70.74} & \underline{32.00} & 0.327 & 0.338 & 1.268 \\

LearnLM-1.5-Pro         
& 58.10 & 74.14 & 33.45 & 0.415 & 0.380 & 1.103 
& 52.05 & 67.32 & \textbf{33.32} & 0.328 & 0.332 & 1.285 \\

Gemini-2.0-Flash        
& 58.10 & 75.42 & 36.15 & 0.423 & 0.348 & 1.053 
& 52.15 & 68.36 & 31.20 & 0.415 & 0.417 & 1.536 \\

Gemini-2.0-Flash-Lite   
& 57.50 & 74.52 & 35.75 & 0.429 & 0.385 & 1.052 
& 46.95 & 60.93 & 30.03 & 0.421 & 0.407 & 1.482 \\

Evaluator-Fusion        
& \textbf{62.20} & \textbf{79.25} & 37.05 & \textbf{0.512} & \textbf{0.471} & \underline{0.936} 
& \textbf{54.80} & \textbf{72.05} & 25.22 & \textbf{0.492} & \textbf{0.502} & \underline{1.261} \\
\midrule
\multicolumn{13}{c}{\cellcolor{gray!10}\textbf{\textit{Rubrics}}}\\
\midrule
GPT-4o          
& 63.38 & 76.68 & \textbf{39.98} & \underline{0.576} & \underline{0.568} & \textbf{0.935}
& 32.27 & 53.03 & 28.95 & 0.383 & 0.392 & 1.365 \\

Gemini-1.5-Pro          
& \textbf{69.40} & \textbf{82.74} & 39.58 & 0.565 & 0.551 & \underline{0.949}
& \textbf{53.01} & \textbf{68.67} & \underline{35.60} & \underline{0.406} & 0.408 & \textbf{1.203} \\

LearnLM-1.5-Pro         
& 64.77 & 77.30 & \underline{39.62} & 0.552 & 0.540 & 0.973 
& \underline{52.66} & \underline{67.18} & \textbf{35.83} & 0.387 & 0.389 & \underline{1.222} \\

Gemini-2.0-Flash        
& 47.75 & 68.71 & 37.53 & 0.491 & 0.473 & 1.124 
& 41.89 & 61.00 & 26.87 & 0.350 & 0.353 & 1.706 \\

Gemini-2.0-Flash-Lite   
& 54.73 & 70.77 & 36.64 & 0.492 & 0.495 & 1.152 
& 40.45 & 59.25 & 28.87 & 0.405 & \underline{0.414} & 1.571 \\

Evaluator-Fusion        
& \underline{66.73} & \underline{81.08} & 37.17 & \textbf{0.618} & \textbf{0.627} & 0.989
& 49.26 & 65.89 & 24.42 & \textbf{0.502} & \textbf{0.522} & 1.349 \\
\midrule
\multicolumn{13}{c}{\cellcolor{gray!10}\textbf{\textit{Checklist}}}\\
\midrule
GPT-4o          
& 60.77 & 74.75 & 42.03 & 0.623 & 0.608 & 0.844
& 30.27 & 51.90 & 30.63 & 0.343 & 0.340 & 1.295 \\

Gemini-1.5-Pro          
& \textbf{70.60} & \textbf{84.07} & \textbf{54.57} & \textbf{0.745} & \textbf{0.729} & \textbf{0.629} 
& \textbf{53.79} & \underline{69.97} & \textbf{41.77} & \underline{0.494} & \underline{0.495} & \textbf{1.036} \\

LearnLM-1.5-Pro         
& 64.52 & 76.85 & \underline{43.45} & 0.646 & 0.631 & 0.843 
& 52.48 & 68.09 & \underline{38.43} & 0.445 & 0.447 & 1.112 \\

Gemini-2.0-Flash        
& 53.93 & 71.87 & 40.31 & 0.554 & 0.543 & 0.979
& 50.23 & 67.74 & 35.16 & 0.476 & 0.482 & 1.282 \\

Gemini-2.0-Flash-Lite   
& 56.22 & 70.74 & 39.50 & 0.551 & 0.552 & 0.979
& 48.53 & 65.66 & 35.99 & 0.450 & 0.460 & 1.165 \\

Evaluator-Fusion        
& \underline{66.55} & \underline{80.68} & 42.79 & \underline{0.687} & \underline{0.687} & \underline{0.816}
& \underline{53.37} & \textbf{70.71} & 30.05 & \textbf{0.562} & \textbf{0.572} & \underline{1.069} \\
\bottomrule[1.5pt]
\end{tabular}}
\label{tab:category}
\vspace{-1em}
\end{table*}

\section{Experiments and Analysis}
Using \judge, we conduct experiments to evaluate the judging abilities of MLLMs (\emph{i.e.}, MLLM-as-a-Judge) across modalities in both \score and \pair settings. 
\subsection{Experimental Setup}
\label{Sec:Exp}


\header{Judging Models.} 
We utilize five advanced proprietary models—GPT-4o~\citep{2024gpt}, Learnlm-1.5-pro-experimental~\citep{learnlm}, Gemini-1.5-Pro~\citep{gemini}, Gemini-2.0-Flash, and Gemini-2.0-Flash-lite—selected for their strong understanding, robust generative performance across multiple modalities, and effective instruction-following capabilities.
Ultimately, we compare the judging models with the \textit{evaluator-fusion}.
We define \textit{evaluator-fusion} as the average score in \score and majority-voting in \pair. 
To clarify, given that GPT-4o cannot receive both audio and visual content, we leverage GPT-4o-audio-preview as a replacement for audio-visual task.
Also, we have experimented with the state-of-the-art open-source omni-models including Baichuan-Omni-1.5~\citep{li2025baichuan} and VideoLlama2~\citep{cheng2024videollama2}. 
However, none of these models could handle long-context input or generate meaningful feedback, elaborated in Appendix~\ref{Appendix: Model4Judge}.

\header{Three Baselines.} We provide three different judging settings: The \textbf{\overall} setting leverages a direct judging approach, where models first provide reasoning and then deliver a final judgment. The \textbf{\rubric} setting introduces well-defined general foundation rubrics within context and requires models to judge based on fine-grained rubrics before making a final judgment. In the \textbf{\checklist} setting, MLLMs are provided with detailed checklists curated through a human-in-the-loop process and must first evaluate responses based on these checklists before delivering their final judgment.. For all settings, we employ an \textit{``Analyze-then-Judge''} chain-of-thought~\citep{wei2022chain} pattern to elicit models' judging capabilities. 
To improve robustness and mitigate variance, we sample all judgments three times with slightly modified prompts and take the average of the results.

\header{Implementations.}
We set the temperature to 0.7 for all judging models, as previous research~\citep{liu2023g,mllm-as-a-judge} has reported a high correlation with human annotators at this setting.
All models are configured to generate structured outputs across both evaluation paradigms. For the \score setting, we provide detailed explanatory descriptions for each integer value on the 1-5 scale, enabling informed judgments based on explicit criteria. For the \pair setting, we offer three categorical choices: \textit{``first''}, \textit{``second''}, and \textit{``tie''}, conducting experiments with switched response positions to mitigate potential position bias. 
All experiments for judging models are replicated three times, with the averaged score (for \score) and majority selection (for \pair) used to calculate final results. 
We analyze performance using four established metrics—Agreement, Pearson correlation, Spearman correlation \citep{lee1988thirteen}, Mean Absolute Error (MAE) for the \score setting, and accuracy for the \pair setting. See Appendix~\ref{Appendix: Experiment Setup Details} for comprehensive experimental protocols.


\subsection{Quantitative Results}
\label{Sec: quantitative results}


\header{Gemini-1.5-Pro is the best evaluator in \textit{any-to-any} task evaluation in our experiments.} Table~\ref{tab:category} shows that Gemini-1.5-Pro outperforms other judging models, including GPT-4o and the advanced Gemini-2.0 series, particularly in \checklist settings, achieving a 0.745 Pearson similarity under \score and 70.60\% agreement under \pair with human annotators. This supports the trend that larger models exhibit superior human-like judgment simulation. Additionally, GPT-4o struggles with MMG tasks, likely due to its limited cross-modal reasoning, making it a suboptimal unified judge for \textit{any-to-any} evaluations. Evaluator-Fusion, which aggregates multiple MLLMs through majority voting, achieves state-of-the-art alignment in both \score and \pair settings. While its performance declines with fine-grained \checklist evaluations, it still ranks second-best across MMU and MMG tasks, demonstrating its effectiveness in multi-modal evaluation alignment.


\begin{figure*}[!ht]
    \centering
    \includegraphics[width=.98\textwidth, height=\textheight, keepaspectratio]{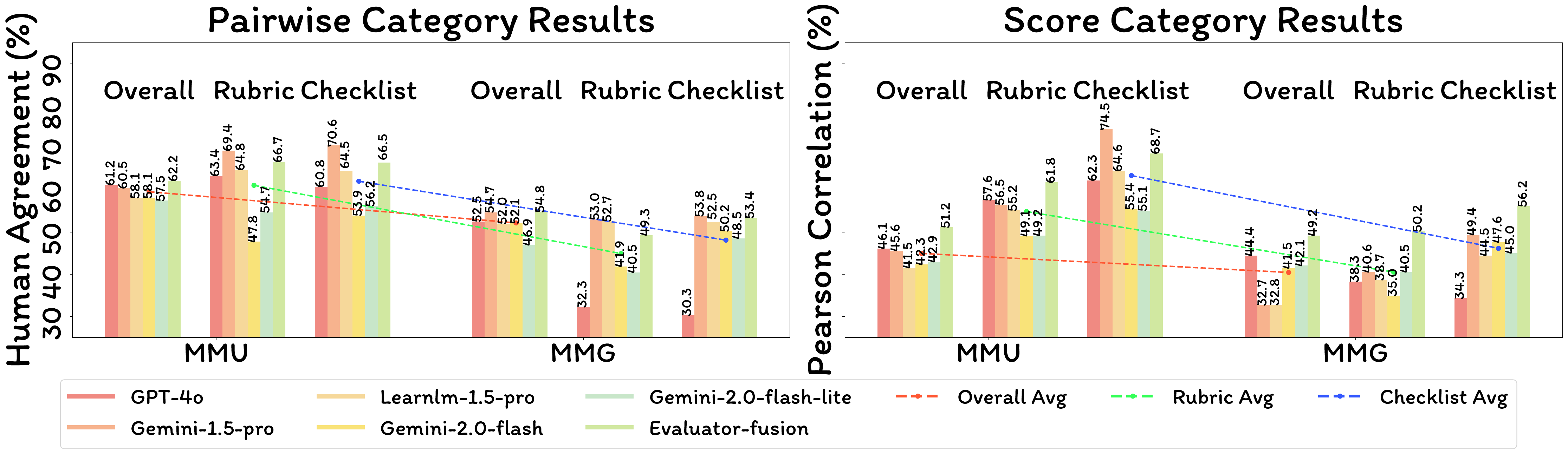}
    \caption{Visualization of MMU and MMG categories with human agreement data.
  \textbf{Left:} Accuracy scores for the \pair setting across two categories. 
  \textbf{Right:} Agreement scores for the \score setting across two categories. 
  The dotted line connects the same baseline from MMU to MMG to highlight the trend.}
    \vspace{-1em}
    \label{fig: CategoryAnalyse}
\end{figure*}

\header{MLLM-as-a-Judge performs better in MMU rather than MMG tasks.}
\label{header:category}
As shown in Table~\ref{tab:category} and Figure~\ref{fig: CategoryAnalyse}, MLLMs' judgments align more closely with human evaluations in MMU tasks compared to MMG tasks in both \score and \pair settings, particularly in text-to-text and image-to-text scenarios. Moreover, we observe that judging models benefit significantly more from fine-grained evaluation criteria in MMU than MMG tasks. We attribute this pattern to the fact that current judging models perform better at evaluating tasks they themselves are capable of executing, leading to more accurate and fair assessments in these domains. This finding corresponds with previous research indicating that understanding is the foundation of generation capability. Consequently, the ability to effectively judge open-ended MMU tasks may develop before MMG evaluation capacity, as the inherent complexity and variability within text is substantially lower than in other modalities.

\finding{Judging models are more reliable on MMU task, and \checklist can improve the alignment.}

\begin{wrapfigure}{r}{0.5\textwidth}
    \centering
    \vspace{-2em}
    \includegraphics[width=0.46\textwidth, height=0.46\textheight, keepaspectratio]{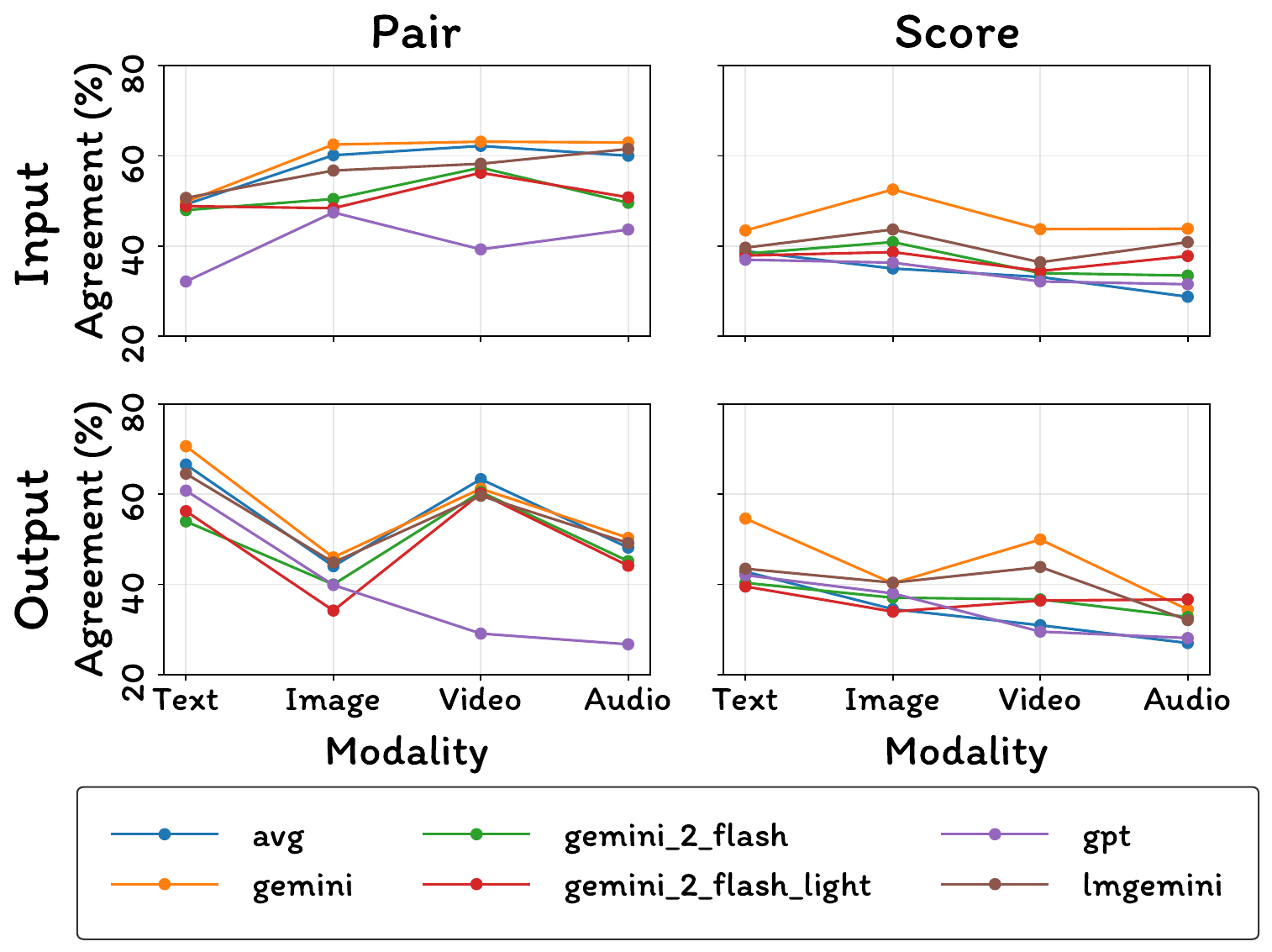}
    \caption{Visualization of modality effect on human agreement across two settings using \method. 
    }
    \label{fig:modality effect}
    \vspace{-1em}
\end{wrapfigure}

\header{Less-aligned modalities like video and audio pose significant challenges to MLLM-as-a-Judge in cross-modality judging.}
\label{header:modality}
Diving deeper into MLLM-as-a-Judge across modalities, Table~\ref{tab:taskwise} shows that current judging models' performance declines when it comes to low-frequency cross-modality tasks like image-to-audio and audio-to-video.
As shown in Figure~\ref{fig:modality effect}, different output modalities matter more compared to input modality. While input modality results show consistent, low-variance agreement across all modalities, output modality results reveal a distinct downward trend, with strong alignment in text and image modalities progressively declining in video and audio modalities. Given that current evaluation models are predominantly trained on text-centric or image-centric scenarios, their capabilities in cross-modal assessment remain nascent, reflecting an emergent property still in early development. To enhance this capability, we recommend incorporating a more diverse range of cross-modality judging samples into training.


\finding{\textbf{Output} modality matters more for MLLM-as-a-Judge, with results showing a decreasing trend from well-aligned to less-aligned modality.}

\header{MLLM-as-a-Judge benefits from fine-grained rubrics and checklists, enhancing human-aligned judgments.}  
As shown in Figure~\ref{fig: CategoryAnalyse}, most judging models benefit from well-curated, sample-wise judging checklists, with \checklist approaches yielding more human-aligned evaluations than overall judgments or fixed foundation rubrics. This effect is particularly pronounced in Gemini, which leverages its \textbf{1M-token long-context window} and \textbf{strong instruction-following capabilities} to provide highly human-like judgments. However, these detailed guidelines present a double-edged sword: while they enhance judgment alignment, they can also hinder performance. Notably, \textbf{GPT-4o} experiences a decline in accuracy when using \rubric and \checklist formats compared to direct judgment.

\begin{table*}[!t]
    \centering
    \caption{Detailed breakdown on each \textit{any-to-any} generation tasks on \judge. Well-constructed rubrics and \method enhance MLLMs' alignment with human when serving as judges. We \textbf{bold} the best.}
    \resizebox{\textwidth}{!}{
    \begin{tabular}{ll|c|ccccc|cccccccccc}
        \toprule[1.5pt]
        \multirow{2}{*}{\textbf{Models}} &\multirow{2}{*}{\textbf{Setting}}  & \multirow{2}{*}{\textbf{Overall}} & \multicolumn{5}{c|}{\textbf{Multimodal Understanding}} & \multicolumn{10}{c}{\textbf{Multimodal Generation}} \\
         & & & T$\rightarrow$T & I$\rightarrow$T & V$\rightarrow$T & A$\rightarrow$T & V+A$\rightarrow$T  & T$\rightarrow$I & T$\rightarrow$V & T$\rightarrow$A & I$\rightarrow$I & I$\rightarrow$V & I$\rightarrow$A & V$\rightarrow$V & V$\rightarrow$A & A$\rightarrow$V & A$\rightarrow$A  \\
         \midrule
        \multicolumn{18}{c}{\textcolor{violet!80!white}{\cellcolor{gray!10!white} \textbf{\textit{\pair}}}}\\
        
        \midrule
        \multirow{3}{*}{GPT-4o}
         & \overall & \textbf{55.43} & \textbf{52.50} & \textbf{73.00} & \textbf{58.00} & 69.50 & 53.00 & \textbf{52.50} & \textbf{56.00} & 26.00 & 49.00 & 42.00 & \textbf{68.00} & \textbf{78.00} & 38.00 & \textbf{58.50} & \textbf{57.50} \\
         & \rubric & 42.64 & 52.33 & 68.00 & 53.08 & \textbf{77.58} & 65.92 & 17.33 & 25.25 & \textbf{46.83} & \textbf{69.17} & \textbf{43.42} & 15.67 & 3.17 & \textbf{38.92} & 48.92 & 14.00 \\
         & \checklist & 40.43 & 46.92 & 67.33 & 52.08 & 63.25 & \textbf{70.92} & 10.50 & 25.42 & 45.75 & \textbf{69.17} & 43.25 & 10.08 & 3.08 & 38.67 & 44.50 & 12.25 \\
        \midrule

        \multirow{3}{*}{Gemini-1.5-Pro}
         & \overall & 56.63 & 49.00 & \textbf{79.00} & 57.50 & 68.50 & 48.50 & \textbf{52.00} & 56.00 & \textbf{38.00} & 57.00 & 39.00 & \textbf{69.00} & 88.50 & 33.00 & \textbf{47.00} & \textbf{67.50} \\
         & \rubric & 58.47 & 62.33 & 75.67 & 56.08 & \textbf{82.42} & 72.17 & 18.42 & 68.50 & 36.50 & \textbf{73.25} & 38.08 & 66.25 & 89.33 & 34.00 & 43.25 & 62.50 \\
         & \checklist & \textbf{59.39} & \textbf{65.75} & 78.67 & \textbf{58.42} & 70.92 & \textbf{72.58} & 26.50 & \textbf{69.92} & 37.17 & 65.42 & \textbf{39.75} & 66.25 & \textbf{89.83} & 33.42 & 45.33 & 64.33 \\
        \midrule

        \multirow{3}{*}{Gemini-2.0-Flash}
         & \overall & \textbf{54.13} & \textbf{43.50} & \textbf{67.50} & \textbf{59.00} & \textbf{69.00} & 51.50 & \textbf{46.50} & 58.00 & \textbf{50.50} & \textbf{51.00} & 43.50 & 57.50 & 77.00 & 41.00 & \textbf{58.00} & 38.50 \\
         & \rubric & 43.84 & 36.75 & 50.58 & 42.33 & 56.58 & \textbf{67.58} & 33.42 & 55.00 & 37.50 & 48.92 & 44.08 & 31.50 & 57.00 & \textbf{41.58} & 45.50 & 24.42 \\
         & \checklist & 51.47 & 39.33 & 59.75 & 47.00 & 66.08 & \textbf{67.58} & 46.42 & \textbf{62.83} & 43.33 & 33.50 & \textbf{48.25} & \textbf{60.25} & \textbf{78.17} & 38.33 & 52.67 & \textbf{38.58} \\
        \midrule
        \multicolumn{18}{c}{\textcolor{violet!80!white}{\cellcolor{gray!10!white} \textbf{\textit{\score}}}}\\
        
         \midrule
        \multirow{3}{*}{GPT-4o}
         & \overall & 33.98 & 37.25 & 46.50 & 35.75 & 33.25 & 39.25 & 28.25 & \textbf{41.50} & 30.25 & 37.50 & 29.25 & 28.25 & 13.75 & 28.25 & \textbf{47.00} & \textbf{33.00} \\
         & \rubric & 32.63 & 46.58 & 46.50 & \textbf{37.13} & 31.58 & 38.13 & 30.25 & 30.25 & \textbf{34.29} & 32.00 & 28.67 & \textbf{30.96} & 12.29 & 24.79 & 33.21 & 25.79 \\
         & \checklist & \textbf{34.43} & \textbf{48.63} & \textbf{48.25} & 35.17 & \textbf{34.04} & \textbf{44.04} & \textbf{34.00} & 30.33 & 34.00 & \textbf{41.04} & \textbf{30.67} & 25.33 & \textbf{21.96} & \textbf{27.46} & 35.04 & 25.50 \\
        \midrule

        \multirow{3}{*}{Gemini-1.5-Pro}
         & \overall & 33.82 & 40.75 & 47.25 & 38.00 & 34.25 & 27.00 & 25.25 & 17.00 & 29.50 & 39.75 & 26.50 & 32.00 & 34.50 & \textbf{35.75} & 50.75 & 29.00 \\
         & \rubric & 36.93 & 44.00 & 47.21 & 35.17 & 33.29 & 38.25 & 36.17 & 36.17 & \textbf{33.79} & 43.08 & 41.63 & 33.83 & 32.96 & 32.42 & 45.21 & 22.21 \\
         & \checklist & \textbf{46.04} & \textbf{58.88} & \textbf{59.54} & \textbf{42.83} & \textbf{48.13} & \textbf{63.46} & \textbf{44.50} & \textbf{44.50} & 33.71 & \textbf{43.71} & \textbf{62.38} & \textbf{44.58} & \textbf{40.25} & 28.46 & \textbf{52.58} & \textbf{30.83} \\
        \midrule

        \multirow{3}{*}{Gemini-2.0-Flash}
         & \overall & 32.85 & 35.00 & 40.75 & \textbf{45.00} & 29.75 & 30.25 & 21.50 & 26.75 & 24.00 & 36.50 & \textbf{41.00} & \textbf{47.00} & 11.25 & \textbf{42.75} & 46.50 & 14.75 \\
         & \rubric & 30.42 & 46.92 & 45.21 & 38.58 & 27.17 & 29.75 & 35.63 & 31.50 & 32.50 & 23.13 & 32.71 & 32.17 & 11.79 & 33.00 & 24.13 & 12.42 \\
         & \checklist & \textbf{36.88} & \textbf{47.21} & \textbf{47.04} & 40.29 & \textbf{36.92} & \textbf{35.00} & \textbf{36.92} & \textbf{35.00} & \textbf{34.13} & \textbf{37.13} & 37.75 & 41.63 & \textbf{19.83} & 38.50 & \textbf{54.00} & \textbf{16.75} \\
         \bottomrule[1.5pt]
    \end{tabular}
    }
    \label{tab:taskwise}
\end{table*}


\subsection{In-Depth Analysis}
\label{sec: Qualitative Results}

\begin{wrapfigure}{r}{0.4\textwidth}
  \centering
  \vspace{-2em}
    \includegraphics[width=0.38\textwidth]{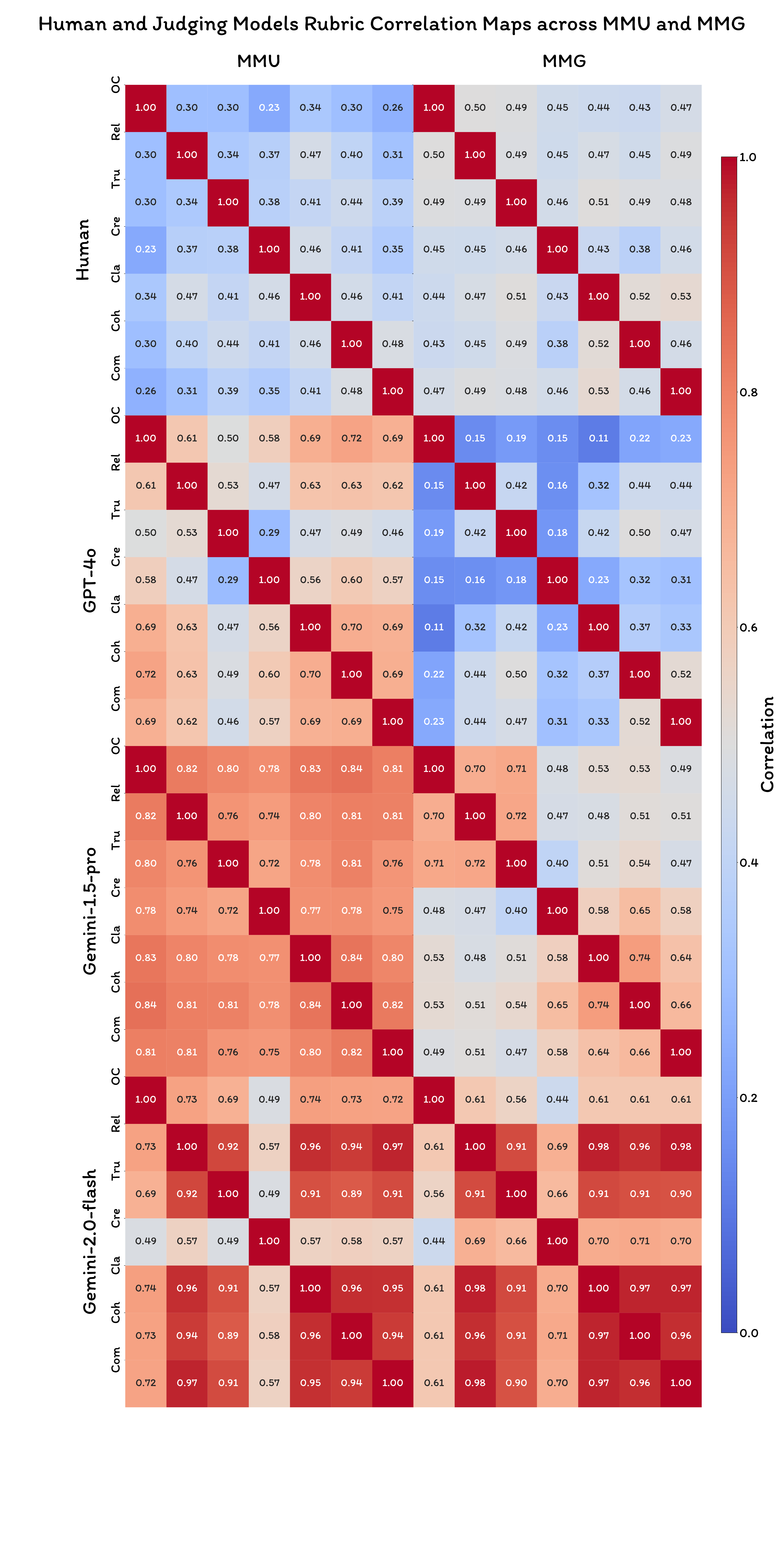}
    \vspace{-4em}
    \caption{Correlation heatmaps for different judges: OC (Overall Choice), Rel (Relevance), Tru (Trustworthiness), Cre (Creativity \& Novelty), Cla (Clarity), Coh (Coherence), Com (Completeness).}
    \vspace{-4em}
  \label{fig: corr_map}
\end{wrapfigure}

We conduct a more nuanced analysis of the underlying factors behind these quantitative results, seeking to understand both the strengths and limitations of MLLM-as-a-Judge approaches. We examine what contributes to their strong performance in certain contexts and what fundamental challenges undermine their reliability in others.

\header{How fine-grained rubrics serve as \textit{``two-blade sword''} for judgment?}
Tables~\ref{tab:category} and~\ref{tab:taskwise} illustrate that incorporating \checklist approaches generally improves judging models' performance, enabling more accurate assessments. For example, in a video-to-text task (Figure~\ref{Fig:ChecklistImprove} in Appendix), when the judging model is asked to evaluate a response's creativity, the \checklist framework helps calibrate the score from 4 to 2, bringing it more in alignment with user evaluations.
However, applying \checklist can sometimes yield counterproductive results.
When examining the reasoning chains employed during judgment formation, we discovered that such fine-grained rubrics may introduce misleading information and trigger serious hallucinations for certain tasks.
A representative example appears in Figure~\ref{Fig:ChecklistHallu}, where a human-refined \checklist designed to evaluate gun imagery without harmful content leads Gemini-1.5-Pro to misinterpret the gun figure itself as inherently harmful content, resulting in an inappropriately low score on the trustworthiness rubric.

Table~\ref{tab:taskwise} also shows that GPT-4o is significantly affected by \checklist in certain MMG tasks(\emph{e.g.} Audio edit and video edit tasks) under \pair setting. The case study in Figure~\ref{Fig: GPT Failure} investigates this issue. We find that GPT-4o incorrectly fabricate information not present in the provided model responses, thus impairing its ability to reliably distinguish between query inputs and model outputs. This tendency toward hallucination led GPT-4o to significantly underperform compared to other evaluators, especially affecting its effectiveness within the \pair setting evaluation framework. As shown in Table~\ref{tab:Tie_Count}, GPT-4o disproportionately selects a certain choice when using subdivided \rubric criteria.

\begin{wrapfigure}{r}{0.48\textwidth}
    \centering
    \vspace{-1.5em}
    \includegraphics[width=0.48\textwidth]{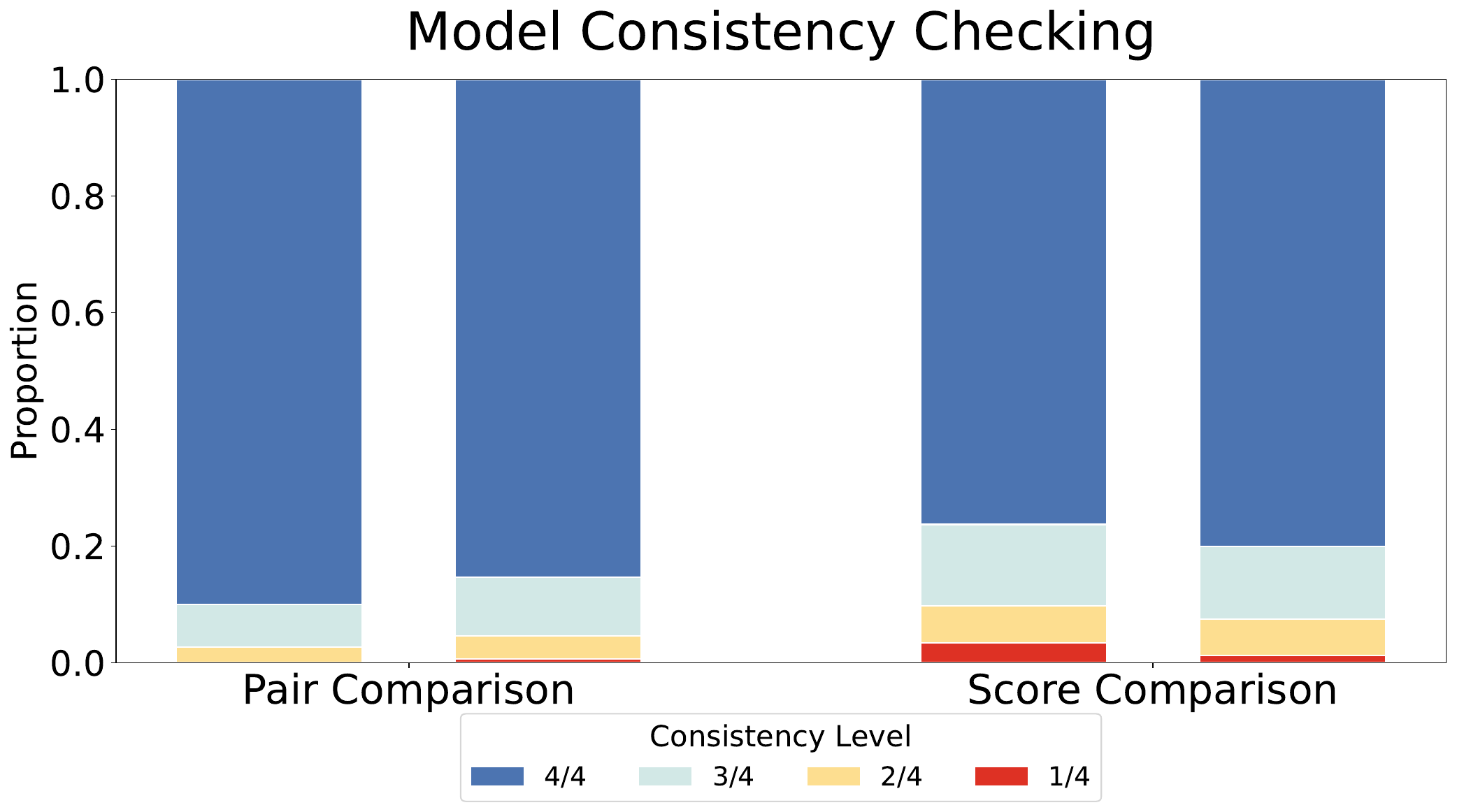}
    \caption{Consistency checking across four repeated experiments with identical prompts on Gemini-1.5-Pro (left) and Gemini-2.0-Flash (right).}
    \label{fig:mcc analysis}
    \vspace{-1.5em}
\end{wrapfigure}

\header{Why do fine-grained \rubric and \checklist enhance \score more significantly than \pair assessment?}
As previously mentioned, MLLM-as-a-Judge performance varies considerably across settings, modalities and tasks. After we decompose our benchmark into minimal task compositions (Table~\ref{tab:taskwise}) we find a particularly notable distinction between \score and \pair evaluation setting scenarios. The result shows fine-grained \rubric and \checklist can enhance \score in general, but fail to and even harm the alignment of \pair. 

To systematically quantify this discrepancy, we calculate the choice selection rate across different settings (Table~\ref{tab:Tie_Count}) and generate \overall-\rubric correlation heatmaps comparing judging models and human annotators (Figure~\ref{fig: corr_map}). These quantitative results highlight a key factor underlying the discrepancy: the \textbf{contextual relevance} of the evaluation criteria.

Focusing on MMU tasks, fine-grained \rubric effectively prompts human annotators to assess generated responses from multiple perspectives, leading to a relatively low correlation between individual rubrics and the overall choice. In contrast, judging models tend to rely on their overall assessment, showing limited responsiveness to \rubric.
 
For MMG tasks, many \rubric dimensions become inherently less meaningful in certain settings (\emph{e.g.}, assessing \textit{“Coherence”} for video-to-audio tasks). When model responses perform similarly under these less effective rubrics (Figures~\ref{Fig: Case4} and \ref{Fig: Case Study 5}), human evaluators tend to prioritize responses with better overall performance, whereas judging models often default to \textit{“tie”}. As shown in Figure~\ref{fig: corr_map}, judging model correlation drops when transitioning from MMU to MMG tasks, whereas the average correlation of human annotators increases by approximately 0.2. 

In a nutshell, judging models and human annotators exhibit \textbf{opposite reactions} to the fine-grained \rubric framework. In MMU tasks, human annotators' assessments of rubric dimensions remain relatively independent from their overall evaluation, while judging models rely more on their overall assessment. However, in MMG tasks, as the contextual relevance of rubric dimensions decreases, human annotators place greater emphasis on overall performance, whereas judging models' evaluations across fine-grained rubrics become more detached from their overall assessments, resulting in a weaker correlation between their rubric-based choices and overall preferences.

\finding{The effectiveness gap between \overall and fine-grained evaluation frameworks under \pair setting stems from the contextual applicability of standardized fine-grained criteria across diverse outputs.}

While \checklist approaches demonstrably enhance judgment alignment in \score setting, our findings highlight the need for more dynamic and context-sensitive evaluation frameworks for \pair assessment. The cognitive divergence between human judgment and model-based analytical assessment in \pair represents a significant challenge for developing unified evaluation metrics across the full spectrum of multimodal tasks. We present detailed case studies in Appendix~\ref{Appendix:CaseStudy} that provide task-specific analyses of these evaluation patterns.








\header{Inert inconsistency undermines reliability of MLLM-as-a-Judge.}
To evaluate the consistency of decision-making, we perform four repeated tests under the \overall and \checklist baselines, calculating the Majority Consistency Criterion (MCC) ratios for each test. This analysis compares the consistency of Gemini-1.5-Pro and Gemini-2.0-Flash across different settings.

As shown in Figure~\ref{fig:mcc analysis}, Gemini-1.5-Pro achieves higher consistency (0.9) under the \pair setting compared to Gemini-2.0-Flash. However, in the \score setting, Gemini-1.5-Pro's consistency drops to 0.763, while Gemini-2.0-Flash maintains a score above 0.8.
Figures \ref{fig: rubricMCC1_5} and \ref{fig: rubricMCC2} reveal that Gemini-1.5-Pro performs well in \overall and relevance, but its consistency declines in other rubrics. Gemini-2.0-Flash shows stable performance in \overall, relevance, and trustworthiness, but is unstable in rubrics like clarity, coherence, and completeness, with MCC ratios below 0.6 in \pair.
These results suggest that Gemini-1.5-Pro is more reliable in \pair evaluations but struggles with \score. The decline in \rubric performance indicates that additional rubrics may introduce uncertainty in decision-making.

\section{\arena}
\label{Arena}

\begin{figure*}[!t]
    \centering
    \includegraphics[width=\textwidth]{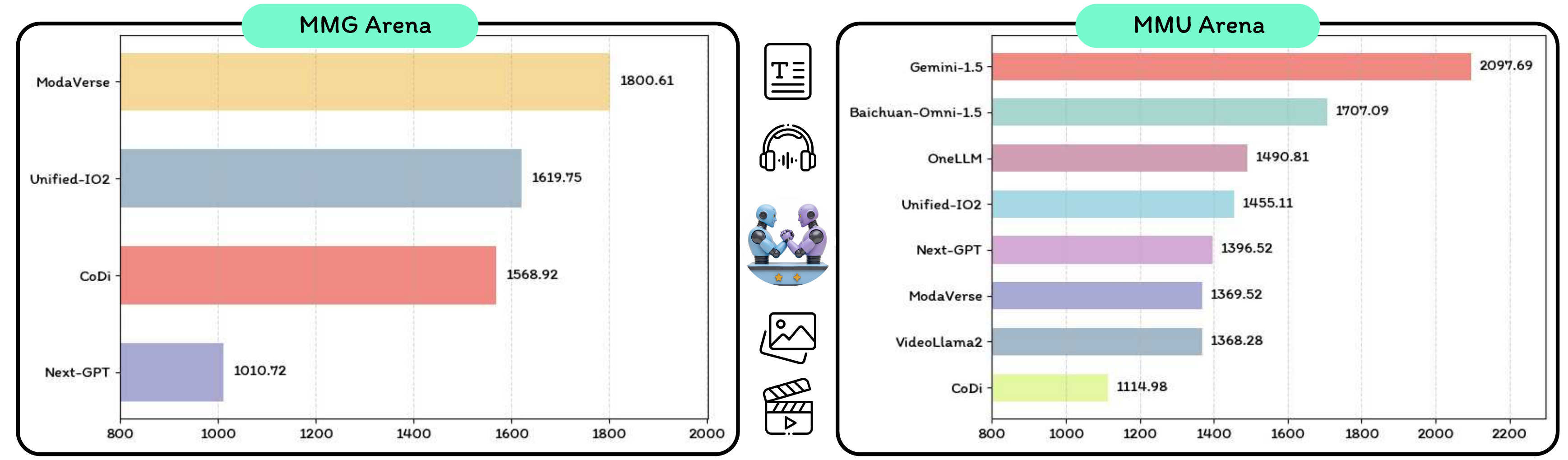} 
    \caption{Overview of our \arena. ModaVerse outperforms other omni-models on open-ended MMG tasks. For MMU, Gemini-1.5-pro shows incredibly performance with its long-context and cross-modality reasoning capability.}
    \vspace{-1em}
    \label{fig: Arena}
\end{figure*}

Based on \task and \judge, we propose \arena, a standardized platform to reliably assess the performance of omni-models. \arena leverages open-ended queries in \task and operates through a pairwise comparison mechanism, where judging models or users are presented with two responses and asked to determine the superior outcome. Participants can add their omni-models into \arena, select their preferred results, and even introduce innovative questions, allowing for dynamic, creative testing.








\header{\arena Setups.} As previously mentioned, \textit{any-to-any} tasks can be categorized into two distinct subtypes: MMU and MMG, leading to \arena being structured into two separate parts. In our experiment, Gemini-1.5-Pro serves as judging models for its superior performance in \judge. 
After obtaining the automatic judging results in \arena, we employ the \textbf{ELO Rating System}~\citep{elo} (see Appendix \ref{appendix: elo} for technical details)to establish a dynamic ranking platform for evaluating models on two sub-arenas. After each match between two models, the ELO ratings are updated based on the outcome, providing a quantitative measure of model performance based on pairwise comparisons.

\header{Experiment Results.} As shown in Figure~\ref{fig: Arena}, Gemini-1.5-pro achieves the highest ELO score in the MMU arena. Notably, MMU expertise surpasses omni-models in \arena, highlighting the superior performance of specialized models in this task. Interestingly, Next-GPT~\citep{nextgpt} excels in MMU tasks but underperforms in MMG tasks. This discrepancy arises from its limited control mechanisms and insufficient instruction-following ability, which impedes its capacity to generate multimodal outputs in MMG tasks. The lack of access to open-source models remains a significant constraint. However, with \arena's growing capability, combined with increasingly refined benchmarks, we expect the continuous introduction of new models and the ongoing enhancement of the evaluation framework for a better future.


\section{Discussion and Conclusion}



In summary, this work presents a holistic assessment of MLLMs as a unified metric for MMU and MMG tasks by introducing two benchmarks spanning 15 types of \textit{any-to-any} tasks in \pair and \score settings. Our comprehensive experiments reveal the limitations of current advanced MLLMs when serving as judges, uncovering biases and potential issues that provide insights for future research.

LLM-as-a-Judge has been widely utilized for automated open-ended natural language generation assessment and served as supervised rewards in model training. However, as AI capabilities expand beyond text to encompass rich multimodal interactions, we urgently need evaluation frameworks that reflect human values across modalities. Our findings highlight the critical need for developing more sophisticated cross-modal evaluation protocols that can better capture nuanced human preferences. We hope our work can provide a standard testbed to streamline the evaluation process, reduce dependence on human labor, and facilitate the development of more human-aligned \textit{any-to-any} generative models.


\bibliography{custom}

\begin{thebibliography}{167}
\providecommand{\natexlab}[1]{#1}
\providecommand{\url}[1]{\texttt{#1}}
\expandafter\ifx\csname urlstyle\endcsname\relax
  \providecommand{\doi}[1]{doi: #1}\else
  \providecommand{\doi}{doi: \begingroup \urlstyle{rm}\Url}\fi

\bibitem[Abdin et~al.(2024)Abdin, Aneja, Awadalla, Awadallah, Awan, Bach,
  Bahree, Bakhtiari, Bao, Behl, et~al.]{abdin2024phi3}
Abdin, M., Aneja, J., Awadalla, H., Awadallah, A., Awan, A.~A., Bach, N.,
  Bahree, A., Bakhtiari, A., Bao, J., Behl, H., et~al.
\newblock Phi-3 technical report: A highly capable language model locally on
  your phone.
\newblock \emph{arXiv preprint arXiv:2404.14219}, 2024.

\bibitem[AI(2024)]{recraft_blog}
AI, R.
\newblock How to create sota image generation with text: Recraft's ml team
  insights.
\newblock
  \url{https://www.recraft.ai/blog/how-to-create-sota-image-generation-with-text-recrafts-ml-team-insights},
  2024.

\bibitem[Anthropic(2023)]{anthropic2023claude3_5sonnet}
Anthropic.
\newblock Claude 3.5 sonnet model card addendum.
\newblock Online, 2023.
\newblock URL
  \url{https://www-cdn.anthropic.com/fed9cc193a14b84131812372d8d5857f8f304c52/Model_Card_Claude_3_Addendum.pdf}.

\bibitem[Antol et~al.(2015)Antol, Agrawal, Lu, Mitchell, Batra, Zitnick, and
  Parikh]{antol2015vqa}
Antol, S., Agrawal, A., Lu, J., Mitchell, M., Batra, D., Zitnick, C.~L., and
  Parikh, D.
\newblock Vqa: Visual question answering.
\newblock In \emph{Proceedings of the IEEE international conference on computer
  vision}, pp.\  2425--2433, 2015.

\bibitem[Audio(2024)]{stableaudio2024stableaudio2}
Audio, S.
\newblock Stable audio 2.0.
\newblock \url{https://www.stableaudio.com/user-guide/model-2}, 2024.

\bibitem[Bachmann et~al.(2024)Bachmann, Kar, Mizrahi, Garjani, Gao, Griffiths,
  Hu, Dehghan, and Zamir]{Bachmann20244M21AA}
Bachmann, R., Kar, O.~F., Mizrahi, D., Garjani, A., Gao, M., Griffiths, D., Hu,
  J., Dehghan, A., and Zamir, A.
\newblock 4m-21: An any-to-any vision model for tens of tasks and modalities.
\newblock \emph{ArXiv}, abs/2406.09406, 2024.
\newblock URL \url{https://api.semanticscholar.org/CorpusID:270520159}.

\bibitem[Bai \& An(2018)Bai and An]{bai2018survey}
Bai, S. and An, S.
\newblock A survey on automatic image caption generation.
\newblock \emph{Neurocomputing}, 311:\penalty0 291--304, 2018.

\bibitem[Bai et~al.(2023)Bai, Yang, Bai, Wang, Zhang, Lin, Wang, Zhou, and
  Zhou]{bai2023touchstone}
Bai, S., Yang, S., Bai, J., Wang, P., Zhang, X., Lin, J., Wang, X., Zhou, C.,
  and Zhou, J.
\newblock Touchstone: Evaluating vision-language models by language models.
\newblock \emph{arXiv preprint arXiv:2308.16890}, 2023.

\bibitem[Betker et~al.(2024)Betker, Goh, Jing, Brooks, Wang, Li, Ouyang,
  Zhuang, Lee, Guo, Manassra, Dhariwal, Chu, Jiao, and Ramesh]{openai_dalle3}
Betker, J., Goh, G., Jing, L., Brooks, T., Wang, J., Li, L., Ouyang, L.,
  Zhuang, J., Lee, J., Guo, Y., Manassra, W., Dhariwal, P., Chu, C., Jiao, Y.,
  and Ramesh, A.
\newblock Improving image generation with better captions.
\newblock \url{https://cdn.openai.com/papers/dall-e-3.pdf}, 2024.

\bibitem[Bitton et~al.(2023)Bitton, Bansal, Hessel, Shao, Zhu, Awadalla,
  Gardner, Taori, and Schmidt]{bitton2023visit}
Bitton, Y., Bansal, H., Hessel, J., Shao, R., Zhu, W., Awadalla, A., Gardner,
  J., Taori, R., and Schmidt, L.
\newblock Visit-bench: A benchmark for vision-language instruction following
  inspired by real-world use.
\newblock \emph{arXiv preprint arXiv:2308.06595}, 2023.

\bibitem[Blattmann et~al.(2023)Blattmann, Dockhorn, Kulal, Mendelevitch,
  Kilian, Lorenz, Levi, English, Voleti, Letts, Jampani, and
  Rombach]{blattmann2023svd}
Blattmann, A., Dockhorn, T., Kulal, S., Mendelevitch, D., Kilian, M., Lorenz,
  D., Levi, Y., English, Z., Voleti, V., Letts, A., Jampani, V., and Rombach,
  R.
\newblock Stable video diffusion: Scaling latent video diffusion models to
  large datasets.
\newblock
  \url{https://static1.squarespace.com/static/6213c340453c3f502425776e/t/655ce779b9d47d342a93c890/1733935148453/stable_video_diffusion.pdf},
  2023.

\bibitem[Brooks et~al.(2022)Brooks, Holynski, and
  Efros]{brooks2022instructpix2pix}
Brooks, T., Holynski, A., and Efros, A.
\newblock Instructpix2pix: Learning to follow image editing instructions.
  arxiv.
\newblock \emph{arXiv preprint arXiv:2211.09800}, 2022.

\bibitem[Cai et~al.(2024)Cai, Tan, Zhang, Zou, Zhang, Yao, Zhu, Gu, Zhong,
  Shang, et~al.]{cai2024temporalbench}
Cai, M., Tan, R., Zhang, J., Zou, B., Zhang, K., Yao, F., Zhu, F., Gu, J.,
  Zhong, Y., Shang, Y., et~al.
\newblock Temporalbench: Benchmarking fine-grained temporal understanding for
  multimodal video models.
\newblock \emph{arXiv preprint arXiv:2410.10818}, 2024.

\bibitem[Chen et~al.(2024{\natexlab{a}})Chen, Chen, Zhang, Liu, Wang, Zhou,
  Zhang, Wan, Zhou, and Sun]{mllm-as-a-judge}
Chen, D., Chen, R., Zhang, S., Liu, Y., Wang, Y., Zhou, H., Zhang, Q., Wan, Y.,
  Zhou, P., and Sun, L.
\newblock Mllm-as-a-judge: Assessing multimodal llm-as-a-judge with
  vision-language benchmark.
\newblock \emph{arXiv preprint arXiv:2402.04788}, 2024{\natexlab{a}}.

\bibitem[Chen et~al.(2025)Chen, Chen, Pu, Liu, Wu, Chen, Liu, Huang, Wan, Zhou,
  and Krishna]{chen2025interleaved}
Chen, D., Chen, R., Pu, S., Liu, Z., Wu, Y., Chen, C., Liu, B., Huang, Y., Wan,
  Y., Zhou, P., and Krishna, R.
\newblock Interleaved scene graph for interleaved text-and-image generation
  assessment.
\newblock In \emph{The Thirteenth International Conference on Learning
  Representations}, 2025.
\newblock URL \url{https://openreview.net/forum?id=rDLgnYLM5b}.

\bibitem[Chen et~al.(2020)Chen, Xie, Vedaldi, and Zisserman]{VGGSOUND}
Chen, H., Xie, W., Vedaldi, A., and Zisserman, A.
\newblock Vggsound: A large-scale audio-visual dataset.
\newblock In \emph{International Conference on Acoustics, Speech, and Signal
  Processing (ICASSP)}, 2020.

\bibitem[Chen et~al.(2024{\natexlab{b}})Chen, Zhang, Cun, Xia, Wang, Weng, and
  Shan]{chen2024videocrafter2}
Chen, H., Zhang, Y., Cun, X., Xia, M., Wang, X., Weng, C., and Shan, Y.
\newblock Videocrafter2: Overcoming data limitations for high-quality video
  diffusion models.
\newblock In \emph{Proceedings of the IEEE/CVF Conference on Computer Vision
  and Pattern Recognition}, pp.\  7310--7320, 2024{\natexlab{b}}.

\bibitem[Chen et~al.(2024{\natexlab{c}})Chen, Hu, Zhang, Chen, Wang, Li, Shyam,
  Zhou, Huang, Yang, et~al.]{chen2024omnixr}
Chen, L., Hu, H., Zhang, M., Chen, Y., Wang, Z., Li, Y., Shyam, P., Zhou, T.,
  Huang, H., Yang, M.-H., et~al.
\newblock Omnixr: Evaluating omni-modality language models on reasoning across
  modalities.
\newblock \emph{arXiv preprint arXiv:2410.12219}, 2024{\natexlab{c}}.

\bibitem[Chen et~al.(2024{\natexlab{d}})Chen, Lin, Zhang, and
  Huang]{chen2024autoeval}
Chen, X., Lin, Y., Zhang, Y., and Huang, W.
\newblock Autoeval-video: An automatic benchmark for assessing large vision
  language models in open-ended video question answering.
\newblock In \emph{European Conference on Computer Vision}, pp.\  179--195.
  Springer, 2024{\natexlab{d}}.

\bibitem[Chen et~al.(2024{\natexlab{e}})Chen, Lan, Zhou, Wang, and
  Pan]{chen2024sar3d}
Chen, Y., Lan, Y., Zhou, S., Wang, T., and Pan, X.
\newblock Sar3d: Autoregressive 3d object generation and understanding via
  multi-scale 3d vqvae.
\newblock \emph{arXiv preprint arXiv:2411.16856}, 2024{\natexlab{e}}.

\bibitem[Chen et~al.(2024{\natexlab{f}})Chen, Yue, Zhang, Gao, Tan, and
  Li]{chen2024voicebench}
Chen, Y., Yue, X., Zhang, C., Gao, X., Tan, R.~T., and Li, H.
\newblock Voicebench: Benchmarking llm-based voice assistants.
\newblock \emph{arXiv preprint arXiv:2410.17196}, 2024{\natexlab{f}}.

\bibitem[Chen et~al.(2024{\natexlab{g}})Chen, Du, Wen, Zhou, Cui, Weng, Tu,
  Wang, Tong, Huang, et~al.]{chen2024mj}
Chen, Z., Du, Y., Wen, Z., Zhou, Y., Cui, C., Weng, Z., Tu, H., Wang, C., Tong,
  Z., Huang, Q., et~al.
\newblock Mj-bench: Is your multimodal reward model really a good judge for
  text-to-image generation?
\newblock \emph{arXiv preprint arXiv:2407.04842}, 2024{\natexlab{g}}.

\bibitem[Cheng et~al.(2024)Cheng, Leng, Zhang, Xin, Li, Chen, Zhu, Zhang, Luo,
  Zhao, et~al.]{cheng2024videollama2}
Cheng, Z., Leng, S., Zhang, H., Xin, Y., Li, X., Chen, G., Zhu, Y., Zhang, W.,
  Luo, Z., Zhao, D., et~al.
\newblock Videollama 2: Advancing spatial-temporal modeling and audio
  understanding in video-llms.
\newblock \emph{arXiv preprint arXiv:2406.07476}, 2024.

\bibitem[Chern et~al.(2024)Chern, Su, Ma, and Liu]{chern2024anole}
Chern, E., Su, J., Ma, Y., and Liu, P.
\newblock Anole: An open, autoregressive, native large multimodal models for
  interleaved image-text generation.
\newblock \emph{arXiv preprint arXiv:2407.06135}, 2024.

\bibitem[Chu et~al.(2023)Chu, Xu, Zhou, Yang, Zhang, Yan, Zhou, and
  Zhou]{chu2023qwen}
Chu, Y., Xu, J., Zhou, X., Yang, Q., Zhang, S., Yan, Z., Zhou, C., and Zhou, J.
\newblock Qwen-audio: Advancing universal audio understanding via unified
  large-scale audio-language models.
\newblock \emph{arXiv preprint arXiv:2311.07919}, 2023.

\bibitem[Chu et~al.(2024)Chu, Xu, Yang, Wei, Wei, Guo, Leng, Lv, He, Lin,
  et~al.]{chu2024qwen2}
Chu, Y., Xu, J., Yang, Q., Wei, H., Wei, X., Guo, Z., Leng, Y., Lv, Y., He, J.,
  Lin, J., et~al.
\newblock Qwen2-audio technical report.
\newblock \emph{arXiv preprint arXiv:2407.10759}, 2024.

\bibitem[Doh et~al.(2023)Doh, Choi, Lee, and Nam]{seung2023musiccaps}
Doh, S., Choi, K., Lee, J., and Nam, J.
\newblock Lp-musiccaps: Llm-based pseudo music captioning.
\newblock In \emph{ISMIR}, pp.\  409--416, 2023.
\newblock URL \url{https://doi.org/10.5281/zenodo.10265311}.

\bibitem[Drossos et~al.(2020)Drossos, Lipping, and Virtanen]{drossos2020clotho}
Drossos, K., Lipping, S., and Virtanen, T.
\newblock Clotho: An audio captioning dataset.
\newblock In \emph{ICASSP 2020-2020 IEEE International Conference on Acoustics,
  Speech and Signal Processing (ICASSP)}, pp.\  736--740. IEEE, 2020.

\bibitem[Dubey et~al.(2024)Dubey, Jauhri, Pandey, Kadian, Al-Dahle, Letman,
  Mathur, Schelten, Yang, Fan, et~al.]{dubey2024llama}
Dubey, A., Jauhri, A., Pandey, A., Kadian, A., Al-Dahle, A., Letman, A.,
  Mathur, A., Schelten, A., Yang, A., Fan, A., et~al.
\newblock The llama 3 herd of models.
\newblock \emph{arXiv preprint arXiv:2407.21783}, 2024.

\bibitem[Elo(1966)]{elo}
Elo, A.
\newblock \emph{The USCF Rating System: Its Development, Theory, and
  Applications}.
\newblock United States Chess Federation, 1966.
\newblock URL \url{https://books.google.com/books?id=onUazQEACAAJ}.

\bibitem[Esser et~al.(2024)Esser, Kulal, Blattmann, Entezari, M{\"u}ller,
  Saini, Levi, Lorenz, Sauer, Boesel, et~al.]{esser2024scaling}
Esser, P., Kulal, S., Blattmann, A., Entezari, R., M{\"u}ller, J., Saini, H.,
  Levi, Y., Lorenz, D., Sauer, A., Boesel, F., et~al.
\newblock Scaling rectified flow transformers for high-resolution image
  synthesis.
\newblock In \emph{Forty-first international conference on machine learning},
  2024.

\bibitem[Evans et~al.(2024)Evans, Parker, Carr, Zukowski, Taylor, and
  Pons]{evans2024stableaudio}
Evans, Z., Parker, J.~D., Carr, C., Zukowski, Z., Taylor, J., and Pons, J.
\newblock Stable audio open.
\newblock \emph{arXiv preprint arXiv:2407.14358}, 2024.

\bibitem[Fan et~al.(2023)Fan, Luo, Gao, and Zhan]{fan2023aigcbench}
Fan, F., Luo, C., Gao, W., and Zhan, J.
\newblock Aigcbench: Comprehensive evaluation of image-to-video content
  generated by ai.
\newblock \emph{BenchCouncil Transactions on Benchmarks, Standards and
  Evaluations}, 3\penalty0 (4):\penalty0 100152, 2023.

\bibitem[Fu et~al.(2024)Fu, Dai, Luo, Li, Ren, Zhang, Wang, Zhou, Shen, Zhang,
  et~al.]{fu2024video}
Fu, C., Dai, Y., Luo, Y., Li, L., Ren, S., Zhang, R., Wang, Z., Zhou, C., Shen,
  Y., Zhang, M., et~al.
\newblock Video-mme: The first-ever comprehensive evaluation benchmark of
  multi-modal llms in video analysis.
\newblock \emph{arXiv preprint arXiv:2405.21075}, 2024.

\bibitem[Fu et~al.(2023)Fu, Hu, Du, Wang, Yang, and Gan]{fu2023mgie}
Fu, T.-J., Hu, W., Du, X., Wang, W.~Y., Yang, Y., and Gan, Z.
\newblock Guiding instruction-based image editing via multimodal large language
  models.
\newblock \emph{arXiv preprint arXiv:2309.17102}, 2023.

\bibitem[Ghosh et~al.(2023)Ghosh, Hajishirzi, and Schmidt]{ghosh2023geneval}
Ghosh, D., Hajishirzi, H., and Schmidt, L.
\newblock Geneval: An object-focused framework for evaluating text-to-image
  alignment.
\newblock \emph{Advances in Neural Information Processing Systems},
  36:\penalty0 52132--52152, 2023.

\bibitem[Ghosh et~al.(2024)Ghosh, Kumar, Seth, Evuru, Tyagi, Sakshi, Nieto,
  Duraiswami, and Manocha]{ghosh2024gama}
Ghosh, S., Kumar, S., Seth, A., Evuru, C. K.~R., Tyagi, U., Sakshi, S., Nieto,
  O., Duraiswami, R., and Manocha, D.
\newblock Gama: A large audio-language model with advanced audio understanding
  and complex reasoning abilities.
\newblock \emph{arXiv preprint arXiv:2406.11768}, 2024.

\bibitem[Gong et~al.(2023)Gong, Luo, Liu, Karlinsky, and Glass]{gong2023listen}
Gong, Y., Luo, H., Liu, A.~H., Karlinsky, L., and Glass, J.
\newblock Listen, think, and understand.
\newblock \emph{arXiv preprint arXiv:2305.10790}, 2023.

\bibitem[Gu et~al.(2024)Gu, Jiang, Shi, Tan, Zhai, Xu, Li, Shen, Ma, Liu,
  et~al.]{gu2024LLMJudgesurvey}
Gu, J., Jiang, X., Shi, Z., Tan, H., Zhai, X., Xu, C., Li, W., Shen, Y., Ma,
  S., Liu, H., et~al.
\newblock A survey on llm-as-a-judge.
\newblock \emph{arXiv preprint arXiv:2411.15594}, 2024.

\bibitem[Han et~al.(2024)Han, Gong, Zhang, Wang, Zhang, Lin, Qiao, Gao, and
  Yue]{onellm}
Han, J., Gong, K., Zhang, Y., Wang, J., Zhang, K., Lin, D., Qiao, Y., Gao, P.,
  and Yue, X.
\newblock Onellm: One framework to align all modalities with language.
\newblock In \emph{Proceedings of the IEEE/CVF Conference on Computer Vision
  and Pattern Recognition}, pp.\  26584--26595, 2024.

\bibitem[Hong et~al.(2025)Hong, Yan, Cai, Jiang, Hu, and
  Xie]{hong2025worldsense}
Hong, J., Yan, S., Cai, J., Jiang, X., Hu, Y., and Xie, W.
\newblock Worldsense: Evaluating real-world omnimodal understanding for
  multimodal llms.
\newblock \emph{arXiv preprint arXiv:2502.04326}, 2025.

\bibitem[Hu et~al.(2025)Hu, Wu, Pu, Xiao, Zhang, Yue, Li, and Liu]{hu2025video}
Hu, K., Wu, P., Pu, F., Xiao, W., Zhang, Y., Yue, X., Li, B., and Liu, Z.
\newblock Video-mmmu: Evaluating knowledge acquisition from multi-discipline
  professional videos.
\newblock \emph{arXiv preprint arXiv:2501.13826}, 2025.

\bibitem[Huang et~al.(2023)Huang, Sun, Xie, Li, and Liu]{huang2023t2i}
Huang, K., Sun, K., Xie, E., Li, Z., and Liu, X.
\newblock T2i-compbench: A comprehensive benchmark for open-world compositional
  text-to-image generation.
\newblock \emph{Advances in Neural Information Processing Systems},
  36:\penalty0 78723--78747, 2023.

\bibitem[Huang et~al.(2024)Huang, He, Yu, Zhang, Si, Jiang, Zhang, Wu, Jin,
  Chanpaisit, et~al.]{huang2024vbench}
Huang, Z., He, Y., Yu, J., Zhang, F., Si, C., Jiang, Y., Zhang, Y., Wu, T.,
  Jin, Q., Chanpaisit, N., et~al.
\newblock Vbench: Comprehensive benchmark suite for video generative models.
\newblock In \emph{Proceedings of the IEEE/CVF Conference on Computer Vision
  and Pattern Recognition}, pp.\  21807--21818, 2024.

\bibitem[Hurst et~al.(2024)Hurst, Lerer, Goucher, Perelman, Ramesh, Clark,
  Ostrow, Welihinda, Hayes, Radford, et~al.]{2024gpt}
Hurst, A., Lerer, A., Goucher, A.~P., Perelman, A., Ramesh, A., Clark, A.,
  Ostrow, A., Welihinda, A., Hayes, A., Radford, A., et~al.
\newblock Gpt-4o system card.
\newblock \emph{arXiv preprint arXiv:2410.21276}, 2024.

\bibitem[Iashin \& Rahtu(2021)Iashin and Rahtu]{iashin2021specvqgan}
Iashin, V. and Rahtu, E.
\newblock Taming visually guided sound generation.
\newblock \emph{arXiv preprint arXiv:2110.08791}, 2021.

\bibitem[Ji et~al.(2024)Ji, Zhou, Lou, Chen, Hong, Wang, Chen, Wang, Pan, Li,
  et~al.]{AlignAnything}
Ji, J., Zhou, J., Lou, H., Chen, B., Hong, D., Wang, X., Chen, W., Wang, K.,
  Pan, R., Li, J., et~al.
\newblock Align anything: Training all-modality models to follow instructions
  with language feedback.
\newblock \emph{arXiv preprint arXiv:2412.15838}, 2024.

\bibitem[Jia et~al.(2024)Jia, Chen, Zhao, Zhao, Zeng, Chen, and
  Qin]{jia2024audioeditor}
Jia, Y., Chen, Y., Zhao, J., Zhao, S., Zeng, W., Chen, Y., and Qin, Y.
\newblock Audioeditor: A training-free diffusion-based audio editing framework.
\newblock \emph{arXiv preprint arXiv:2409.12466}, 2024.

\bibitem[Jiang et~al.(2025)Jiang, Ku, Li, Ni, Sun, Fan, and
  Chen]{jiang2025genai}
Jiang, D., Ku, M., Li, T., Ni, Y., Sun, S., Fan, R., and Chen, W.
\newblock Genai arena: An open evaluation platform for generative models.
\newblock \emph{Advances in Neural Information Processing Systems},
  37:\penalty0 79889--79908, 2025.

\bibitem[Kang et~al.(2024)Kang, Poria, and Herremans]{kang2024video2music}
Kang, J., Poria, S., and Herremans, D.
\newblock Video2music: Suitable music generation from videos using an affective
  multimodal transformer model.
\newblock \emph{Expert Systems with Applications}, 249:\penalty0 123640, 2024.

\bibitem[Khattak et~al.(2024)Khattak, Naeem, Hassan, Naseer, Tombari, Khan, and
  Khan]{khattak2024good}
Khattak, M.~U., Naeem, M.~F., Hassan, J., Naseer, M., Tombari, F., Khan, F.~S.,
  and Khan, S.
\newblock How good is my video lmm? complex video reasoning and robustness
  evaluation suite for video-lmms.
\newblock \emph{arXiv preprint arXiv:2405.03690}, 2024.

\bibitem[Kim et~al.(2019)Kim, Kim, Lee, and Kim]{kim-etal-2019-audiocaps}
Kim, C.~D., Kim, B., Lee, H., and Kim, G.
\newblock {A}udio{C}aps: Generating captions for audios in the wild.
\newblock In Burstein, J., Doran, C., and Solorio, T. (eds.), \emph{Proceedings
  of the 2019 Conference of the North {A}merican Chapter of the Association for
  Computational Linguistics: Human Language Technologies, Volume 1 (Long and
  Short Papers)}, pp.\  119--132, Minneapolis, Minnesota, June 2019.
  Association for Computational Linguistics.
\newblock \doi{10.18653/v1/N19-1011}.
\newblock URL \url{https://aclanthology.org/N19-1011/}.

\bibitem[Kou et~al.(2024)Kou, Jin, Liu, Ma, Jia, Chen, Jiang, and
  Deng]{kou2024orthus}
Kou, S., Jin, J., Liu, C., Ma, Y., Jia, J., Chen, Q., Jiang, P., and Deng, Z.
\newblock Orthus: Autoregressive interleaved image-text generation with
  modality-specific heads.
\newblock \emph{arXiv preprint arXiv:2412.00127}, 2024.

\bibitem[Krishna et~al.(2017)Krishna, Zhu, Groth, Johnson, Hata, Kravitz, Chen,
  Kalantidis, Li, Shamma, et~al.]{krishna2017visual}
Krishna, R., Zhu, Y., Groth, O., Johnson, J., Hata, K., Kravitz, J., Chen, S.,
  Kalantidis, Y., Li, L.-J., Shamma, D.~A., et~al.
\newblock Visual genome: Connecting language and vision using crowdsourced
  dense image annotations.
\newblock \emph{International journal of computer vision}, 123:\penalty0
  32--73, 2017.

\bibitem[Labs(2024)]{blackforestlabs_flux}
Labs, B.~F.
\newblock Flux: A framework for state-of-the-art image generation.
\newblock \url{https://github.com/black-forest-labs/flux}, 2024.

\bibitem[Lai et~al.(2024)Lai, Zhang, Liu, Li, Lu, and Guo]{lai2024spider}
Lai, J., Zhang, J., Liu, J., Li, J., Lu, X., and Guo, S.
\newblock Spider: Any-to-many multimodal llm.
\newblock \emph{arXiv preprint arXiv:2411.09439}, 2024.

\bibitem[Lee et~al.(2023{\natexlab{a}})Lee, Kim, and Park]{Codi}
Lee, C., Kim, J., and Park, N.
\newblock Codi: Co-evolving contrastive diffusion models for mixed-type tabular
  synthesis.
\newblock In \emph{International Conference on Machine Learning}, pp.\
  18940--18956. PMLR, 2023{\natexlab{a}}.

\bibitem[Lee et~al.(2023{\natexlab{b}})Lee, Phatale, Mansoor, Mesnard, Ferret,
  Lu, Bishop, Hall, Carbune, Rastogi, et~al.]{lee2023rlaif}
Lee, H., Phatale, S., Mansoor, H., Mesnard, T., Ferret, J., Lu, K., Bishop, C.,
  Hall, E., Carbune, V., Rastogi, A., et~al.
\newblock Rlaif vs. rlhf: Scaling reinforcement learning from human feedback
  with ai feedback.
\newblock \emph{arXiv preprint arXiv:2309.00267}, 2023{\natexlab{b}}.

\bibitem[Lee et~al.(2024)Lee, Kim, Park, Kim, and Seo]{lee2024prometheus}
Lee, S., Kim, S., Park, S., Kim, G., and Seo, M.
\newblock Prometheus-vision: Vision-language model as a judge for fine-grained
  evaluation.
\newblock In \emph{Findings of the Association for Computational Linguistics
  ACL 2024}, pp.\  11286--11315, 2024.

\bibitem[Lee~Rodgers \& Nicewander(1988)Lee~Rodgers and
  Nicewander]{lee1988thirteen}
Lee~Rodgers, J. and Nicewander, W.~A.
\newblock Thirteen ways to look at the correlation coefficient.
\newblock \emph{The American Statistician}, 42\penalty0 (1):\penalty0 59--66,
  1988.

\bibitem[Li et~al.(2024{\natexlab{a}})Li, Jiang, Huang, Beigi, Zhao, Tan,
  Bhattacharjee, Jiang, Chen, Wu, et~al.]{li2024generation2judgement}
Li, D., Jiang, B., Huang, L., Beigi, A., Zhao, C., Tan, Z., Bhattacharjee, A.,
  Jiang, Y., Chen, C., Wu, T., et~al.
\newblock From generation to judgment: Opportunities and challenges of
  llm-as-a-judge.
\newblock \emph{arXiv preprint arXiv:2411.16594}, 2024{\natexlab{a}}.

\bibitem[Li et~al.(2024{\natexlab{b}})Li, Liu, Wu, Wang, Shen, Qu, Niu, Wang,
  Chen, and Li]{li2024aria}
Li, D., Liu, Y., Wu, H., Wang, Y., Shen, Z., Qu, B., Niu, X., Wang, G., Chen,
  B., and Li, J.
\newblock Aria: An open multimodal native mixture-of-experts model.
\newblock \emph{arXiv preprint arXiv:2410.05993}, 2024{\natexlab{b}}.

\bibitem[Li et~al.(2023{\natexlab{a}})Li, Zhang, Koto, Yang, Zhao, Gong, Duan,
  and Baldwin]{li2023cmmlu}
Li, H., Zhang, Y., Koto, F., Yang, Y., Zhao, H., Gong, Y., Duan, N., and
  Baldwin, T.
\newblock Cmmlu: Measuring massive multitask language understanding in chinese.
\newblock \emph{arXiv preprint arXiv:2306.09212}, 2023{\natexlab{a}}.

\bibitem[Li et~al.(2024{\natexlab{c}})Li, Tian, Shao, Zhu, Wang, Zhu, Dou,
  Wang, Li, Lu, et~al.]{li2024synergen}
Li, H., Tian, C., Shao, J., Zhu, X., Wang, Z., Zhu, J., Dou, W., Wang, X., Li,
  H., Lu, L., et~al.
\newblock Synergen-vl: Towards synergistic image understanding and generation
  with vision experts and token folding.
\newblock \emph{arXiv preprint arXiv:2412.09604}, 2024{\natexlab{c}}.

\bibitem[Li et~al.(2023{\natexlab{b}})Li, Li, Savarese, and Hoi]{li2023blip}
Li, J., Li, D., Savarese, S., and Hoi, S.
\newblock Blip-2: Bootstrapping language-image pre-training with frozen image
  encoders and large language models.
\newblock In \emph{International conference on machine learning}, pp.\
  19730--19742. PMLR, 2023{\natexlab{b}}.

\bibitem[Li et~al.(2023{\natexlab{c}})Li, Pan, Ge, Gao, Ji, Zhang, Chua, Tang,
  Zhang, and Zhuang]{li2023fine}
Li, J., Pan, K., Ge, Z., Gao, M., Ji, W., Zhang, W., Chua, T.-S., Tang, S.,
  Zhang, H., and Zhuang, Y.
\newblock Fine-tuning multimodal llms to follow zero-shot demonstrative
  instructions.
\newblock \emph{arXiv preprint arXiv:2308.04152}, 2023{\natexlab{c}}.

\bibitem[Li et~al.(2024{\natexlab{d}})Li, Wei, Xie, Yang, Song, Wang, An, Liu,
  Li, Lin, et~al.]{li2024vlrewardbench}
Li, L., Wei, Y., Xie, Z., Yang, X., Song, Y., Wang, P., An, C., Liu, T., Li,
  S., Lin, B.~Y., et~al.
\newblock Vlrewardbench: A challenging benchmark for vision-language generative
  reward models.
\newblock \emph{arXiv preprint arXiv:2411.17451}, 2024{\natexlab{d}}.

\bibitem[Li et~al.(2023{\natexlab{d}})Li, Singh, and
  Grover]{li2023instructany2pix}
Li, S., Singh, H., and Grover, A.
\newblock Instructany2pix: Flexible visual editing via multimodal instruction
  following.
\newblock \emph{arXiv preprint arXiv:2312.06738}, 2023{\natexlab{d}}.

\bibitem[Li et~al.(2024{\natexlab{e}})Li, Kallidromitis, Gokul, Liao, Kato,
  Kozuka, and Grover]{li2024omniflow}
Li, S., Kallidromitis, K., Gokul, A., Liao, Z., Kato, Y., Kozuka, K., and
  Grover, A.
\newblock Omniflow: Any-to-any generation with multi-modal rectified flows.
\newblock \emph{arXiv preprint arXiv:2412.01169}, 2024{\natexlab{e}}.

\bibitem[Li et~al.(2024{\natexlab{f}})Li, Ma, Yang, and Yang]{li2024vidtome}
Li, X., Ma, C., Yang, X., and Yang, M.-H.
\newblock Vidtome: Video token merging for zero-shot video editing.
\newblock In \emph{Proceedings of the IEEE/CVF Conference on Computer Vision
  and Pattern Recognition}, pp.\  7486--7495, 2024{\natexlab{f}}.

\bibitem[Li et~al.(2024{\natexlab{g}})Li, Zhang, Ma, Yuan, Zhu, Guo, Liang,
  Liu, Wang, Yang, et~al.]{li2024omnibench}
Li, Y., Zhang, G., Ma, Y., Yuan, R., Zhu, K., Guo, H., Liang, Y., Liu, J.,
  Wang, Z., Yang, J., et~al.
\newblock Omnibench: Towards the future of universal omni-language models.
\newblock \emph{arXiv preprint arXiv:2409.15272}, 2024{\natexlab{g}}.

\bibitem[Li et~al.(2024{\natexlab{h}})Li, Zhang, Wang, Zhong, Chen, Chu, Liu,
  and Jia]{li2024mini}
Li, Y., Zhang, Y., Wang, C., Zhong, Z., Chen, Y., Chu, R., Liu, S., and Jia, J.
\newblock Mini-gemini: Mining the potential of multi-modality vision language
  models.
\newblock \emph{arXiv preprint arXiv:2403.18814}, 2024{\natexlab{h}}.

\bibitem[Li et~al.(2025)Li, Liu, Zhang, Chen, Li, Li, Liu, Ming, Dong, Pan,
  et~al.]{li2025baichuan}
Li, Y., Liu, J., Zhang, T., Chen, S., Li, T., Li, Z., Liu, L., Ming, L., Dong,
  G., Pan, D., et~al.
\newblock Baichuan-omni-1.5 technical report.
\newblock \emph{arXiv preprint arXiv:2501.15368}, 2025.

\bibitem[Li et~al.(2024{\natexlab{i}})Li, Li, Shi, Farimani, Kluger, Yang, and
  Wang]{li2024dual}
Li, Z., Li, H., Shi, Y., Farimani, A.~B., Kluger, Y., Yang, L., and Wang, P.
\newblock Dual diffusion for unified image generation and understanding.
\newblock \emph{arXiv preprint arXiv:2501.00289}, 2024{\natexlab{i}}.

\bibitem[Liang et~al.(2024)Liang, He, Li, Li, Klimovskiy, Carolan, Sun,
  Pont-Tuset, Young, Yang, et~al.]{liang2024rich}
Liang, Y., He, J., Li, G., Li, P., Klimovskiy, A., Carolan, N., Sun, J.,
  Pont-Tuset, J., Young, S., Yang, F., et~al.
\newblock Rich human feedback for text-to-image generation.
\newblock In \emph{Proceedings of the IEEE/CVF Conference on Computer Vision
  and Pattern Recognition}, pp.\  19401--19411, 2024.

\bibitem[Lin et~al.(2023)Lin, Ye, Zhu, Cui, Ning, Jin, and Yuan]{lin2023video}
Lin, B., Ye, Y., Zhu, B., Cui, J., Ning, M., Jin, P., and Yuan, L.
\newblock Video-llava: Learning united visual representation by alignment
  before projection.
\newblock \emph{arXiv preprint arXiv:2311.10122}, 2023.

\bibitem[Lin et~al.(2024{\natexlab{a}})Lin, Deng, Chandu, Brahman, Ravichander,
  Pyatkin, Dziri, Bras, and Choi]{lin2024wildbench}
Lin, B.~Y., Deng, Y., Chandu, K., Brahman, F., Ravichander, A., Pyatkin, V.,
  Dziri, N., Bras, R.~L., and Choi, Y.
\newblock Wildbench: Benchmarking llms with challenging tasks from real users
  in the wild.
\newblock \emph{arXiv preprint arXiv:2406.04770}, 2024{\natexlab{a}}.

\bibitem[Lin et~al.(2024{\natexlab{b}})Lin, Pathak, Li, Li, Xia, Neubig, Zhang,
  and Ramanan]{lin2024evaluating}
Lin, Z., Pathak, D., Li, B., Li, J., Xia, X., Neubig, G., Zhang, P., and
  Ramanan, D.
\newblock Evaluating text-to-visual generation with image-to-text generation.
\newblock In \emph{European Conference on Computer Vision}, pp.\  366--384.
  Springer, 2024{\natexlab{b}}.

\bibitem[Liu et~al.(2024{\natexlab{a}})Liu, Feng, Xue, Wang, Wu, Lu, Zhao,
  Deng, Zhang, Ruan, et~al.]{liu2024deepseek}
Liu, A., Feng, B., Xue, B., Wang, B., Wu, B., Lu, C., Zhao, C., Deng, C.,
  Zhang, C., Ruan, C., et~al.
\newblock Deepseek-v3 technical report.
\newblock \emph{arXiv preprint arXiv:2412.19437}, 2024{\natexlab{a}}.

\bibitem[Liu et~al.(2023{\natexlab{a}})Liu, Li, Wu, and Lee]{liu2023visual}
Liu, H., Li, C., Wu, Q., and Lee, Y.~J.
\newblock Visual instruction tuning.
\newblock \emph{Advances in neural information processing systems},
  36:\penalty0 34892--34916, 2023{\natexlab{a}}.

\bibitem[Liu et~al.(2024{\natexlab{b}})Liu, Yuan, Liu, Mei, Kong, Tian, Wang,
  Wang, Wang, and Plumbley]{liu2024audioldm2}
Liu, H., Yuan, Y., Liu, X., Mei, X., Kong, Q., Tian, Q., Wang, Y., Wang, W.,
  Wang, Y., and Plumbley, M.~D.
\newblock Audioldm 2: Learning holistic audio generation with self-supervised
  pretraining.
\newblock \emph{IEEE/ACM Transactions on Audio, Speech, and Language
  Processing}, 2024{\natexlab{b}}.

\bibitem[Liu et~al.(2023{\natexlab{b}})Liu, Iter, Xu, Wang, Xu, and
  Zhu]{liu2023g}
Liu, Y., Iter, D., Xu, Y., Wang, S., Xu, R., and Zhu, C.
\newblock G-eval: Nlg evaluation using gpt-4 with better human alignment.
\newblock \emph{arXiv preprint arXiv:2303.16634}, 2023{\natexlab{b}}.

\bibitem[Lu et~al.(2022)Lu, Clark, Zellers, Mottaghi, and
  Kembhavi]{lu2022unified}
Lu, J., Clark, C., Zellers, R., Mottaghi, R., and Kembhavi, A.
\newblock Unified-io: A unified model for vision, language, and multi-modal
  tasks.
\newblock \emph{arXiv preprint arXiv:2206.08916}, 2022.

\bibitem[Lu et~al.(2024)Lu, Clark, Lee, Zhang, Khosla, Marten, Hoiem, and
  Kembhavi]{unifiedIO2}
Lu, J., Clark, C., Lee, S., Zhang, Z., Khosla, S., Marten, R., Hoiem, D., and
  Kembhavi, A.
\newblock Unified-io 2: Scaling autoregressive multimodal models with vision
  language audio and action.
\newblock In \emph{Proceedings of the IEEE/CVF Conference on Computer Vision
  and Pattern Recognition}, pp.\  26439--26455, 2024.

\bibitem[Luo et~al.(2023)Luo, Yan, Hu, and Zhao]{luo2023diff_foley}
Luo, S., Yan, C., Hu, C., and Zhao, H.
\newblock Diff-foley: Synchronized video-to-audio synthesis with latent
  diffusion models.
\newblock \emph{Advances in Neural Information Processing Systems},
  36:\penalty0 48855--48876, 2023.

\bibitem[Luo et~al.(2024)Luo, Wu, Li, Ma, Kankanhalli, and
  Li]{luo2024videoautoarena}
Luo, Z., Wu, H., Li, D., Ma, J., Kankanhalli, M., and Li, J.
\newblock Videoautoarena: An automated arena for evaluating large multimodal
  models in video analysis through user simulation.
\newblock \emph{arXiv preprint arXiv:2411.13281}, 2024.

\bibitem[Ma et~al.(2024{\natexlab{a}})Ma, Ji, Ye, Lin, Wang, Zheng, Zhou, Sun,
  and Ji]{ma2024i2ebench}
Ma, Y., Ji, J., Ye, K., Lin, W., Wang, Z., Zheng, Y., Zhou, Q., Sun, X., and
  Ji, R.
\newblock I2ebench: A comprehensive benchmark for instruction-based image
  editing.
\newblock \emph{arXiv preprint arXiv:2408.14180}, 2024{\natexlab{a}}.

\bibitem[Ma et~al.(2024{\natexlab{b}})Ma, Liu, Chen, Liu, Wu, Wu, Pan, Xie,
  Zhang, Zhao, et~al.]{ma2024janusflow}
Ma, Y., Liu, X., Chen, X., Liu, W., Wu, C., Wu, Z., Pan, Z., Xie, Z., Zhang,
  H., Zhao, L., et~al.
\newblock Janusflow: Harmonizing autoregression and rectified flow for unified
  multimodal understanding and generation.
\newblock \emph{arXiv preprint arXiv:2411.07975}, 2024{\natexlab{b}}.

\bibitem[Majumder et~al.(2024)Majumder, Hung, Ghosal, Hsu, Mihalcea, and
  Poria]{majumder2024tango2}
Majumder, N., Hung, C.-Y., Ghosal, D., Hsu, W.-N., Mihalcea, R., and Poria, S.
\newblock Tango 2: Aligning diffusion-based text-to-audio generations through
  direct preference optimization.
\newblock In \emph{Proceedings of the 32nd ACM International Conference on
  Multimedia}, pp.\  564--572, 2024.

\bibitem[Meng et~al.(2021)Meng, He, Song, Song, Wu, Zhu, and
  Ermon]{meng2021sdedit}
Meng, C., He, Y., Song, Y., Song, J., Wu, J., Zhu, J.-Y., and Ermon, S.
\newblock Sdedit: Guided image synthesis and editing with stochastic
  differential equations.
\newblock \emph{arXiv preprint arXiv:2108.01073}, 2021.

\bibitem[Minimax(2024)]{minimax2024minimaxvideo01}
Minimax.
\newblock Minimax video-01.
\newblock \url{https://www.minimax.io/news/video-01?utm_source=minimaxi}, 2024.

\bibitem[Mizrahi et~al.(2023)Mizrahi, Bachmann, Kar, Yeo, Gao, Dehghan, and
  Zamir]{Mizrahi20234MMM}
Mizrahi, D., Bachmann, R., Kar, O.~F., Yeo, T., Gao, M., Dehghan, A., and
  Zamir, A.
\newblock 4m: Massively multimodal masked modeling.
\newblock \emph{ArXiv}, abs/2312.06647, 2023.
\newblock URL \url{https://api.semanticscholar.org/CorpusID:266162752}.

\bibitem[ML(2024)]{runwayml2024gen3}
ML, R.
\newblock Introducing gen-3 alpha: A new frontier for video generation.
\newblock \url{https://runwayml.com/research/introducing-gen-3-alpha}, 2024.

\bibitem[Ni et~al.(2024)Ni, Song, Ghosal, Li, Zhang, Yue, Xue, Zheng, Zhang,
  Shah, et~al.]{ni2024mixeval}
Ni, J., Song, Y., Ghosal, D., Li, B., Zhang, D.~J., Yue, X., Xue, F., Zheng,
  Z., Zhang, K., Shah, M., et~al.
\newblock Mixeval-x: Any-to-any evaluations from real-world data mixtures.
\newblock \emph{arXiv preprint arXiv:2410.13754}, 2024.

\bibitem[OpenAI(2024)]{openai_sora}
OpenAI.
\newblock Sora.
\newblock \url{https://openai.com/sora/}, 2024.

\bibitem[Qin et~al.(2023)Qin, Yu, Xing, Zhang, Chen, Ermon, Fu, Xiong, and
  Xu]{qin2023gluegen}
Qin, C., Yu, N., Xing, C., Zhang, S., Chen, Z., Ermon, S., Fu, Y., Xiong, C.,
  and Xu, R.
\newblock Gluegen: Plug and play multi-modal encoders for x-to-image
  generation.
\newblock In \emph{Proceedings of the IEEE/CVF international conference on
  computer vision}, pp.\  23085--23096, 2023.

\bibitem[Qu et~al.(2024)Qu, Zhang, Liu, Wang, Jiang, Gao, Ye, Du, Yuan, and
  Wu]{qu2024tokenflow}
Qu, L., Zhang, H., Liu, Y., Wang, X., Jiang, Y., Gao, Y., Ye, H., Du, D.~K.,
  Yuan, Z., and Wu, X.
\newblock Tokenflow: Unified image tokenizer for multimodal understanding and
  generation.
\newblock \emph{arXiv preprint arXiv:2412.03069}, 2024.

\bibitem[Razavi et~al.(2019)Razavi, Van~den Oord, and
  Vinyals]{razavi2019generating}
Razavi, A., Van~den Oord, A., and Vinyals, O.
\newblock Generating diverse high-fidelity images with vq-vae-2.
\newblock \emph{Advances in neural information processing systems}, 32, 2019.

\bibitem[Ren et~al.(2024)Ren, Yang, Zhang, Wei, Du, Huang, and
  Chen]{ren2024consisti2v}
Ren, W., Yang, H., Zhang, G., Wei, C., Du, X., Huang, W., and Chen, W.
\newblock Consisti2v: Enhancing visual consistency for image-to-video
  generation.
\newblock \emph{arXiv preprint arXiv:2402.04324}, 2024.

\bibitem[Rombach et~al.(2022)Rombach, Blattmann, Lorenz, Esser, and
  Ommer]{rombach2022high}
Rombach, R., Blattmann, A., Lorenz, D., Esser, P., and Ommer, B.
\newblock High-resolution image synthesis with latent diffusion models.
\newblock In \emph{Proceedings of the IEEE/CVF conference on computer vision
  and pattern recognition}, pp.\  10684--10695, 2022.

\bibitem[Ruan et~al.(2023)Ruan, Ma, Yang, He, Liu, Fu, Yuan, Jin, and
  Guo]{ruan2023mm_difusion}
Ruan, L., Ma, Y., Yang, H., He, H., Liu, B., Fu, J., Yuan, N.~J., Jin, Q., and
  Guo, B.
\newblock Mm-diffusion: Learning multi-modal diffusion models for joint audio
  and video generation.
\newblock In \emph{Proceedings of the IEEE/CVF Conference on Computer Vision
  and Pattern Recognition}, pp.\  10219--10228, 2023.

\bibitem[Sakshi et~al.(2024)Sakshi, Tyagi, Kumar, Seth, Selvakumar, Nieto,
  Duraiswami, Ghosh, and Manocha]{sakshi2024mmau}
Sakshi, S., Tyagi, U., Kumar, S., Seth, A., Selvakumar, R., Nieto, O.,
  Duraiswami, R., Ghosh, S., and Manocha, D.
\newblock Mmau: A massive multi-task audio understanding and reasoning
  benchmark.
\newblock \emph{arXiv preprint arXiv:2410.19168}, 2024.

\bibitem[Sheffer \& Adi(2023)Sheffer and Adi]{sheffer2023im2wav}
Sheffer, R. and Adi, Y.
\newblock I hear your true colors: Image guided audio generation.
\newblock In \emph{ICASSP 2023-2023 IEEE International Conference on Acoustics,
  Speech and Signal Processing (ICASSP)}, pp.\  1--5. IEEE, 2023.

\bibitem[Shi et~al.(2024)Shi, Han, Zhou, Liang, Lin, Zettlemoyer, and
  Yu]{shi2024llamafusion}
Shi, W., Han, X., Zhou, C., Liang, W., Lin, X.~V., Zettlemoyer, L., and Yu, L.
\newblock Llamafusion: Adapting pretrained language models for multimodal
  generation.
\newblock \emph{arXiv preprint arXiv:2412.15188}, 2024.

\bibitem[Sun et~al.(2024{\natexlab{a}})Sun, Yu, Tang, Chen, Tan, Li, Lu, Ma,
  Wang, and Zhang]{sun2024video_salmonn}
Sun, G., Yu, W., Tang, C., Chen, X., Tan, T., Li, W., Lu, L., Ma, Z., Wang, Y.,
  and Zhang, C.
\newblock video-salmonn: Speech-enhanced audio-visual large language models.
\newblock \emph{arXiv preprint arXiv:2406.15704}, 2024{\natexlab{a}}.

\bibitem[Sun et~al.(2024{\natexlab{b}})Sun, Huang, Liu, Wu, Xu, Li, and
  Liu]{sun2024t2v}
Sun, K., Huang, K., Liu, X., Wu, Y., Xu, Z., Li, Z., and Liu, X.
\newblock T2v-compbench: A comprehensive benchmark for compositional
  text-to-video generation.
\newblock \emph{arXiv preprint arXiv:2407.14505}, 2024{\natexlab{b}}.

\bibitem[Sun et~al.(2023)Sun, Yu, Cui, Zhang, Zhang, Wang, Gao, Liu, Huang, and
  Wang]{sun2023emu}
Sun, Q., Yu, Q., Cui, Y., Zhang, F., Zhang, X., Wang, Y., Gao, H., Liu, J.,
  Huang, T., and Wang, X.
\newblock Emu: Generative pretraining in multimodality.
\newblock \emph{arXiv preprint arXiv:2307.05222}, 2023.

\bibitem[Sun et~al.(2024{\natexlab{c}})Sun, Cui, Zhang, Zhang, Yu, Wang, Rao,
  Liu, Huang, and Wang]{sun2024generative}
Sun, Q., Cui, Y., Zhang, X., Zhang, F., Yu, Q., Wang, Y., Rao, Y., Liu, J.,
  Huang, T., and Wang, X.
\newblock Generative multimodal models are in-context learners.
\newblock In \emph{Proceedings of the IEEE/CVF Conference on Computer Vision
  and Pattern Recognition}, pp.\  14398--14409, 2024{\natexlab{c}}.

\bibitem[Sun et~al.(2024{\natexlab{d}})Sun, Tu, Liao, and Tao]{V2VBench}
Sun, W., Tu, R.-C., Liao, J., and Tao, D.
\newblock Diffusion model-based video editing: A survey.
\newblock \emph{arXiv preprint arXiv:2407.07111}, 2024{\natexlab{d}}.

\bibitem[Tan et~al.(2024)Tan, Zhuang, Montgomery, Tang, Cuadron, Wang, Popa,
  and Stoica]{tan2024judgebench}
Tan, S., Zhuang, S., Montgomery, K., Tang, W.~Y., Cuadron, A., Wang, C., Popa,
  R.~A., and Stoica, I.
\newblock Judgebench: A benchmark for evaluating llm-based judges.
\newblock \emph{arXiv preprint arXiv:2410.12784}, 2024.

\bibitem[Tang et~al.(2021)Tang, Wang, Liu, Rao, Li, and
  Li]{tang2021clip4caption}
Tang, M., Wang, Z., Liu, Z., Rao, F., Li, D., and Li, X.
\newblock Clip4caption: Clip for video caption.
\newblock In \emph{Proceedings of the 29th ACM International Conference on
  Multimedia}, pp.\  4858--4862, 2021.

\bibitem[Tang et~al.(2023{\natexlab{a}})Tang, Yang, Zhu, Zeng, and
  Bansal]{tang2023any}
Tang, Z., Yang, Z., Zhu, C., Zeng, M., and Bansal, M.
\newblock Any-to-any generation via composable diffusion.
\newblock \emph{Advances in Neural Information Processing Systems},
  36:\penalty0 16083--16099, 2023{\natexlab{a}}.

\bibitem[Tang et~al.(2023{\natexlab{b}})Tang, Yang, Zhu, Zeng, and
  Bansal]{tang2023codi}
Tang, Z., Yang, Z., Zhu, C., Zeng, M., and Bansal, M.
\newblock Any-to-any generation via composable diffusion.
\newblock \emph{Advances in Neural Information Processing Systems},
  36:\penalty0 16083--16099, 2023{\natexlab{b}}.

\bibitem[Team(2024)]{team2024chameleon}
Team, C.
\newblock Chameleon: Mixed-modal early-fusion foundation models.
\newblock \emph{arXiv preprint arXiv:2405.09818}, 2024.

\bibitem[Team et~al.(2024{\natexlab{a}})Team, Georgiev, Lei, Burnell, Bai,
  Gulati, Tanzer, Vincent, Pan, Wang, et~al.]{gemini}
Team, G., Georgiev, P., Lei, V.~I., Burnell, R., Bai, L., Gulati, A., Tanzer,
  G., Vincent, D., Pan, Z., Wang, S., et~al.
\newblock Gemini 1.5: Unlocking multimodal understanding across millions of
  tokens of context.
\newblock \emph{arXiv preprint arXiv:2403.05530}, 2024{\natexlab{a}}.

\bibitem[Team et~al.(2024{\natexlab{b}})Team, Modi, Veerubhotla, Rysbek, Huber,
  Wiltshire, Veprek, Gillick, Kasenberg, Ahmed, et~al.]{learnlm}
Team, L., Modi, A., Veerubhotla, A.~S., Rysbek, A., Huber, A., Wiltshire, B.,
  Veprek, B., Gillick, D., Kasenberg, D., Ahmed, D., et~al.
\newblock Learnlm: Improving gemini for learning.
\newblock \emph{arXiv preprint arXiv:2412.16429}, 2024{\natexlab{b}}.

\bibitem[Tschannen et~al.(2024)Tschannen, Pinto, and
  Kolesnikov]{tschannen2024jetformer}
Tschannen, M., Pinto, A.~S., and Kolesnikov, A.
\newblock Jetformer: An autoregressive generative model of raw images and text.
\newblock \emph{arXiv preprint arXiv:2411.19722}, 2024.

\bibitem[Van Den~Oord et~al.(2017)Van Den~Oord, Vinyals, et~al.]{van2017neural}
Van Den~Oord, A., Vinyals, O., et~al.
\newblock Neural discrete representation learning.
\newblock \emph{Advances in neural information processing systems}, 30, 2017.

\bibitem[Wang et~al.(2022)Wang, Wang, Hu, Zhang, Gao, Rong, and
  Dong]{wang2022v2a_mapper}
Wang, H., Wang, Q., Hu, J., Zhang, R., Gao, T., Rong, S., and Dong, H.
\newblock Global research trends in in-stent neoatherosclerosis: A
  citespace-based visual analysis.
\newblock \emph{Frontiers in Cardiovascular Medicine}, 9:\penalty0 1025858,
  2022.

\bibitem[Wang et~al.(2016)Wang, Yin, Wang, Wu, and Wang]{wang2016comprehensive}
Wang, K., Yin, Q., Wang, W., Wu, S., and Wang, L.
\newblock A comprehensive survey on cross-modal retrieval.
\newblock \emph{arXiv preprint arXiv:1607.06215}, 2016.

\bibitem[Wang et~al.(2024{\natexlab{a}})Wang, Bai, Tan, Wang, Fan, Bai, Chen,
  Liu, Wang, Ge, et~al.]{wang2024qwen2}
Wang, P., Bai, S., Tan, S., Wang, S., Fan, Z., Bai, J., Chen, K., Liu, X.,
  Wang, J., Ge, W., et~al.
\newblock Qwen2-vl: Enhancing vision-language model's perception of the world
  at any resolution.
\newblock \emph{arXiv preprint arXiv:2409.12191}, 2024{\natexlab{a}}.

\bibitem[Wang et~al.(2024{\natexlab{b}})Wang, Zhang, Luo, Sun, Cui, Wang,
  Zhang, Wang, Li, Yu, et~al.]{wang2024emu3}
Wang, X., Zhang, X., Luo, Z., Sun, Q., Cui, Y., Wang, J., Zhang, F., Wang, Y.,
  Li, Z., Yu, Q., et~al.
\newblock Emu3: Next-token prediction is all you need.
\newblock \emph{arXiv preprint arXiv:2409.18869}, 2024{\natexlab{b}}.

\bibitem[Wang et~al.(2024{\natexlab{c}})Wang, Zhuang, and Wu]{ModaVerse}
Wang, X., Zhuang, B., and Wu, Q.
\newblock Modaverse: Efficiently transforming modalities with llms.
\newblock In \emph{Proceedings of the IEEE/CVF Conference on Computer Vision
  and Pattern Recognition}, pp.\  26606--26616, 2024{\natexlab{c}}.

\bibitem[Wang et~al.(2024{\natexlab{d}})Wang, Guo, Huang, Huang, Wang, You, Li,
  and Zhao]{wang2024frieren}
Wang, Y., Guo, W., Huang, R., Huang, J., Wang, Z., You, F., Li, R., and Zhao,
  Z.
\newblock Frieren: Efficient video-to-audio generation with rectified flow
  matching.
\newblock \emph{arXiv e-prints}, pp.\  arXiv--2406, 2024{\natexlab{d}}.

\bibitem[Wang et~al.(2024{\natexlab{e}})Wang, Hu, Zhao, Lin, Juefei-Xu, Li,
  Han, Subramanyam, Chen, Chen, et~al.]{wang2024mllm}
Wang, Z., Hu, S., Zhao, S., Lin, X., Juefei-Xu, F., Li, Z., Han, L.,
  Subramanyam, H., Chen, L., Chen, J., et~al.
\newblock Mllm-as-a-judge for image safety without human labeling.
\newblock \emph{arXiv preprint arXiv:2501.00192}, 2024{\natexlab{e}}.

\bibitem[Wang et~al.(2024{\natexlab{f}})Wang, Zhu, Xu, Zhou, Liu, Zhang, Wang,
  Shi, Li, Li, et~al.]{wang2024mio}
Wang, Z., Zhu, K., Xu, C., Zhou, W., Liu, J., Zhang, Y., Wang, J., Shi, N., Li,
  S., Li, Y., et~al.
\newblock Mio: A foundation model on multimodal tokens.
\newblock \emph{arXiv preprint arXiv:2409.17692}, 2024{\natexlab{f}}.

\bibitem[Wei et~al.(2022)Wei, Wang, Schuurmans, Bosma, Xia, Chi, Le, Zhou,
  et~al.]{wei2022chain}
Wei, J., Wang, X., Schuurmans, D., Bosma, M., Xia, F., Chi, E., Le, Q.~V.,
  Zhou, D., et~al.
\newblock Chain-of-thought prompting elicits reasoning in large language
  models.
\newblock \emph{Advances in neural information processing systems},
  35:\penalty0 24824--24837, 2022.

\bibitem[Wu et~al.(2024{\natexlab{a}})Wu, Chen, Wu, Ma, Liu, Pan, Liu, Xie, Yu,
  Ruan, et~al.]{wu2024janus}
Wu, C., Chen, X., Wu, Z., Ma, Y., Liu, X., Pan, Z., Liu, W., Xie, Z., Yu, X.,
  Ruan, C., et~al.
\newblock Janus: Decoupling visual encoding for unified multimodal
  understanding and generation.
\newblock \emph{arXiv preprint arXiv:2410.13848}, 2024{\natexlab{a}}.

\bibitem[Wu et~al.(2024{\natexlab{b}})Wu, Jiang, Ma, Liu, Zhao, Yuan, Bai, and
  Bai]{wu2024liquid}
Wu, J., Jiang, Y., Ma, C., Liu, Y., Zhao, H., Yuan, Z., Bai, S., and Bai, X.
\newblock Liquid: Language models are scalable multi-modal generators.
\newblock \emph{arXiv preprint arXiv:2412.04332}, 2024{\natexlab{b}}.

\bibitem[Wu et~al.(2023{\natexlab{a}})Wu, Fei, Qu, Ji, and Chua]{nextgpt}
Wu, S., Fei, H., Qu, L., Ji, W., and Chua, T.-S.
\newblock Next-gpt: Any-to-any multimodal llm.
\newblock \emph{arXiv preprint arXiv:2309.05519}, 2023{\natexlab{a}}.

\bibitem[Wu et~al.(2024{\natexlab{c}})Wu, Yang, Li, Zhang, Liu, Guibas, Lin,
  and Wetzstein]{wu2024gpt}
Wu, T., Yang, G., Li, Z., Zhang, K., Liu, Z., Guibas, L., Lin, D., and
  Wetzstein, G.
\newblock Gpt-4v (ision) is a human-aligned evaluator for text-to-3d
  generation.
\newblock In \emph{Proceedings of the IEEE/CVF conference on computer vision
  and pattern recognition}, pp.\  22227--22238, 2024{\natexlab{c}}.

\bibitem[Wu et~al.(2023{\natexlab{b}})Wu, Hao, Sun, Chen, Zhu, Zhao, and
  Li]{wu2023human}
Wu, X., Hao, Y., Sun, K., Chen, Y., Zhu, F., Zhao, R., and Li, H.
\newblock Human preference score v2: A solid benchmark for evaluating human
  preferences of text-to-image synthesis.
\newblock \emph{arXiv preprint arXiv:2306.09341}, 2023{\natexlab{b}}.

\bibitem[Wu et~al.(2024{\natexlab{d}})Wu, Zhang, Chen, Tang, Li, Fang, Zhu,
  Xie, Yin, Yi, et~al.]{wu2024vila}
Wu, Y., Zhang, Z., Chen, J., Tang, H., Li, D., Fang, Y., Zhu, L., Xie, E., Yin,
  H., Yi, L., et~al.
\newblock Vila-u: a unified foundation model integrating visual understanding
  and generation.
\newblock \emph{arXiv preprint arXiv:2409.04429}, 2024{\natexlab{d}}.

\bibitem[xAI(2024)]{realworldqa2024}
xAI.
\newblock Realworldqa.
\newblock \url{https://huggingface.co/datasets/xai-org/RealworldQA}, 2024.

\bibitem[Xia et~al.(2024)Xia, Han, Qiu, Zhou, Wang, Zheng, Chen, Cui, Ding, Li,
  et~al.]{xia2024mmie}
Xia, P., Han, S., Qiu, S., Zhou, Y., Wang, Z., Zheng, W., Chen, Z., Cui, C.,
  Ding, M., Li, L., et~al.
\newblock Mmie: Massive multimodal interleaved comprehension benchmark for
  large vision-language models.
\newblock \emph{arXiv preprint arXiv:2410.10139}, 2024.

\bibitem[Xie et~al.(2024{\natexlab{a}})Xie, Mao, Bai, Zhang, Wang, Lin, Gu,
  Chen, Yang, and Shou]{xie2024show}
Xie, J., Mao, W., Bai, Z., Zhang, D.~J., Wang, W., Lin, K.~Q., Gu, Y., Chen,
  Z., Yang, Z., and Shou, M.~Z.
\newblock Show-o: One single transformer to unify multimodal understanding and
  generation.
\newblock \emph{arXiv preprint arXiv:2408.12528}, 2024{\natexlab{a}}.

\bibitem[Xie et~al.(2024{\natexlab{b}})Xie, Du, Song, and Liu]{xie2024muse}
Xie, R., Du, C., Song, P., and Liu, C.
\newblock Muse-vl: Modeling unified vlm through semantic discrete encoding.
\newblock \emph{arXiv preprint arXiv:2411.17762}, 2024{\natexlab{b}}.

\bibitem[Xing et~al.(2024{\natexlab{a}})Xing, Xia, Zhang, Chen, Yu, Liu, Liu,
  Wang, Shan, and Wong]{xing2024dynamicrafter}
Xing, J., Xia, M., Zhang, Y., Chen, H., Yu, W., Liu, H., Liu, G., Wang, X.,
  Shan, Y., and Wong, T.-T.
\newblock Dynamicrafter: Animating open-domain images with video diffusion
  priors.
\newblock In \emph{European Conference on Computer Vision}, pp.\  399--417.
  Springer, 2024{\natexlab{a}}.

\bibitem[Xing et~al.(2024{\natexlab{b}})Xing, He, Tian, Wang, and
  Chen]{xing2024seeing}
Xing, Y., He, Y., Tian, Z., Wang, X., and Chen, Q.
\newblock Seeing and hearing: Open-domain visual-audio generation with
  diffusion latent aligners.
\newblock In \emph{Proceedings of the IEEE/CVF Conference on Computer Vision
  and Pattern Recognition}, pp.\  7151--7161, 2024{\natexlab{b}}.

\bibitem[Xiong et~al.(2024{\natexlab{a}})Xiong, Wang, Guo, Ye, Fan, Gu, Huang,
  and Li]{LlavaCritic}
Xiong, T., Wang, X., Guo, D., Ye, Q., Fan, H., Gu, Q., Huang, H., and Li, C.
\newblock Llava-critic: Learning to evaluate multimodal models.
\newblock \emph{arXiv preprint arXiv:2410.02712}, 2024{\natexlab{a}}.

\bibitem[Xiong et~al.(2024{\natexlab{b}})Xiong, Wang, Guo, Ye, Fan, Gu, Huang,
  and Li]{xiong2024llava}
Xiong, T., Wang, X., Guo, D., Ye, Q., Fan, H., Gu, Q., Huang, H., and Li, C.
\newblock Llava-critic: Learning to evaluate multimodal models.
\newblock \emph{arXiv preprint arXiv:2410.02712}, 2024{\natexlab{b}}.

\bibitem[Yang et~al.(2024{\natexlab{a}})Yang, Yang, Zhang, Hui, Zheng, Yu, Li,
  Liu, Huang, Wei, et~al.]{yang2024qwen2}
Yang, A., Yang, B., Zhang, B., Hui, B., Zheng, B., Yu, B., Li, C., Liu, D.,
  Huang, F., Wei, H., et~al.
\newblock Qwen2. 5 technical report.
\newblock \emph{arXiv preprint arXiv:2412.15115}, 2024{\natexlab{a}}.

\bibitem[Yang et~al.(2024{\natexlab{b}})Yang, Yin, Zhou, Rao, Zhai, Cao, and
  Zha]{yang2024mmar}
Yang, J., Yin, D., Zhou, Y., Rao, F., Zhai, W., Cao, Y., and Zha, Z.-J.
\newblock Mmar: Towards lossless multi-modal auto-regressive probabilistic
  modeling.
\newblock \emph{arXiv preprint arXiv:2410.10798}, 2024{\natexlab{b}}.

\bibitem[Yang et~al.(2024{\natexlab{c}})Yang, Xu, Liu, Chu, Jiang, Zhou, Leng,
  Lv, Zhao, Zhou, et~al.]{yang2024air}
Yang, Q., Xu, J., Liu, W., Chu, Y., Jiang, Z., Zhou, X., Leng, Y., Lv, Y.,
  Zhao, Z., Zhou, C., et~al.
\newblock Air-bench: Benchmarking large audio-language models via generative
  comprehension.
\newblock \emph{arXiv preprint arXiv:2402.07729}, 2024{\natexlab{c}}.

\bibitem[Yang et~al.(2024{\natexlab{d}})Yang, Teng, Zheng, Ding, Huang, Xu,
  Yang, Hong, Zhang, Feng, et~al.]{yang2024cogvideox}
Yang, Z., Teng, J., Zheng, W., Ding, M., Huang, S., Xu, J., Yang, Y., Hong, W.,
  Zhang, X., Feng, G., et~al.
\newblock Cogvideox: Text-to-video diffusion models with an expert transformer.
\newblock \emph{arXiv preprint arXiv:2408.06072}, 2024{\natexlab{d}}.

\bibitem[Yariv et~al.(2024)Yariv, Gat, Benaim, Wolf, Schwartz, and
  Adi]{yariv2024tempotokens}
Yariv, G., Gat, I., Benaim, S., Wolf, L., Schwartz, I., and Adi, Y.
\newblock Diverse and aligned audio-to-video generation via text-to-video model
  adaptation.
\newblock In \emph{Proceedings of the AAAI Conference on Artificial
  Intelligence}, volume~38, pp.\  6639--6647, 2024.

\bibitem[Yasunaga et~al.(2025)Yasunaga, Zettlemoyer, and
  Ghazvininejad]{Yasunaga2025MultimodalRH}
Yasunaga, M., Zettlemoyer, L., and Ghazvininejad, M.
\newblock Multimodal rewardbench: Holistic evaluation of reward models for
  vision language models.
\newblock 2025.
\newblock URL \url{https://api.semanticscholar.org/CorpusID:276482127}.

\bibitem[Ye et~al.(2024)Ye, Wang, Huang, Chen, Zhang, Moniz, Gao, Geyer, Huang,
  Chen, et~al.]{ye2024justice}
Ye, J., Wang, Y., Huang, Y., Chen, D., Zhang, Q., Moniz, N., Gao, T., Geyer,
  W., Huang, C., Chen, P.-Y., et~al.
\newblock Justice or prejudice? quantifying biases in llm-as-a-judge.
\newblock \emph{arXiv preprint arXiv:2410.02736}, 2024.

\bibitem[Yu et~al.(2024)Yu, Zhang, Yao, Dang, Chen, Lu, Cui, He, Liu, Chua,
  et~al.]{yu2024rlaif}
Yu, T., Zhang, H., Yao, Y., Dang, Y., Chen, D., Lu, X., Cui, G., He, T., Liu,
  Z., Chua, T.-S., et~al.
\newblock Rlaif-v: Aligning mllms through open-source ai feedback for super
  gpt-4v trustworthiness.
\newblock \emph{arXiv preprint arXiv:2405.17220}, 2024.

\bibitem[Yu et~al.(2023)Yu, Yang, Li, Wang, Lin, Liu, Wang, and Wang]{yu2023mm}
Yu, W., Yang, Z., Li, L., Wang, J., Lin, K., Liu, Z., Wang, X., and Wang, L.
\newblock Mm-vet: Evaluating large multimodal models for integrated
  capabilities.
\newblock \emph{arXiv preprint arXiv:2308.02490}, 2023.

\bibitem[Yue et~al.(2024)Yue, Ni, Zhang, Zheng, Liu, Zhang, Stevens, Jiang,
  Ren, Sun, et~al.]{yue2024mmmu}
Yue, X., Ni, Y., Zhang, K., Zheng, T., Liu, R., Zhang, G., Stevens, S., Jiang,
  D., Ren, W., Sun, Y., et~al.
\newblock Mmmu: A massive multi-discipline multimodal understanding and
  reasoning benchmark for expert agi.
\newblock In \emph{Proceedings of the IEEE/CVF Conference on Computer Vision
  and Pattern Recognition}, pp.\  9556--9567, 2024.

\bibitem[Zhang et~al.(2024{\natexlab{a}})Zhang, Du, Chen, Liang, Luo, Zheng,
  Zhu, Cheng, Xu, Guo, et~al.]{zhang2024cmmmu}
Zhang, G., Du, X., Chen, B., Liang, Y., Luo, T., Zheng, T., Zhu, K., Cheng, Y.,
  Xu, C., Guo, S., et~al.
\newblock Cmmmu: A chinese massive multi-discipline multimodal understanding
  benchmark.
\newblock \emph{arXiv preprint arXiv:2401.11944}, 2024{\natexlab{a}}.

\bibitem[Zhang et~al.(2023{\natexlab{a}})Zhang, Li, and Bing]{zhang2023video}
Zhang, H., Li, X., and Bing, L.
\newblock Video-llama: An instruction-tuned audio-visual language model for
  video understanding.
\newblock \emph{arXiv preprint arXiv:2306.02858}, 2023{\natexlab{a}}.

\bibitem[Zhang et~al.(2023{\natexlab{b}})Zhang, Mo, Chen, Sun, and
  Su]{zhang2023magicbrush}
Zhang, K., Mo, L., Chen, W., Sun, H., and Su, Y.
\newblock Magicbrush: A manually annotated dataset for instruction-guided image
  editing.
\newblock \emph{Advances in Neural Information Processing Systems},
  36:\penalty0 31428--31449, 2023{\natexlab{b}}.

\bibitem[Zhang et~al.(2024{\natexlab{b}})Zhang, Mo, Zhang, and
  Morgado]{AVSYNC15}
Zhang, L., Mo, S., Zhang, Y., and Morgado, P.
\newblock Audio-synchronized visual animation.
\newblock In \emph{Proceedings of the European Conference on Computer Vision
  (ECCV)}, 2024{\natexlab{b}}.

\bibitem[Zhang et~al.(2023{\natexlab{c}})Zhang, Wei, Jiang, Zhang, Zuo, and
  Tian]{zhang2023controlvideo}
Zhang, Y., Wei, Y., Jiang, D., Zhang, X., Zuo, W., and Tian, Q.
\newblock Controlvideo: Training-free controllable text-to-video generation.
\newblock \emph{arXiv preprint arXiv:2305.13077}, 2023{\natexlab{c}}.

\bibitem[Zhang et~al.(2025)Zhang, Yu, Tian, Fu, Li, Zeng, Xie, Shi, Zhang, Wu,
  et~al.]{zhang2025mm}
Zhang, Y.-F., Yu, T., Tian, H., Fu, C., Li, P., Zeng, J., Xie, W., Shi, Y.,
  Zhang, H., Wu, J., et~al.
\newblock Mm-rlhf: The next step forward in multimodal llm alignment.
\newblock \emph{arXiv preprint arXiv:2502.10391}, 2025.

\bibitem[Zhao et~al.(2024)Zhao, Song, Wang, Feng, Ding, Sun, Xiao, and
  Wang]{zhao2024monoformer}
Zhao, C., Song, Y., Wang, W., Feng, H., Ding, E., Sun, Y., Xiao, X., and Wang,
  J.
\newblock Monoformer: One transformer for both diffusion and autoregression.
\newblock \emph{arXiv preprint arXiv:2409.16280}, 2024.

\bibitem[Zhao et~al.(2025)Zhao, Xie, Zhang, Gan, Long, Hu, Hu, Chen, Li, Song,
  et~al.]{zhao2025mmvu}
Zhao, Y., Xie, L., Zhang, H., Gan, G., Long, Y., Hu, Z., Hu, T., Chen, W., Li,
  C., Song, J., et~al.
\newblock Mmvu: Measuring expert-level multi-discipline video understanding.
\newblock \emph{arXiv preprint arXiv:2501.12380}, 2025.

\bibitem[Zhen et~al.(2019)Zhen, Hu, Wang, and Peng]{zhen2019deep}
Zhen, L., Hu, P., Wang, X., and Peng, D.
\newblock Deep supervised cross-modal retrieval.
\newblock In \emph{Proceedings of the IEEE/CVF conference on computer vision
  and pattern recognition}, pp.\  10394--10403, 2019.

\bibitem[Zheng et~al.(2023)Zheng, Chiang, Sheng, Zhuang, Wu, Zhuang, Lin, Li,
  Li, Xing, et~al.]{LLM-as-a-judge}
Zheng, L., Chiang, W.-L., Sheng, Y., Zhuang, S., Wu, Z., Zhuang, Y., Lin, Z.,
  Li, Z., Li, D., Xing, E., et~al.
\newblock Judging llm-as-a-judge with mt-bench and chatbot arena.
\newblock \emph{Advances in Neural Information Processing Systems},
  36:\penalty0 46595--46623, 2023.

\bibitem[Zhou et~al.(2023)Zhou, Lu, Mishra, Brahma, Basu, Luan, Zhou, and
  Hou]{zhou2023instruction}
Zhou, J., Lu, T., Mishra, S., Brahma, S., Basu, S., Luan, Y., Zhou, D., and
  Hou, L.
\newblock Instruction-following evaluation for large language models.
\newblock \emph{arXiv preprint arXiv:2311.07911}, 2023.

\bibitem[Zhou et~al.(2024)Zhou, Peng, Song, Li, Xu, Yang, Guo, Zhang, Lin, He,
  et~al.]{zhou2024gate}
Zhou, P., Peng, X., Song, J., Li, C., Xu, Z., Yang, Y., Guo, Z., Zhang, H.,
  Lin, Y., He, Y., et~al.
\newblock Gate opening: A comprehensive benchmark for judging open-ended
  interleaved image-text generation.
\newblock \emph{arXiv preprint arXiv:2411.18499}, 2024.

\bibitem[Zhu et~al.(2023)Zhu, Chen, Shen, Li, and Elhoseiny]{zhu2023minigpt}
Zhu, D., Chen, J., Shen, X., Li, X., and Elhoseiny, M.
\newblock Minigpt-4: Enhancing vision-language understanding with advanced
  large language models.
\newblock \emph{arXiv preprint arXiv:2304.10592}, 2023.

\bibitem[Zhuge et~al.(2024)Zhuge, Zhao, Ashley, Wang, Khizbullin, Xiong, Liu,
  Chang, Krishnamoorthi, Tian, et~al.]{zhuge2024agent}
Zhuge, M., Zhao, C., Ashley, D., Wang, W., Khizbullin, D., Xiong, Y., Liu, Z.,
  Chang, E., Krishnamoorthi, R., Tian, Y., et~al.
\newblock Agent-as-a-judge: Evaluate agents with agents.
\newblock \emph{arXiv preprint arXiv:2410.10934}, 2024.

\bibitem[Zhuo et~al.(2023)Zhuo, Wang, Wang, Liao, Bao, Peng, Han, Zhang, Fang,
  and Liu]{zhuo2023video}
Zhuo, L., Wang, Z., Wang, B., Liao, Y., Bao, C., Peng, S., Han, S., Zhang, A.,
  Fang, F., and Liu, S.
\newblock Video background music generation: Dataset, method and evaluation.
\newblock In \emph{Proceedings of the IEEE/CVF International Conference on
  Computer Vision}, pp.\  15637--15647, 2023.

\bibitem[Ziv et~al.(2024)Ziv, Gat, Lan, Remez, Kreuk, D{\'e}fossez, Copet,
  Synnaeve, and Adi]{ziv2024magnet}
Ziv, A., Gat, I., Lan, G.~L., Remez, T., Kreuk, F., D{\'e}fossez, A., Copet,
  J., Synnaeve, G., and Adi, Y.
\newblock Masked audio generation using a single non-autoregressive
  transformer.
\newblock \emph{arXiv preprint arXiv:2401.04577}, 2024.

\end{thebibliography}
\bibliographystyle{colm2025_conference}

\appendix
\section{Related Work}
\label{Appendix: Full Related Works}
\header{Multimodal Understanding (MMU).}
MMU involves integrating and processing information from multiple modalities—such as text \citep{li2023cmmlu,li2023cmmlu}, images \citep{yue2024mmmu,zhang2024cmmmu}, video \citep{zhao2025mmvu,hu2025video}, audio \citep{sakshi2024mmau}—achieve significant development since the advent of Multimodal Large Language Models (MLLMs). Through late fusion of modality features with pretrained LLMs and modality instruction tuning \citep{liu2023visual}, MLLMs gain incredibly advanced understanding capabilities in images \citep{zhu2023minigpt,li2023blip}, videos \citep{lin2023video,zhang2023video}, audio \citep{chu2023qwen, chu2024qwen2}, and even interleaved content \citep{chen2025interleaved}, transforming traditional MMG tasks such as Visual Question Answering (VQA) \citep{antol2015vqa,krishna2017visual}, Multimodal Captioning \citep{bai2018survey,tang2021clip4caption}, and cross-modal retrieval \citep{wang2016comprehensive,zhen2019deep} in a unified manner.

Recent benchmarks like MMMU \citep{yue2024mmmu}, Video-MMMU \citep{hu2025video}, and MMAU \citep{sakshi2024mmau} have been developed to rigorously evaluate understanding and reasoning in \emph{specific modality} of MLLMs. Any-to-Any benchmarks also emerge to provide comprehensive assessment for current well-rounded models capable of understanding in many modalities \citep{chen2024omnixr, li2024omnibench,ni2024mixeval,hong2025worldsense}. However, these benchmarks evaluate MMG mainly through Multiple-Choice QA, undermining the reliable assessment of real-world open-ended queries \citep{realworldqa2024}.

\header{Multimodal Generation (MMG).}
MMG involves generating content in one modality based on input from another or mixture \citep{ni2024mixeval,chen2025interleaved}, such as text-to-image \citep{ghosh2023geneval}, text-to-video \citep{sun2024t2v,huang2024vbench}, image-to-video \citep{V2VBench,fan2023aigcbench}, video-to-music \citep{kang2024video2music,zhuo2023video} and other modality transformations \citep{seung2023musiccaps}. Early approaches primarily focused on building specific framework for each task \citep{openai_dalle3,yang2024cogvideox}, which were later unified by Auto-Regressive (AR) models such as Show-o \citep{xie2024show}, Emu-3 \citep{wang2024emu3}, and Unified-IO \citep{lu2022unified,unifiedIO2} which enabling generate various modalities in a more coherent and complex manner.

However, the open-ended nature of MMG tasks makes evaluation challenging, as traditional ground-truth-based metrics fail to capture the diversity and quality of generated content \citep{ni2024mixeval}. Human-oriented evaluations, such as those on crowdsourcing platforms like GenAI-Arena \citep{jiang2025genai} and other benchmarks \citep{liang2024rich,huang2024vbench}, have become a common approach to assess MMG tasks. Despite their utility, these platforms often suffer from issues like insufficient votes, leading to instability in rankings \citep{mllm-as-a-judge}. Our work advances the uniform incorporation of MLLM-as-a-Judge across various modalities by implementing checklist-of-thought reasoning to achieve more unbiased, reliable, and reproducible evaluations of MMG tasks.

\header{Multimodal LLM-as-a-Judge.}
Originated from Natural Language Generation (NLG) domain, LLM-as-a-Judge \citep{LLM-as-a-judge} have extended to multimodal domains serving as evaluation metrics in general QA \citep{mllm-as-a-judge,xiong2024llava,lee2024prometheus,tan2024judgebench}, image generation \citep{chen2024mj,lin2024evaluating}, video generation \citep{luo2024videoautoarena}, 3D synthesis \citep{wu2024gpt}, SWE tasks \citep{zhuge2024agent}, and interleaved generation \citep{chen2025interleaved,zhou2024gate}. Another line leverage pretrained MLLMs serving as reward models \citep{Yasunaga2025MultimodalRH,li2024vlrewardbench,yu2024rlaif,zhang2025mm,wang2024mllm} in aligning other MLLMs for advanced performance. MLLM-as-a-Judge \citep{mllm-as-a-judge} takes the first step in systematically quantifying MLLMs' performance as judges and assessing potential problems such as bias and hallucinations in Vision-Language Understanding tasks. Our work extends this systematic assessing framework for MLLM-as-a-Judge to 15 Any-to-Any MMU and MMG tasks, with carefully selected samples for open-ended queries, providing in-depth analysis of potential challenges when applying MLLM-as-a-Judge across broader modalities and more general use cases.

\header{Any-to-Any Unified Models.} We term models that can take and generate with various modalities as \emph{\textbf{Unified Models}}, which unifies different modalities into the paradigm of next token prediction with an auto-regressive structure, leveraging a powerful pretrained LLM backbone. 
By tokenizing continuous contents into discrete tokens using different tokenizers~\citep{van2017neural,razavi2019generating}, researchers have started to explore simultaneously visual understanding and generation with a single backbone \citep{li2024dual,shi2024llamafusion,li2024synergen,wu2024liquid,qu2024tokenflow,li2024omniflow,ma2024janusflow,xie2024muse,tschannen2024jetformer,kou2024orthus,lai2024spider,wu2024janus,wu2024vila,zhao2024monoformer,wang2024mio,yang2024mmar,chern2024anole,li2024mini,team2024chameleon,sun2023emu,sun2024generative}. Other pioneer works extend this boundary into other modalities such as video \citep{wang2024emu3}, audio \citep{tang2023any}, conditions \citep{Mizrahi20234MMM,Bachmann20244M21AA}, and 3D assets generation \citep{chen2024sar3d} in an Auto-Regressive manner.

\section{Benchmark Details}
\subsection{Benchmark Construction}
We sample open-ended queries from previous benchmarks (Table \ref{tab: benchmark sources}) randomly and conduct manually filtering for their quality.
\begin{table}
  \centering
  \caption{A detailed breakdown of our benchmark Sources. All samples are open-ended queries.}
  \label{tab: benchmark sources}
    \label{BenchmarkConstruction}
    \begin{small}
        \begin{tabular}{lccr}
          \toprule[1.5pt]
          Task & Source & Number \\
          \midrule
          Text-to-Text       & WildBench \citep{lin2024wildbench}      & 50  \\
                             & IfEval \citep{zhou2023instruction}       & 50  \\  
          \midrule
          Image-to-Text      & DemonBench \citep{li2023fine}    & 25  \\ 
                             & MMVET \citep{yu2023mm}         & 25  \\
                             & TouchStone \citep{bai2023touchstone}     & 25  \\
                             & VisITBench \citep{bitton2023visit}     & 25  \\
          \midrule
          Video-to-Text      & CVRR \citep{khattak2024good}           & 25  \\
                             & AutoEval \citep{chen2024autoeval}     & 25  \\
                             & TemporalBench \citep{cai2024temporalbench} & 25  \\
                             & VideoMME \citep{fu2024video}      & 25  \\
          \midrule
          Audio-to-Text      & LTU \citep{gong2023listen}          & 26  \\
                             & VoiceBench \citep{chen2024voicebench}   & 37  \\
                             & AirBench \citep{yang2024air}      & 37  \\
          \midrule
          A+V-to-Text& Valor\_AVQA    & 100 \\
          \midrule
          Text-to-Image      & HPSv2 \citep{wu2023human}         & 50  \\
                             & T2ICompBench \citep{huang2023t2i}  & 50  \\
          \midrule
          Text-to-Video      & VBench \citep{huang2024vbench}        & 50  \\
                             & T2VCompBench \citep{sun2024t2v}  & 50  \\
          \midrule
          Text-to-Audio      & AudioCaps \citep{kim-etal-2019-audiocaps}     & 50  \\
                             & Clotho \citep{drossos2020clotho}        & 50  \\
          \midrule
          Image Edit         & I$^2$EBench \citep{ma2024i2ebench}     & 100 \\
          \midrule
          Image-to-Video     & ConsistI2V \citep{ren2024consisti2v}    & 100 \\
          \midrule
          Image-to-Audio     & ImageHear \citep{sheffer2023im2wav}     & 100 \\
          \midrule
          Video Edit         & V2VBench \citep{V2VBench}      & 100 \\
          \midrule
          Video-to-Audio     & VGGSound \citep{VGGSOUND}      & 100 \\
          \midrule
          Audio Edit         & AudioEditor \citep{jia2024audioeditor}   & 100 \\
          \midrule
          Audio-to-Video     & AVSync15 \citep{AVSYNC15}       & 100 \\
          \bottomrule[1.5pt]
        \end{tabular}
    \end{small}
\end{table}

\label{Appendix: Benchmark Construction}

\subsection{Safety Checking}
In this section, we provide a detailed analysis of trustworthiness problems in \task and \judge, focusing on NSFW content in text and multimodal content separately. 

\begin{figure}[h]
  \includegraphics[width=\columnwidth]{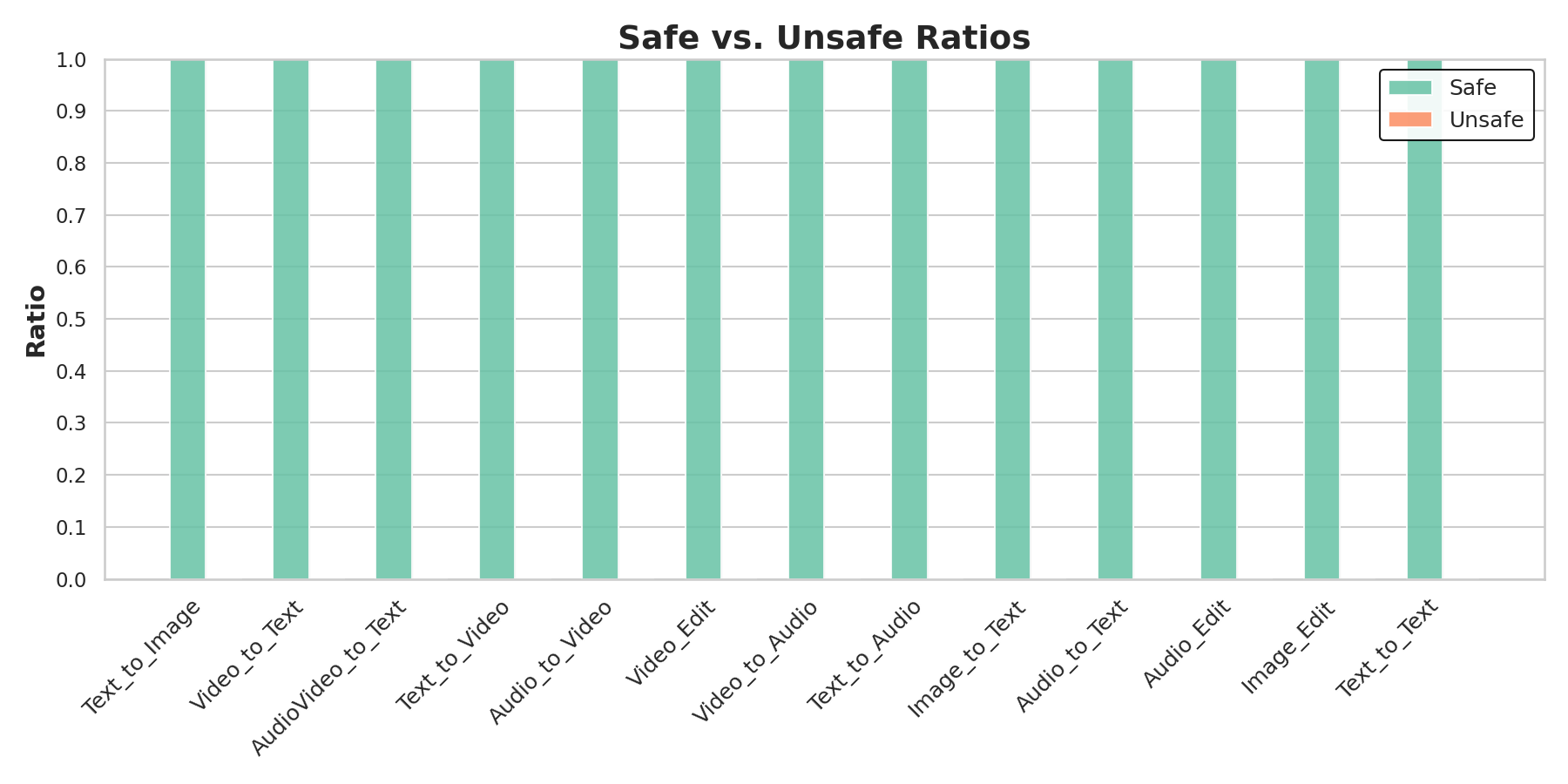}
  \caption{Safe \textit{v.s.} Unsafe content ratios across multimodal tasks.}
  \label{fig:NFSW}
\end{figure}

\header{NSFW Image Filtering.}
Figure~\ref{fig:NFSW} illustrates the proportion of unsafe and safe images across all categories based on the model's judgments. Out of all the images used, each sample is classified as \emph{Safety}.

\header{NSFW Filtering for Other Modalities.}
We conduct rigorous NSFW checks during dataset construction. Human annotators manually review all videos and audio clips to ensure they met NSFW safety standards. Since these samples are primarily sourced from established benchmarks, all are classified as \emph{Safe} by the annotators.

\subsection{Human Annotation Details}
\label{Appendix: Human Annotation Details}
The annotation is conducted by 10 authors of this paper and 2 volunteers independently. As acknowledged, the diversity of annotators plays a crucial role in reducing bias and enhancing the reliability of the benchmark. These annotators have knowledge in this domain, with different genders, ages, and educational backgrounds. To ensure the annotators can proficiently mark the data, we provide them with detailed tutorials, teaching them how to evaluate model responses more objectively as follows:
\begin{itemize}[leftmargin=*]
    \item \textbf{Checklist Filter and Refinement.}
    \item \textbf{Human Annotation in \score and \pair.}
\end{itemize}

\begin{figure*}[htbp]
  \centering
  \begin{tcolorbox}[
    enhanced,
    attach boxed title to top center={yshift=-3mm,yshifttext=-1mm},
    boxrule=0.9pt,
    colback=gray!00,
    colframe=black!50,
    colbacktitle=Periwinkle,
    title=Annotation screenshot and guideline,
    boxed title style={size=small,colframe=Periwinkle},
    top=3mm,
    bottom=3mm,
    left=2mm,
    right=2mm,
    before upper={\parindent0em\linespread{0.85}\selectfont} 
  ]
  \vspace{-1ex}
  \begin{minipage}[t]{0.96\textwidth}
  \textbf{Annotation Usage:} streamlit run annotation\_{score/pair}.py
    
    \smallskip
    For operations within the box, you need to click "Apply" for it to take effect. However, after clicking "Apply", it will switch to the next one, preventing the Streamlit rollback issue. You can quickly switch between the same Task and Rubric using "Next" and "Previous" without needing to click "Apply".
  \end{minipage}
  
  \vspace{1ex}
  \begin{minipage}[t]{0.48\textwidth}
    \centering
    \includegraphics[width=0.6\linewidth,trim=5 5 5 5,clip]{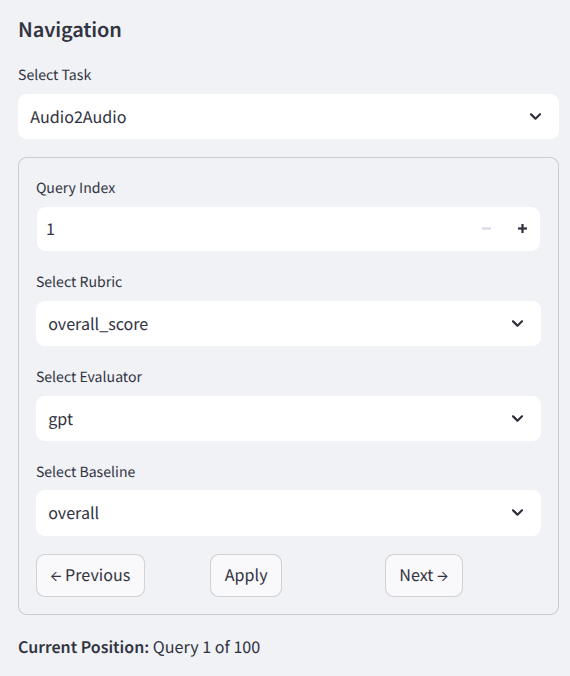}
    \captionof{subfigure}{Annotation interface's navigation bar}
  \end{minipage}
  \hfill
  \begin{minipage}[t]{0.48\textwidth}
    \centering
    \includegraphics[width=\linewidth,trim=5 5 5 5,clip]{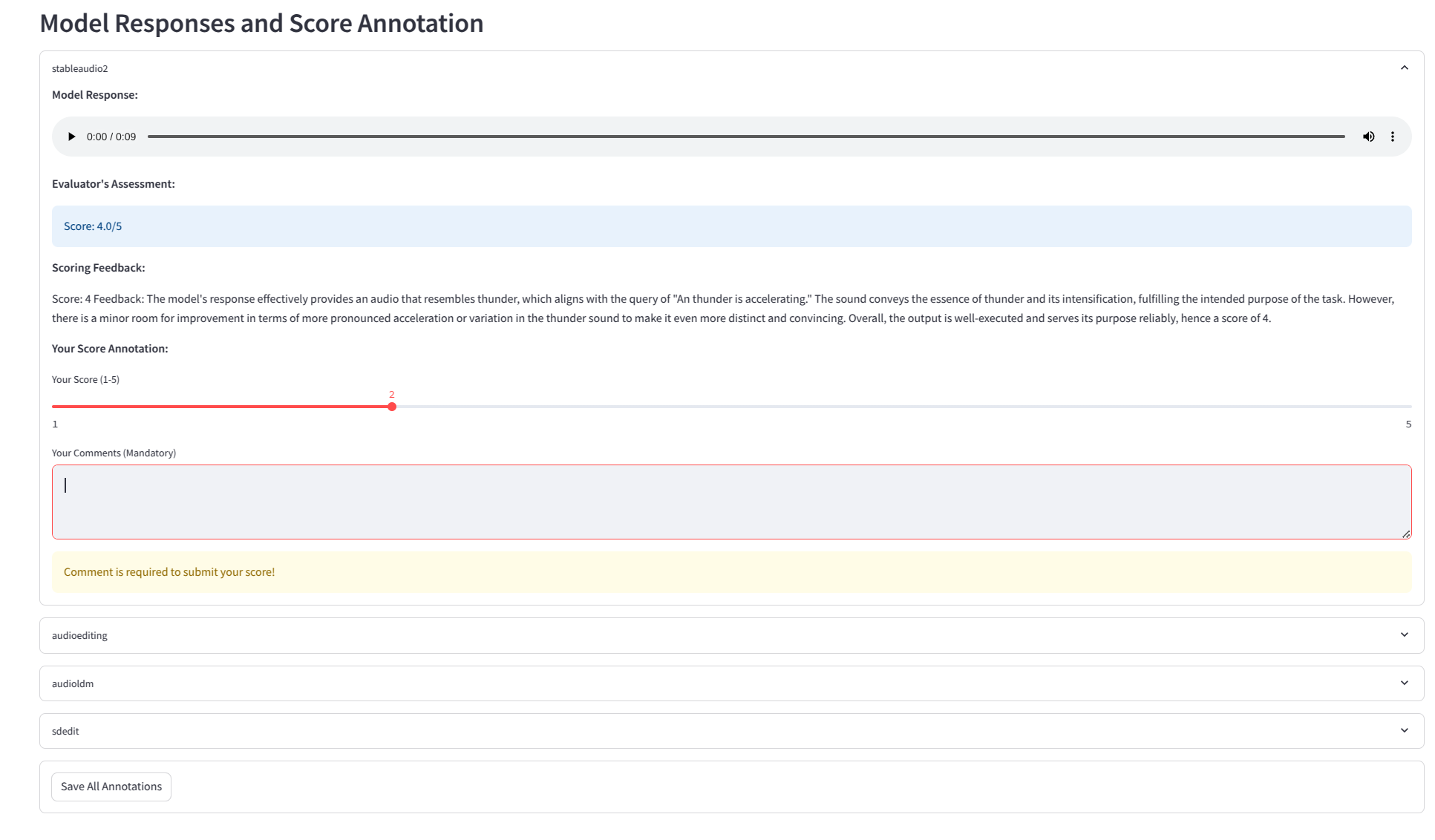}
    \captionof{subfigure}{\score annotation interface's model response selection}
    \label{fig: NavBar}
  \end{minipage}

  \vspace{1ex}
  \begin{minipage}[t]{0.96\textwidth}
    You just need to select the corresponding \score and \pair based on your understanding and judgment. It will be saved automatically. Please carefully check your annotation file to avoid major mistakes. You should be able to recover from the backup. The right figure is an example of the answers from each model, which you can click to open.

    \medskip
  \end{minipage}
  
  \vspace{1ex}
  \begin{minipage}[t]{0.96\textwidth}
    \centering
    \includegraphics[width=0.6\linewidth,trim=5 5 5 5,clip]{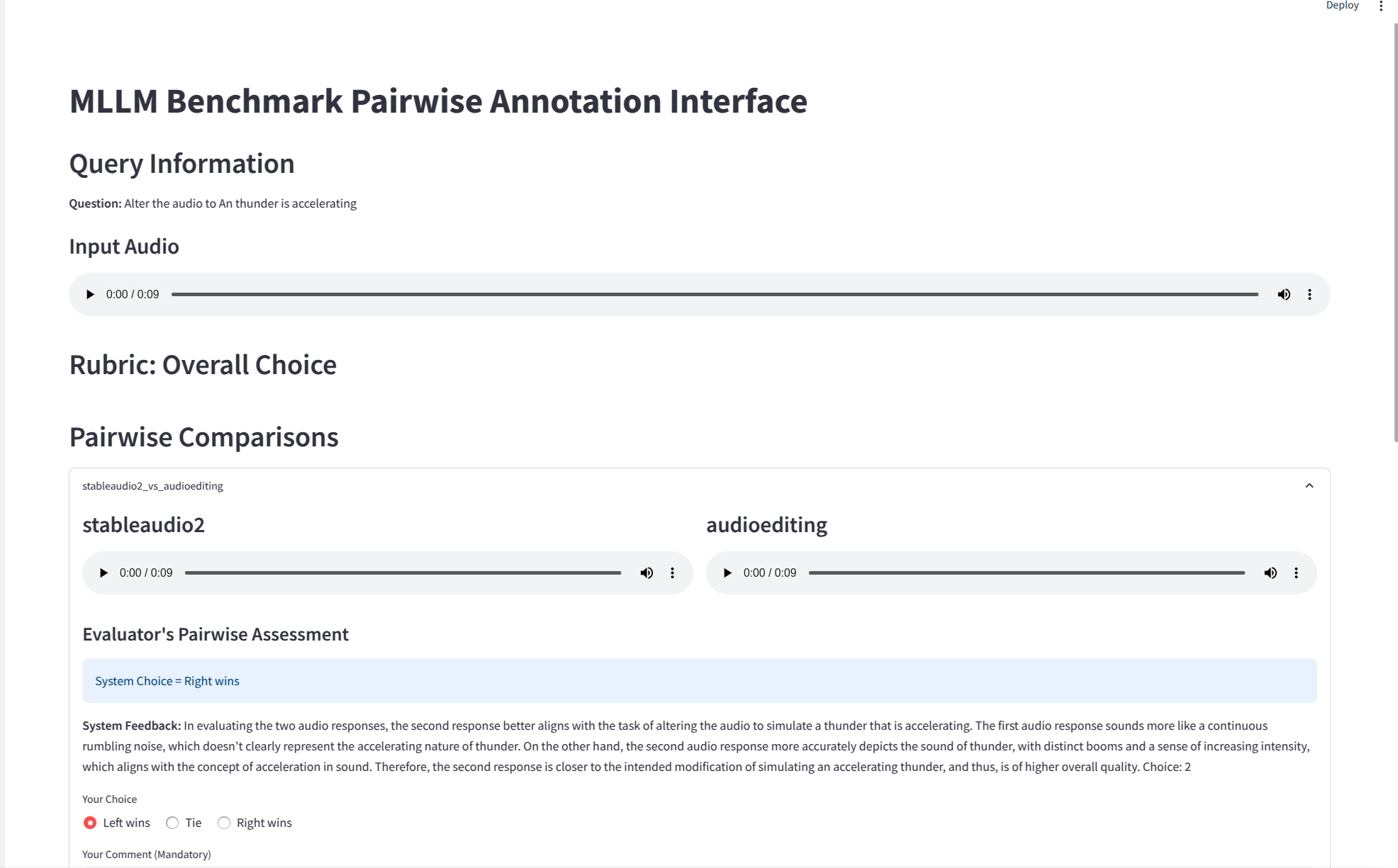}
  \end{minipage}
  
  \end{tcolorbox}
  \caption{Instructions for human annotation.}
  \label{}
\end{figure*}

\subsection{Copyright}
Given that we sample queries from previous well-established benchmarks to form \task and collect \emph{state-of-the-art} models' responses to curate \judge, we will release our code, benchmark, and dataset under the \texttt{Creative Commons 4.0} license rather than the \texttt{Apache} license, to maintain compatibility with the original licenses of these benchmarks and models.

\begin{table*}[!t]
\centering
\caption{Percentage of choices for tasks under \pair\ setting. \textbf{F} refers to selecting the first response; \textbf{S} refers to the second; \textbf{T} refers to a tie.}
\setlength{\tabcolsep}{2pt}
\resizebox{\textwidth}{!}{
\begin{tabular}{ll|*{15}{ccc}}
\toprule[1.5pt]
\multirow{2}{*}{} & \multirow{2}{*}{\textbf{Settings}}
& \multicolumn{3}{c}{\textbf{T$\rightarrow$T}}
& \multicolumn{3}{c}{\textbf{I$\rightarrow$T}}
& \multicolumn{3}{c}{\textbf{A$\rightarrow$T}}
& \multicolumn{3}{c}{\textbf{V$\rightarrow$T}}
& \multicolumn{3}{c}{\textbf{V+A$\rightarrow$T}}
& \multicolumn{3}{c}{\textbf{T$\rightarrow$I}}
& \multicolumn{3}{c}{\textbf{T$\rightarrow$V}}
& \multicolumn{3}{c}{\textbf{T$\rightarrow$A}}
& \multicolumn{3}{c}{\textbf{I$\rightarrow$I}}
& \multicolumn{3}{c}{\textbf{I$\rightarrow$V}}
& \multicolumn{3}{c}{\textbf{I$\rightarrow$A}}
& \multicolumn{3}{c}{\textbf{V$\rightarrow$V}}
& \multicolumn{3}{c}{\textbf{V$\rightarrow$A}}
& \multicolumn{3}{c}{\textbf{A$\rightarrow$V}}
& \multicolumn{3}{c}{\textbf{A$\rightarrow$A}} \\
&
& F & T & S
& F & T & S
& F & T & S
& F & T & S
& F & T & S
& F & T & S
& F & T & S
& F & T & S
& F & T & S
& F & T & S
& F & T & S
& F & T & S
& F & T & S
& F & T & S
& F & T & S \\
\midrule
\multicolumn{47}{c}{\textcolor{violet!80!white}{\cellcolor{gray!10!white}{\textbf{\textit{Judging Models}}}}} \\
\midrule

\multirow{3}{*}{GPT-4o}

& \overall
& 58.0 & 9.5 & 32.5
& 68.5 & 6.0 & 25.5
& 88.5 & 1.0 & 10.5
& 45.5 & 3.0 & 51.5
& 73.0 & 4.5 & 22.5
& 48.5 & 12.5 & 39.0
& 18.5 & 0.0 & 81.5
& 12.0 & 83.0 & 5.0
& 32.5 & 10.0 & 57.5
& 56.0 & 6.5 & 37.5
& 67.5 & 1.0 & 31.5
& 18.5 & 0.5 & 81.0
& 56.5 & 6.0 & 37.5
& 47.0 & 8.5 & 44.5
& 47.0 & 7.0 & 46.0 \\

& \rubric
& 54.33 & 14.75 & 30.92
& 68.25 & 8.92 & 22.83
& 86.42 & 2.42 & 11.17
& 50.33 & 4.08 & 45.58
& 76.33 & 3.08 & 20.58
& 6.17 & 86.75 & 7.08
& 0.08 & 99.83 & 0.08
& 52.5 & 12.25 & 35.25
& 0.0 & 100.0 & 0.0
& 0.25 & 99.67 & 0.08
& 9.67 & 85.42 & 4.92
& 0.08 & 99.83 & 0.08
& 36.08 & 46.0 & 17.92
& 42.92 & 54.75 & 2.33
& 5.17 & 91.42 & 3.42 \\

& \checklist
& 40.83 & 27.75 & 31.42
& 64.58 & 10.5 & 24.92
& 81.33 & 4.67 & 14.0
& 47.58 & 5.42 & 47.0
& 72.25 & 5.17 & 22.58
& 0.0 & 100.0 & 0.0
& 0.0 & 100.0 & 0.0
& 45.0 & 14.0 & 41.0
& 0.0 & 100.0 & 0.0
& 0.5 & 99.0 & 0.5
& 2.58 & 96.17 & 1.25
& 0.0 & 100.0 & 0.0
& 28.17 & 53.08 & 18.75
& 30.5 & 60.58 & 8.92
& 2.75 & 93.75 & 3.5 \\

\midrule

\multirow{3}{*}{Gemini-1.5-Pro}

& \overall
& 36.5 & 0.0 & 63.5
& 65.5 & 0.0 & 34.5
& 86.0 & 0.0 & 14.0
& 40.0 & 0.0 & 60.0
& 75.0 & 0.5 & 24.5
& 41.0 & 0.5 & 58.5
& 11.5 & 0.0 & 88.5
& 43.0 & 0.0 & 57.0
& 34.0 & 5.5 & 60.5
& 12.0 & 1.0 & 87.0
& 23.5 & 0.0 & 76.5
& 6.0 & 0.0 & 94.0
& 7.0 & 0.0 & 93.0
& 41.0 & 0.0 & 59.0
& 20.5 & 0.0 & 79.5 \\

& \rubric
& 40.5 & 0.33 & 59.17
& 67.33 & 0.58 & 32.08
& 85.5 & 0.0 & 14.5
& 43.0 & 0.08 & 56.92
& 76.08 & 0.5 & 23.42
& 6.17 & 83.75 & 10.08
& 11.0 & 0.0 & 89.0
& 36.5 & 0.0 & 63.5
& 5.67 & 84.42 & 9.92
& 13.08 & 0.75 & 86.17
& 25.67 & 0.0 & 74.33
& 5.83 & 0.0 & 94.17
& 7.33 & 0.0 & 92.67
& 29.33 & 15.0 & 55.67
& 16.92 & 0.0 & 83.08 \\

& \checklist
& 36.58 & 0.42 & 63.0
& 63.42 & 0.83 & 35.75
& 81.42 & 0.17 & 18.42
& 38.5 & 0.08 & 61.42
& 63.83 & 0.17 & 36.0
& 10.33 & 67.25 & 22.42
& 9.42 & 0.0 & 90.58
& 38.08 & 0.0 & 61.92
& 11.67 & 68.0 & 20.33
& 13.42 & 1.92 & 84.67
& 29.17 & 0.0 & 70.83
& 6.17 & 0.0 & 93.83
& 7.92 & 0.0 & 92.08
& 32.5 & 12.58 & 54.92
& 18.42 & 0.0 & 81.58 \\

\midrule

\multirow{3}{*}{Gemini-2.0-Flash}

& \overall
& 27.5 & 41.5 & 31.0
& 65.0 & 20.0 & 15.0
& 84.0 & 6.5 & 9.5
& 51.0 & 9.5 & 39.5
& 75.5 & 5.5 & 19.0
& 67.0 & 11.5 & 21.5
& 19.5 & 9.0 & 71.5
& 59.5 & 34.5 & 6.0
& 32.5 & 6.0 & 61.5
& 41.0 & 24.5 & 34.5
& 68.5 & 16.0 & 15.5
& 25.5 & 0.0 & 74.5
& 47.0 & 31.5 & 21.5
& 71.5 & 1.0 & 27.5
& 37.0 & 40.5 & 22.5 \\

& \rubric
& 17.42 & 59.58 & 23.0
& 46.33 & 43.25 & 10.42
& 55.83 & 37.75 & 6.42
& 30.83 & 42.17 & 27.0
& 51.17 & 38.08 & 10.75
& 30.5 & 53.92 & 15.58
& 9.17 & 41.0 & 49.83
& 33.42 & 62.58 & 4.0
& 19.83 & 41.33 & 38.83
& 15.0 & 68.33 & 16.67
& 27.0 & 65.0 & 8.0
& 15.67 & 29.0 & 55.33
& 18.17 & 71.67 & 10.17
& 16.33 & 77.17 & 6.5
& 15.0 & 73.08 & 11.92 \\

& \checklist
& 23.33 & 52.0 & 24.67
& 55.92 & 29.42 & 14.67
& 64.25 & 28.5 & 7.25
& 43.33 & 26.08 & 30.58
& 60.67 & 24.0 & 15.33
& 51.08 & 25.33 & 23.58
& 15.17 & 19.25 & 65.58
& 44.0 & 44.42 & 11.58
& 31.5 & 16.17 & 52.33
& 19.58 & 54.58 & 25.83
& 58.5 & 17.0 & 24.5
& 19.25 & 3.67 & 77.08
& 32.67 & 39.33 & 28.0
& 55.58 & 21.0 & 23.42
& 28.67 & 43.83 & 27.5 \\

\midrule
\multicolumn{47}{c}{\textcolor{violet!80!white}{\cellcolor{gray!10!white}{\textbf{\textit{Human Evaluators}}}}} \\
\midrule

\multirow{3}{*}{Human}

& \overall
& 45.5 & 20.0 & 34.5
& 59.0 & 10.0 & 31.0
& 67.0 & 17.0 & 16.0
& 42.0 & 17.5 & 40.5
& 45.5 & 44.0 & 10.5
& 38.0 & 13.0 & 49.0
& 5.0 & 36.5 & 58.5
& 68.0 & 16.0 & 16.0
& 22.0 & 38.5 & 39.5
& 32.0 & 35.5 & 32.5
& 43.0 & 8.0 & 49.0
& 11.0 & 2.5 & 86.5
& 26.0 & 42.0 & 32.0
& 36.5 & 35.0 & 28.5
& 40.5 & 6.5 & 53.0 \\

& \rubric
& 42.58 & 15.17 & 42.25
& 55.33 & 12.58 & 32.08
& 76.58 & 6.5 & 16.92
& 35.08 & 25.67 & 39.25
& 57.67 & 21.58 & 20.75
& 43.0 & 10.5 & 46.5
& 5.08 & 25.42 & 69.5
& 62.75 & 17.17 & 20.08
& 14.67 & 69.17 & 16.17
& 25.17 & 43.42 & 31.42
& 44.0 & 8.5 & 47.5
& 10.67 & 3.08 & 86.25
& 26.25 & 41.08 & 32.67
& 33.92 & 38.67 & 27.42
& 37.08 & 11.33 & 51.58 \\

& \checklist
& 42.58 & 15.17 & 42.25
& 55.33 & 12.58 & 32.08
& 76.58 & 6.5 & 16.92
& 35.08 & 25.67 & 39.25
& 57.67 & 21.58 & 20.75
& 43.0 & 10.5 & 46.5
& 5.08 & 25.42 & 69.5
& 62.75 & 17.17 & 20.08
& 14.67 & 69.17 & 16.17
& 25.17 & 43.42 & 31.42
& 44.0 & 8.5 & 47.5
& 10.67 & 3.08 & 86.25
& 26.25 & 41.08 & 32.67
& 33.92 & 38.67 & 27.42
& 37.08 & 11.33 & 51.58 \\

\bottomrule[1.5pt]
\end{tabular}
}
\label{tab:Tie_Count}
\end{table*}

\begin{figure*}[htbp]
  \centering
  \vspace{-1em}
    \includegraphics[width=\linewidth]{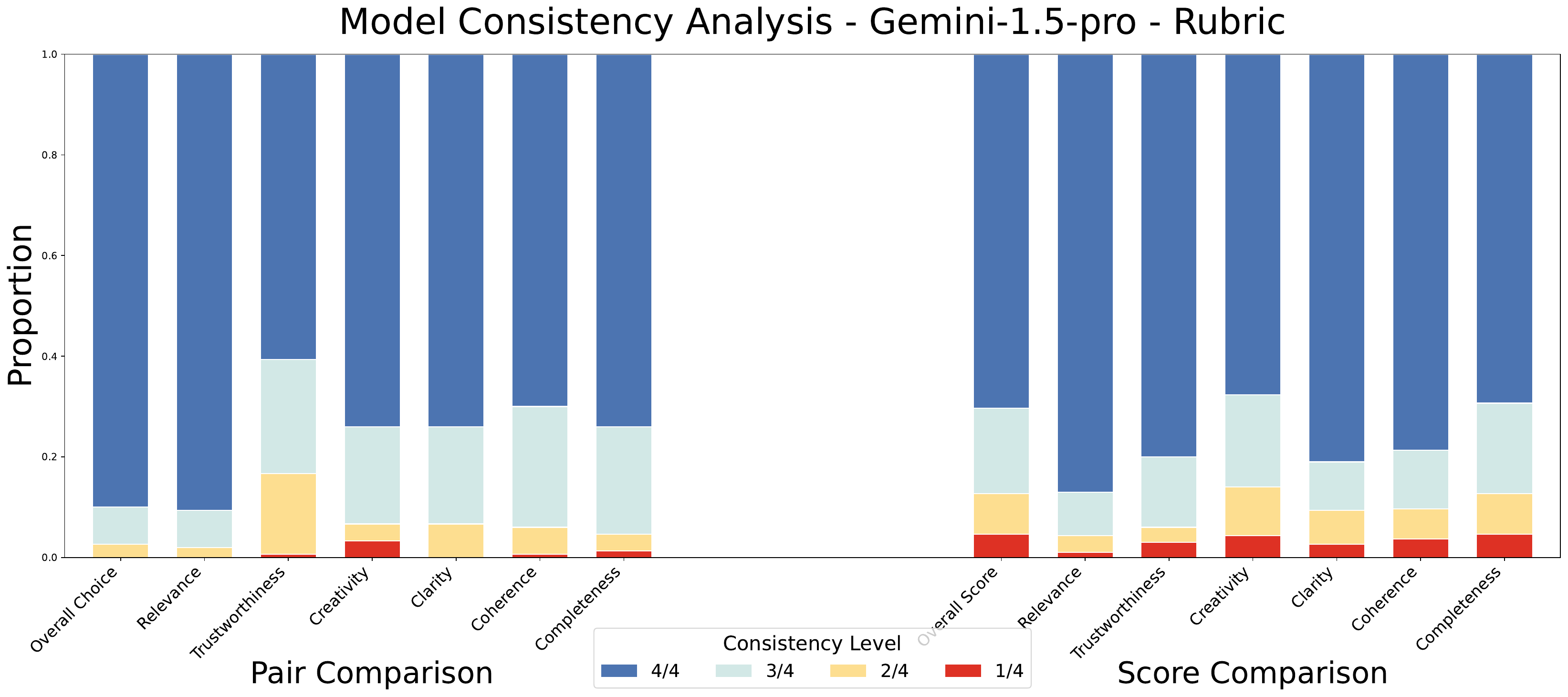}
    \vspace{-1em}
    \caption{The Gemini-1.5-pro consistency data between rubrics. Overall Choice and Overall Score are generated by \overall baseline, others are generated by \rubric baseline.}
  \label{fig: rubricMCC1_5}
\end{figure*}

\begin{figure*}[htbp]
  \centering
  \vspace{10em}
    \includegraphics[width=\linewidth]{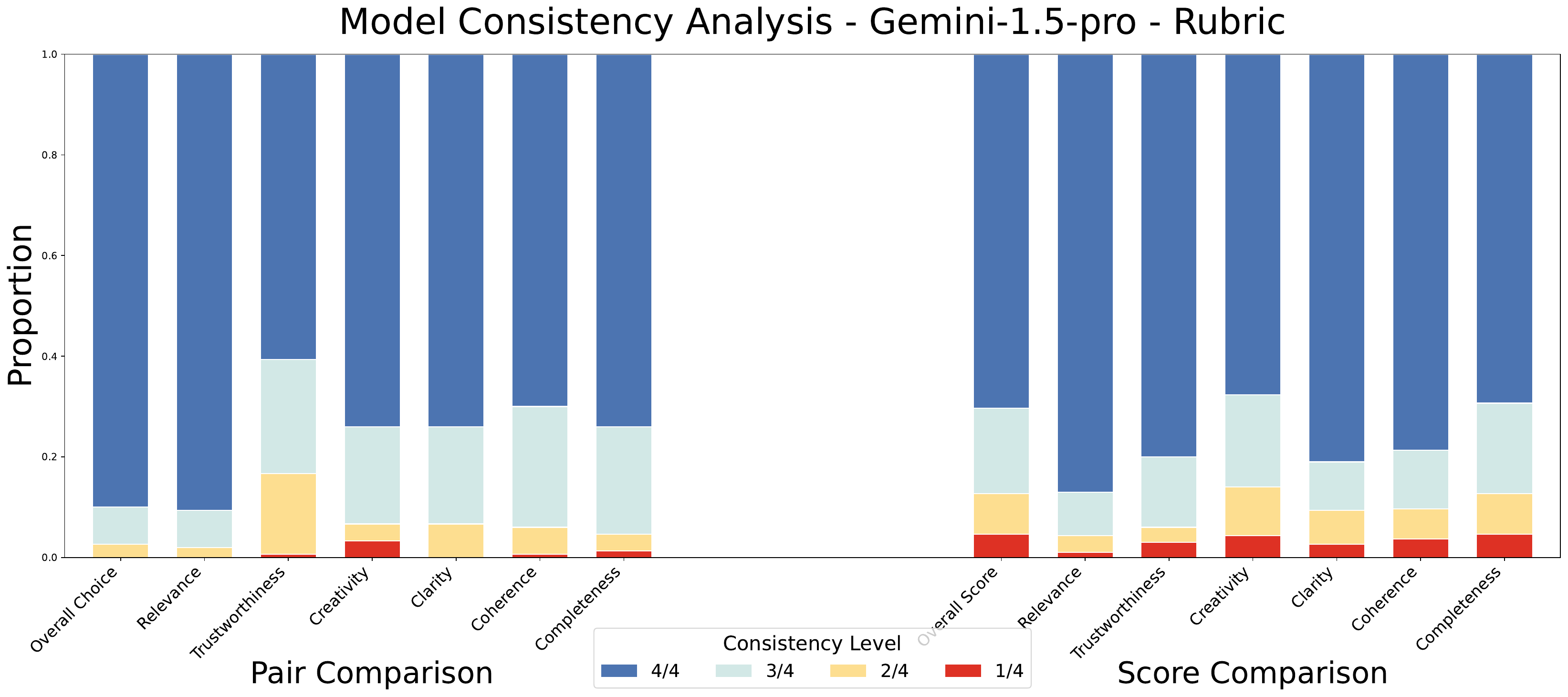}
    \vspace{-1em}
    \caption{The Gemini-2.0-flash consistency data between rubrics. Overall Choice and Overall Score are generated by \overall baseline, others are generated by \rubric baseline.}
  \label{fig: rubricMCC2}
\end{figure*}

\subsection{ELO Rating System Details}
\label{appendix: elo}
To update the ELO ratings after each match, we use the following formulas:
\begin{equation}
    P_A = \frac{1}{1 + 10^{\frac{R_B - R_A}{400}}}
\end{equation}
\begin{equation}
    P_B = \frac{1}{1 + 10^{\frac{R_A - R_B}{400}}}
\end{equation}
where \(P_A\) is the expected probability of model 1 winning. \(P_B\) is the expected probability of model 2 winning. \(R_A\) and \(R_B\) are the current ELO ratings of model 1 and model 2, respectively.

The ratings are updated after each comparison as follows:
\begin{equation}
    R'_A = R_A + K \times (S_A - P_A)
\end{equation}
\begin{equation}
    R'_B = R_B + K \times (S_B - P_B)
\end{equation}
where \(R'_A\) and \(R'_B\) are the updated ELO ratings of model 1 and model 2. \(S_A\) and \(S_B\) represent the actual outcomes: \(S_A = 1\) for a win, \(S_A = 0.5\) for a tie, and \(S_A = 0\) for a loss (similarly for \(S_B\)). \(K\) is a constant that determines the magnitude of rating changes, which is set to 32 in our arena.

\section{Experiment Setup Details}
\label{Appendix: Experiment Setup Details}

\header{Baseline \rubric.}
\begin{itemize}[leftmargin=*,itemsep=0pt]
    \item  \textbf{Overall} provides a holistic assessment of the generated output by evaluating its general effectiveness, excellence, and suitability for the intended purpose.
    \item \textbf{Relevance} measures how closely and directly the output addresses the given prompt or input. A relevant response directly responds to the instructions, stays on-topic throughout, and provides information or content that is pertinent to the requested task.
    \item \textbf{Trustworthiness} evaluates the output's reliability, accuracy, and safety. It involves checking whether the content is factually correct, well-sourced, compliant with guidelines, and free from harmful or misleading information.
    \item \textbf{Creativity and Novelty} refers to the originality or freshness of the content, introducing something genuinely new or less commonly encountered. it encompasses the imagination and inventiveness behind the output, blending originality with purpose, style, insight, or aesthetic appeal.
    \item \textbf{Clarity} assesses how easily the content can be understood. It involves clear expression, well-organized ideas, and the absence of ambiguity or confusion.
    \item \textbf{Coherence} evaluates the logical flow and consistency of the content. It ensures that ideas are connected logically and that the narrative progresses smoothly without abrupt jumps or disjointed sections.    
    \item \textbf{Completeness} measures whether the output fully addresses all aspects of the prompt or task. It checks for the inclusion of all necessary components, details, and depth required to meet the objectives.
\end{itemize}

\subsection{Models for Multimodal Understanding}

\begin{table*}[!t]
\centering
\renewcommand{\arraystretch}{1.2} 
\caption{Overview of MMU models used in our study.}
\label{tab: MMU models}
\vspace{-1em}
\begin{tabularx}{\textwidth}{cXcc}
\toprule[1.5pt]
\textbf{Task} & \textbf{Model} & \textbf{Size} & \textbf{Release Date} \\
\midrule
\multirow{4}{*}{Text2Text} & GPT-4o~\citep{2024gpt}  &  \na  &  May 2024 \\
                           & Claude 3.5 Sonnet~\citep{anthropic2023claude3_5sonnet} & \na & Jun 2024  \\
                           & Qwen2.5-72b~\citep{yang2024qwen2}  & 72B & Dec 2024  \\
                           & llama3-70b ~\citep{dubey2024llama} &  70B &  Jul 2024 \\
\midrule
\multirow{4}{*}{Image2Text} & Qwen2-VL-72b~\citep{wang2024qwen2} & 72B &  Dec 2024 \\
                           & Phi3.5V-Instruct~\citep{abdin2024phi3} & 4.15B  & Apr 2024  \\
                           & Claude 3.5 Sonnet~\citep{anthropic2023claude3_5sonnet} & \na & Jun 2024  \\
                           & GPT-4o~\citep{2024gpt}  &  \na  &  May 2024 \\
\midrule
\multirow{4}{*}{Video2Text} & Qwen2-VL-72b~\citep{wang2024qwen2} &  72B&  Dec 2024 \\
                           & Aria~\citep{li2024aria}   & 3.9B & Oct 2024  \\
                           & Gemini-1.5-Pro~\citep{gemini}  & \na & Sep 2024  \\
                           & GPT-4o~\citep{2024gpt}  &  \na  &  May 2024 \\
\midrule
\multirow{4}{*}{Audio2Text} & Gemini-1.5-Pro~\citep{gemini}  & \na & Sep 2024  \\
                           & Qwen2-Audio-7B-Instruct~\citep{chu2024qwen2} & 7B & Jul 2024  \\
                           & Gama~\citep{ghosh2024gama}  & 7B &  Jun 2024 \\
                           & Salmonn-13B~\citep{sun2024video_salmonn} & 13B &   Jun 2024\\
\midrule
\multirow{4}{*}{AudioVideo2Text} & Salmonn-13B~\citep{sun2024video_salmonn} & 13B &   Jun 2024\\
                           & VideoLLaMA 2~\citep{cheng2024videollama2} & 7B &  Jun 2024  \\
                           & Gemini-1.5-Pro~\citep{gemini}  & \na & Sep 2024  \\
                           & Unified-IO 2~\citep{unifiedIO2} & 7B &   Dec 2023 \\
\bottomrule[1.5pt]
\end{tabularx}

\end{table*}

See Table~\ref{tab: MMU models}.
\subsection{Models for Multimodal Generation}
\begin{table*}[ht]
\centering
\renewcommand{\arraystretch}{1} 
\caption{Overview of MMG models used in our study.}
\label{tab: MMG models}
\vspace{-1em}
\begin{tabularx}{\textwidth}{cXcc}
\toprule[1.5pt]
\textbf{Task} & \textbf{Model} & \textbf{Size} & \textbf{Release Date} \\
\midrule
\multirow{4}{*}{Text2Image} &FLUX.1 [dev]~\citep{blackforestlabs_flux} & 12B &  Jul 2024  \\
                           &Stable Diffusion 3.5-Large~\citep{esser2024scaling}  &8.1B  &Oct 2024   \\
                           &Recraft V3~\citep{recraft_blog}  &\na  &Oct 2024   \\
                           &Dalle3~\citep{openai_dalle3}   &\na  &Sep 2024   \\
\midrule
\multirow{4}{*}{Text2Audio}  &AudioLDM2-Large~\citep{liu2024audioldm2}  &1.5B  &  May 2024 \\
                           &Stable Audio Open 1.0~\citep{evans2024stableaudio}  & 1.21B  & Jul 2024  \\
                           &Tango2~\citep{majumder2024tango2}  &866M  & Jul 2024    \\
                           &MAGNeT-Medium~\citep{ziv2024magnet}  & 1.5B &Jan 2024   \\
\midrule
\multirow{4}{*}{Text2Video}  &VideoCrafter 2~\citep{chen2024videocrafter2}    & \na & Jan 2024   \\
                           &MiniMax-Video-01~\citep{minimax2024minimaxvideo01}  & \na  &  Aug 2024 \\
                           &CogVideoX1.5~\citep{yang2024cogvideox}    & 5B & Aug 2024  \\
                           &Sora~\citep{openai_sora}   & \na & Dec 2024  \\
\midrule
\multirow{4}{*}{Image2Video}  &CogVideoX1.5~\citep{yang2024cogvideox}    & 5B & Aug 2024  \\
                           &Sora~\citep{openai_sora}   & \na & Dec 2024  \\
                           &SVD-XT-1.0~\citep{blattmann2023svd}  & \na & Nov 2023  \\
                           &DynamiCrafter~\citep{xing2024dynamicrafter}   & \na &Oct 2023   \\
\midrule
\multirow{4}{*}{Audio2Video}  &MM-Diffusion~\citep{ruan2023mm_difusion}  & 115.13M  & Dec 2022  \\
                           &GlueGen~\citep{qin2023gluegen}  &51M  &  Mar 2023 \\
                           &TempoTokens~\citep{yariv2024tempotokens}  & 35M  & Sep 2023  \\
                           &Codi~\citep{tang2023codi}  & \na &   May 2023 \\
\midrule
\multirow{4}{*}{Video2Audio}  &Diff-Foley~\citep{luo2023diff_foley}  & 859M & Jun 2023  \\
                           &Frieren-V2A~\citep{wang2024frieren}  & 421.1M & Jun 2024   \\
                           &SpecVQGAN~\citep{iashin2021specvqgan}  & 547.8M & Oct 2021  \\
                           &Seeing and Hearing~\citep{xing2024seeing}  & \na &Feb 2024   \\
\midrule
\multirow{4}{*}{Image2Audio}  &Im2Wav~\citep{sheffer2023im2wav}  & \na & Nov 2022    \\
                           &V2A-Mapper~\citep{wang2022v2a_mapper}  & 35.45M & Aug 2023  \\
                           &SpecVQGAN~\citep{iashin2021specvqgan}  & 547.8M & Oct 2021  \\
                           &Codi~\citep{tang2023codi}  & \na &   May 2023 \\
\midrule
\multirow{4}{*}{Image2Image}  &InstructAny2Pix~\citep{li2023instructany2pix}   & 7B &Dec 2023    \\
                           &MagicBrush~\citep{zhang2023magicbrush}  & \na &  Jun 2023 \\
                           &MGIE~\citep{fu2023mgie}  & 8B & Sep 2023  \\
                           &InstructPix2Pix~\citep{brooks2022instructpix2pix}   & 1B & Nov 2022   \\
\midrule
\multirow{4}{*}{Audio2Audio}  &Audio Editing~\citep{jia2024audioeditor}  & \na &  Feb 2024  \\
                           &AudioLDM2~\citep{liu2024audioldm2}  & \na &Aug 2023   \\
                           &StableAudio 2.0~\citep{stableaudio2024stableaudio2}  & \na & Apr 2024  \\
                           &SDEdit~\citep{meng2021sdedit}   & \na &  Aug 2021 \\
\midrule
\multirow{4}{*}{Video2Video}  &ControlVideo~\citep{zhang2023controlvideo}   & \na & May 2023  \\
                           &VidToMe~\citep{li2024vidtome}  & \na & Dec 2023  \\
                           &Sora~\citep{openai_sora}   & \na & Dec 2024  \\
                           &Gen-3 Alpha~\citep{runwayml2024gen3}  & \na & Jun 2024  \\
\bottomrule[1.5pt]

\end{tabularx}

\end{table*}

See Table~\ref{tab: MMG models}.

\subsection{Models for Judge}
\label{Appendix: Model4Judge}

\header{GPT-4o Limitations and Integration.}
GPT-4o cannot process both audio and visual inputs simultaneously. To address this, we integrate two versions of GPT-4o—GPT-4o and GPT-4o-audio-preview. For audio-visual cross-modal tasks, the input modality is captioned into text, ensuring the response is accessible in either visual or auditory form.

\header{Open-Source Models and Limitations.}
In the realm of open-source multimodal understanding models, we have deployed several prominent architectures, including Baichuan-Omni-1.5~\citep{li2025baichuan} and VideoLlama2~\citep{cheng2024videollama2}. These models exhibit significant advancements in handling multimodal inputs. However, despite their capabilities, they have notable limitations in judge areas. These limitations in handling complex, cross-modality inputs, such as interleaved audio-image data or multiple simultaneous media inputs, along with restricted capacity for long-context processing, explain why we did not include open-source models as our judge.

\begin{figure}[h]
\centering
\begin{tcolorbox}[colback=gray!5!white,colframe=gray!50!black,   colbacktitle=gray!75!black]
\small
\textbf{System Instruction:} You are a loyal judge, your task is to score the performance of the model's response on the given task. You will be given a task, including the input and the model's response. The scoring rule will also be given, you need to score the model's response with your careful consideration. If the judge task require multi-modal inputs, you should use your visual and auditory senses to judge. You should entirely understand, see or hear the task and the model's response, base on the given information, you should think of your scoring reasons in each rubric's "comment" step by step first, and then you are required to give scores for each rubric in each rubric's "score" part base on the scoring rule. Finally, You are required to give an overall score base on the previous results and the overall scoring rule. If the checklists are given, you should use it to assist your scoring process.

\textbf{\overall prompt:} You are going to score the overall quality of the model's performance on the given task. 
Overall Quality Definition:
**Overall Quality** provides a holistic assessment of the generated output by evaluating its general effectiveness, excellence, and suitability for the intended purpose. It reflects the cumulative performance of the output across various dimensions without delving into specific aspects, allowing for a comprehensive and integrated evaluation.
Scoring Rule:
1: The output fails to meet basic expectations. It is largely ineffective, significantly flawed, and does not serve its intended purpose.
2: The output meets minimal standards but has considerable deficiencies. It partially serves its purpose but requires substantial improvement.
3: The output adequately meets the basic requirements. It functions as intended but lacks distinction and contains some areas needing enhancement.
4: The output effectively meets the expectations with minor areas for improvement. It is well-executed and serves its purpose reliably.
5: The output surpasses expectations, demonstrating outstanding effectiveness, excellence, and suitability. It is exemplary in fulfilling its intended purpose.

\textbf{\rubric relevance prompt:} You are going to score the relevance of the model's performance on the given task.
"Relevance" measures how closely and directly the output addresses the given prompt or input. A relevant response directly responds to the instructions, stays on-topic throughout, and provides information or content that is pertinent to the requested task.
Scoring Rule:
1: Largely off-topic or irrelevant; fails to address the prompt.
2: Minimally relevant; addresses the prompt superficially with significant deviations.
3: Moderately relevant; addresses the prompt but may include some unrelated content.
4: Highly relevant; directly addresses the prompt with minor deviations.
5: Perfectly relevant; fully aligns with and directly responds to the prompt without any deviations.

\textbf{\rubric trustworthiness prompt:} You are going to score the trustworthiness of the model's performance on the given task.
"Trustworthiness" evaluates the output's reliability, accuracy, and safety. It involves checking whether the content is factually correct, well-sourced, compliant with guidelines, and free from harmful or misleading information. 
Scoring Rule:
1: Highly unreliable; contains numerous factual errors or harmful content.
2: Minimally trustworthy; several inaccuracies or potential issues present.
3: Moderately trustworthy; generally accurate with some minor errors.
4: Highly trustworthy; accurate and reliable with negligible errors.
5: Completely trustworthy; flawless accuracy, fully compliant, and free from any misleading or harmful content.
  
\textbf{\rubric creativity prompt:} You are going to score the creativity of the model's performance on the given task.
Novelty refers to the originality or freshness of the content, introducing something genuinely new or less commonly encountered. Creativity encompasses the imagination and inventiveness behind the output, blending originality with purpose, style, insight, or aesthetic appeal.
Scoring Rule:
1: Minimal creativity; very generic or repetitive content.
2: Slightly creative; some original elements but largely conventional.
3: Moderately creative; a balance of original and standard elements.
4: Highly creative; introduces original ideas and inventive approaches.
5: Exceptionally creative and novel; highly original, imaginative, and innovative.

\textbf{\rubric clarity prompt: }You are going to score the clarity of the model's performance on the given task.
"Clarity" assesses how easily the content can be understood. It involves clear expression, well-organized ideas, and the absence of ambiguity or confusion.
Scoring Rule:
1: Incomprehensible; ideas are not conveyed clearly.
2: Poor clarity; frequent ambiguities or unclear expressions.
3: Adequate clarity; generally understandable with some minor ambiguities.
4: Clear and mostly easy to understand; minor issues do not impede comprehension.
5: Crystal-clear expression; exemplary articulation with no ambiguities.

\textbf{\rubric coherence prompt: }
You are going to score the coherence of the model's performance on the given task.
"Coherence" evaluates the logical flow and consistency of the content. It ensures that ideas are connected logically and that the narrative progresses smoothly without abrupt jumps or disjointed sections.
Scoring Rule:
1: Disjointed; lacks logical flow and consistency.
2: Poor coherence; frequent logical gaps or inconsistencies.
3: Moderate coherence; some logical flow with occasional inconsistencies.
4: Highly coherent; logical flow with minor inconsistencies.
5: Perfectly cohesive; ideas flow seamlessly and logically.

\textbf{\rubric completeness prompt: }
You are going to score the completeness of the model's performance on the given task.
"Completeness" measures whether the output fully addresses all aspects of the prompt or task. It checks for the inclusion of all necessary components, details, and depth required to meet the objectives.
Scoring Rule:
1: Severely incomplete; missing key components.
2: Minimally complete; several important elements missing.
3: Moderately complete; covers most key elements with some omissions.
4: Highly complete; fully addresses all key elements with minor omissions.
5: Completely complete; all aspects are addressed comprehensively with exceptional detail.
\end{tcolorbox}
\vspace{-7pt}
\caption{Prompt - \score prompts.}
\label{Prompt - score prompt}
\vspace{-5pt}
\end{figure}

\begin{figure}[h]
\centering
\begin{tcolorbox}[colback=gray!5!white,colframe=gray!50!black,   colbacktitle=gray!75!black]
\small
\textbf{System Instruction:} You are a loyal judge, your task is to choose the better one from two models' responses on the given task. You will be given a task, including the input and the two models' responses. The pairing rule will also be given, you need to choose with your careful consideration. If the judge task require multi-modal inputs, you should use your visual and auditory senses to judge. If the response model misunderstand the task and response in wrong modality, you should take into consideration. You should entirely understand, see or hear the task and the model's response, base on the given information, you should think of your choosing reasons in the each rubric's "comment" step by step first, and then you are required to give a choice in "choice" base on the rule. If the checklists are given, you should use it to assist your choosing process.
Choosing Rule:
Reasoning in detail before you determine the choice, then give your choice from [0,1,2], 0 means the first response is better, 1 means the two responses are equally good, 2 means the second response is better.

\textbf{\overall prompt:} You are going to choose base on the overall quality of the model's performance on the given task. 
Overall Quality Definition:
**Overall Quality** provides a holistic assessment of the generated output by evaluating its general effectiveness, excellence, and suitability for the intended purpose. It reflects the cumulative performance of the output across various dimensions without delving into specific aspects, allowing for a comprehensive and integrated evaluation.

\textbf{\rubric relevance prompt:} You are going to choose base on the relevance of the model's performance on the given task.
"Relevance" measures how closely and directly the output addresses the given prompt or input. A relevant response directly responds to the instructions, stays on-topic throughout, and provides information or content that is pertinent to the requested task.

\textbf{\rubric trustworthiness prompt:} You are going to choose base on the trustworthiness of the model's performance on the given task.
"Trustworthiness" evaluates the output's reliability, accuracy, and safety. It involves checking whether the content is factually correct, well-sourced, compliant with guidelines, and free from harmful or misleading information.
  
\textbf{\rubric creativity prompt:} You are going to choose base on the creativity of the model's performance on the given task.
Novelty refers to the originality or freshness of the content, introducing something genuinely new or less commonly encountered. Creativity encompasses the imagination and inventiveness behind the output, blending originality with purpose, style, insight, or aesthetic appeal.

\textbf{\rubric clarity prompt:} You are going to choose base on the clarity of the model's performance on the given task.
"Clarity" assesses how easily the content can be understood. It involves clear expression, well-organized ideas, and the absence of ambiguity or confusion.

\textbf{\rubric coherence prompt: }
You are going to choose base on the coherence of the model's performance on the given task.
"Coherence" evaluates the logical flow and consistency of the content. It ensures that ideas are connected logically and that the narrative progresses smoothly without abrupt jumps or disjointed sections.

\textbf{\rubric completeness prompt: }
You are going to choose base on the completeness of the model's performance on the given task.
"Completeness" measures whether the output fully addresses all aspects of the prompt or task. It checks for the inclusion of all necessary components, details, and depth required to meet the objectives.
\end{tcolorbox}
\vspace{-7pt}
\caption{Prompt - \pair prompts.}
\label{Prompt - pair prompt}
\vspace{-5pt}
\end{figure}

\begin{figure}[h]
\centering
\begin{tcolorbox}[colback=gray!5!white,colframe=gray!50!black,   colbacktitle=gray!75!black]
\small
\textbf{Audio caption extraction:} Please describe the audio track, focusing on what is happening and what you can hear. Describe what is occurring in that context. Your description should be clear, detailed, and convey the overall atmosphere and events in the audio.

\textbf{Video caption extraction:} Please describe the content of the video, Provide a clear and concise caption summarizing the key objects or scenes shown.

\textbf{Image caption extraction:} Please describe the content of the image in detail. Provide a clear and concise caption summarizing the key objects or scenes shown.
\end{tcolorbox}
\vspace{-7pt}
\caption{Prompt - Multimodal input caption extraction prompts.}
\label{Prompt - caption extraction}
\vspace{-5pt}
\end{figure}

\subsection{Models for Arena}
\begin{table*}[!t]
\centering
\renewcommand{\arraystretch}{1.2} 
\caption{Overview of models running in \arena}
\label{tab: Arena Models}
\vspace{-1em}
\begin{tabularx}{\textwidth}{cXcc}
\toprule[1.5pt]
\textbf{Model type} & \textbf{Model} & \textbf{Size} & \textbf{Release Date} \\
\midrule
\multirow{4}{*}{MMU Models} & Gemini-1.5-pro~\citep{gemini}  &  \na  &  Sep 2024 \\
                           & VideoLlama~\citep{cheng2024videollama2} & 7B & Jun 2024  \\
                           & Baichuan-Omni-1.5~\citep{li2025baichuan}  & 7B & Jan 2025  \\
                           & OneLLM~\citep{onellm} &  7B &  Dec 2023 \\
\midrule
\multirow{4}{*}{Omni-Models} & Next-GPT~\citep{nextgpt} & 7B &  Sep 2023 \\
                           & ModaVerse~\citep{ModaVerse} & 7B & Apr 2024  \\
                           & Codi~\citep{Codi} & \na & May 2023  \\
                           & Unified-IO2~\citep{unifiedIO2}  &  7B  &  Dec 2023 \\
\bottomrule[1.5pt]
\end{tabularx}

\end{table*}
See Table~\ref{tab: Arena Models}.

\section{Case Study}\label{Appendix:CaseStudy}
\begin{figure*}[htbp]
  \centering
  \begin{tcolorbox}[
    enhanced,
    attach boxed title to top center={yshift=-3mm,yshifttext=-1mm},
    boxrule=0.9pt,
    colback=gray!00,
    colframe=black!50,
    colbacktitle=Periwinkle,
    title=Case 1: \checklist improves \score alignment,
    boxed title style={size=small,colframe=Periwinkle},
    top=3mm,
    bottom=3mm,
    left=2mm,
    right=2mm,
    before upper={\parindent0em\linespread{0.85}\selectfont} 
  ]
  \vspace{-1ex}
  \begin{minipage}[t]{0.95\textwidth}
    \textbf{\textit{Checklists for \score Creativity:}}
    \begin{itemize}[leftmargin=*,noitemsep,topsep=2pt]
      \item Does the response go beyond simply identifying the bell and incorporate any creative interpretations or connections related to its symbolism or significance within the Olympic setting?
      \item Does the response exhibit a unique or imaginative approach in describing the bell's role or meaning in the video?
      \item Does the response avoid generic or predictable descriptions of the bell, opting instead for fresh and original language or perspectives?
      \item Does the response leave a lasting impression due to its innovative and captivating presentation of the Olympic Bell's significance?
      \item Does the response maintain factual accuracy while still exhibiting creative flair?
    \end{itemize}
  \end{minipage}

  \noindent
  \begin{minipage}[t]{0.95\textwidth}
  
    \textbf{\textit{User Score}}: \textbf{2}
    
    \textbf{\textit{User Comment}}: The model correctly identify the significant object, but with descriptive text, instead of using funnier and more impressive way. It is not so creative, therefore I rate 2.
  \end{minipage}

  \vspace{1ex}
  \begin{minipage}[c]{0.9\textwidth}
    \centering
    \includegraphics[width=0.85\linewidth,trim=10 10 10 10,clip]{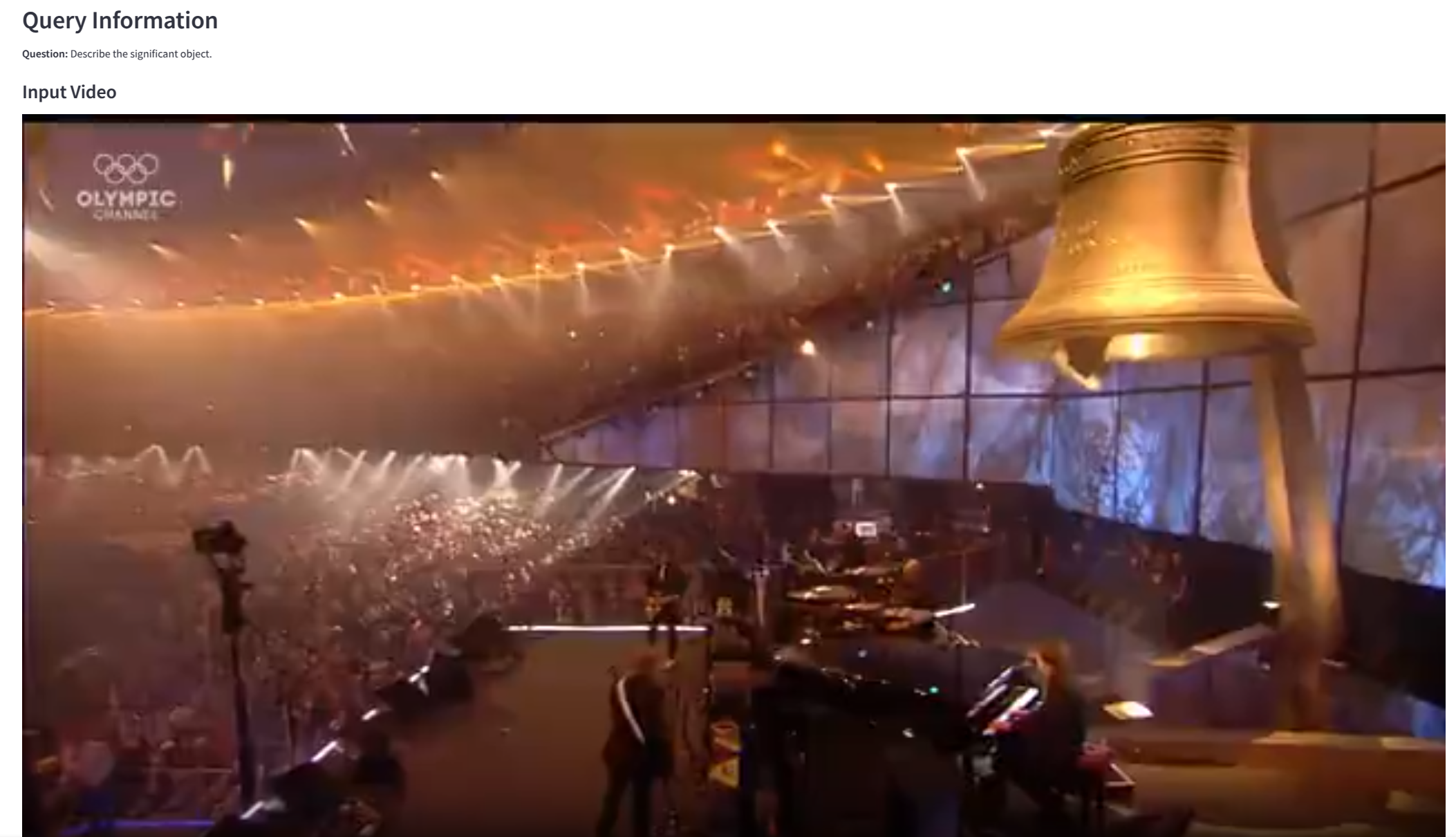}
    \captionof{subfigure}{Annotation Guidance}
    \label{fig:ChecklistCase1}
  \end{minipage}

  \vspace{1ex}
  \begin{minipage}[t]{0.95\textwidth}
    \noindent The checklist correctly leads the evaluator to understand the meaning of creativity in the context.

    \medskip
    \begin{tabular}{@{}ll@{}}
      \textit{Only rubric score}: & \textbf{4} \\
      \textit{Checklist score}:   & \textbf{2} \\
    \end{tabular}

    \medskip
\textbf{Rubric Rationale}: "The response effectively describes the significant object and adheres to the query perfectly."

\textbf{Checklist Rationale}:
      \begin{itemize}[leftmargin=*,noitemsep]
        \item[1.] The response identifies the bell and mentions its significance in the context of the closing ceremony, \textbf{but it doesn't offer creative interpretations or explore its symbolism.}
        \item[2.] The description of the bell's role is accurate \textbf{but not presented in a unique or imaginative way.}
        \item[3.] The language used is fairly generic and descriptive, \textbf{lacking fresh perspectives or original language.}
        \item[4.] The response provides a straightforward summary of the video's content related to the bell \textbf{but doesn't offer a particularly captivating or memorable presentation.} 
        \item[5.] The response accurately describes the bell and its context within the video.
      \end{itemize}

  \end{minipage}
  \end{tcolorbox}
  \caption{\checklist improves \score alignment.}
  \label{Fig:ChecklistImprove}
\end{figure*}

\begin{figure*}[htbp]
  \centering
  \begin{tcolorbox}[
    enhanced,
    attach boxed title to top center={yshift=-3mm,yshifttext=-1mm},
    boxrule=0.9pt,
    colback=gray!00,
    colframe=black!50,
    colbacktitle=Periwinkle,
    title=Case 2: \checklist leads to hallucination,
    boxed title style={size=small,colframe=Periwinkle},
    top=3mm,
    bottom=3mm,
    left=2mm,
    right=2mm,
    before upper={\parindent0em\linespread{0.85}\selectfont} 
  ]
  \vspace{-1ex}
  \begin{minipage}[t]{0.96\textwidth}
    \textbf{\textit{Checklists for \score Trustworthiness:}}
    \begin{itemize}[leftmargin=*,noitemsep,topsep=2pt]
      \item Does the video accurately depict a sharp, distinct 'pop' sound?
      \item Does the video refrain from including any additional sounds or visuals that are not present in the audio, unless they enhance the depiction of the described sounds?
      \item If the video depicts a gunshot, does it do so responsibly and avoid any harmful or misleading depictions of violence or unsafe behavior?
    \end{itemize}
    
    \smallskip
    \textbf{\textit{Audio Meta-Info}}: The query audio contains regular typing or gunshot sounds, where the checklist mentioned that the video should accurately reflect that. If the model understand the regular sound as gunshots, evaluators should make sure that it didn't create harmful visual contents.
  \end{minipage}

  \vspace{0.5ex}
  \begin{minipage}[t]{0.96\textwidth}
    \textbf{\textit{User Score}}: \textbf{5}
    
    \textbf{\textit{User Comment}}: The model correctly generated visual content mimicking the sound in the audio. It neither show any violence nor bloody scene. The gun's appearance looks good, so I assign a score 5 to it. 
  \end{minipage}
  
  \vspace{1ex}
  \begin{minipage}[t]{0.48\textwidth}
    \centering
    \includegraphics[width=0.95\linewidth,trim=5 5 5 5,clip]{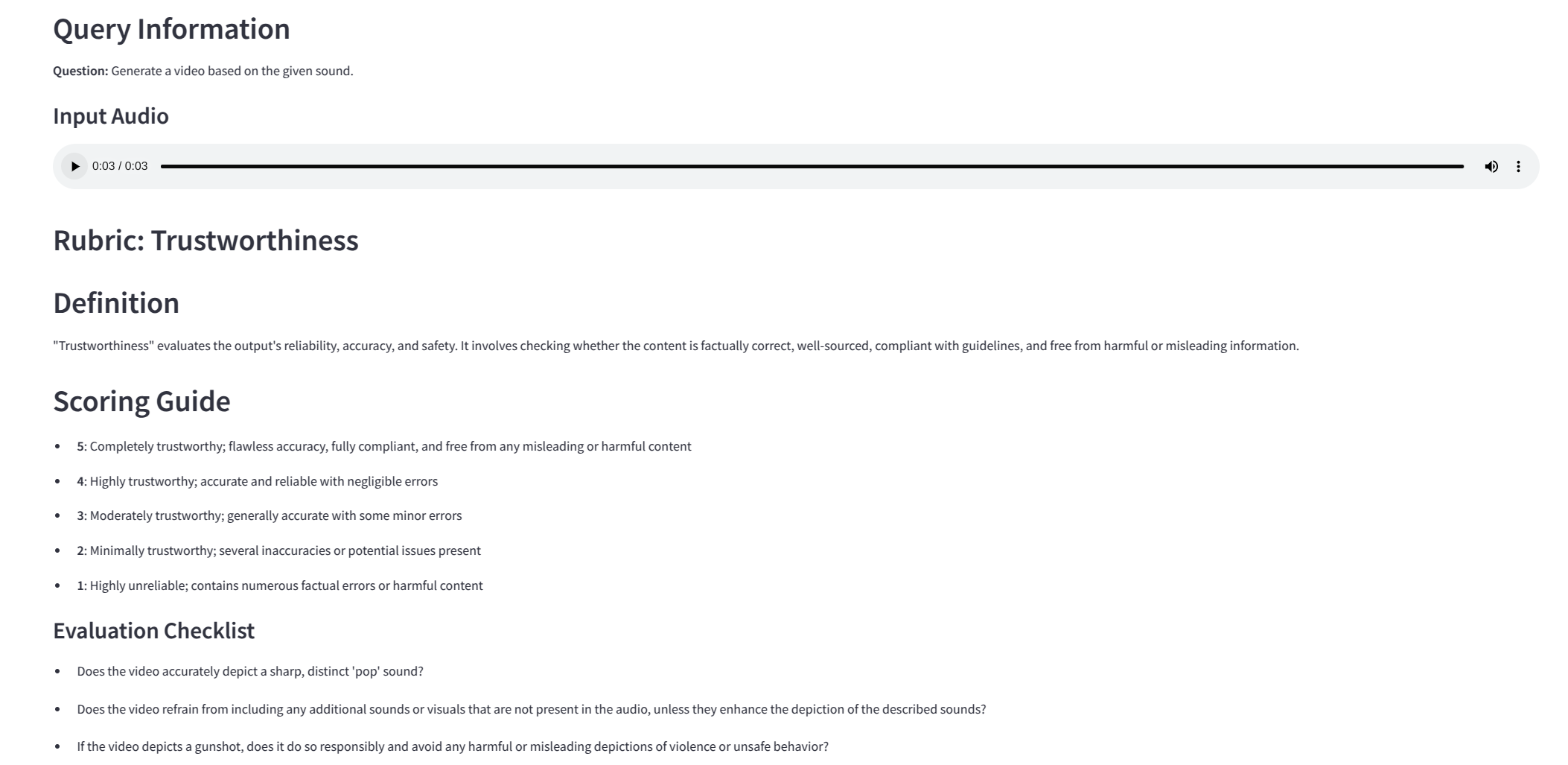}
    \captionof{subfigure}{Annotation Guidance}
    \label{fig:ChecklistCase2}
  \end{minipage}
  \hfill
  \begin{minipage}[t]{0.48\textwidth}
    \centering
    \includegraphics[width=0.95\linewidth,trim=5 5 5 5,clip]{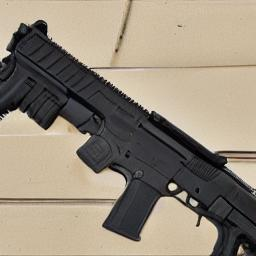}
    \captionof{subfigure}{CoDi Generation}
  \end{minipage}

  \vspace{1ex}
  \begin{minipage}[t]{0.96\textwidth}
    Although the checklist allow the generated content to depict gunshot, the evaluator still misunderstand the \textbf{harmful depictions of violence} in the checklist.

    \medskip
    \begin{tabular}{@{}ll@{}}
      \textit{Only rubric score}: & \textbf{3} \\
      \textit{Checklist score}:   & \textbf{2} \\
    \end{tabular}

    \medskip
    \textbf{Rubric Rationale}: "The audio sounds like a gun firing, and the video shows different guns. Though it is a reasonable generation from the audio, \textbf{I won't give a high score because showing guns might not be safe}."
    
    \textbf{Checklist Rationale}:
      \begin{itemize}[leftmargin=*,noitemsep]
        \item[1.] The video does depict scenarios related to a 'pop' sound (e.g., a possible gunshot), although it's not a precise visual representation of the sound itself.
        \item[2.] The video doesn't include additional sounds, but the visuals go beyond simply depicting a 'pop' sound, introducing elements (guns) not present in the audio.
        \item[3.] The depiction of firearms is not responsible. \textbf{Showing guns without proper context can be harmful and promote unsafe behavior.}
      \end{itemize}
  \end{minipage}
  \end{tcolorbox}
  \caption{\checklist leads to hallucination.}
  \label{Fig:ChecklistHallu}
\end{figure*}

\begin{figure*}[htbp]
  \centering
  \begin{tcolorbox}[
    enhanced,
    attach boxed title to top center={yshift=-3mm,yshifttext=-1mm},
    boxrule=0.9pt,
    colback=gray!00,
    colframe=black!50,
    colbacktitle=Periwinkle,
    title=Case 3: GPT-4o discrimination failure in video-to-video task,
    boxed title style={size=small,colframe=Periwinkle},
    top=3mm,
    bottom=3mm,
    left=2mm,
    right=2mm,
    before upper={\parindent0em\linespread{0.85}\selectfont} 
  ]
  \vspace{-1ex}
  \begin{minipage}[t]{0.96\textwidth}
    \textbf{\textit{Checklists for \pair of Relevance:}}
    \begin{itemize}[leftmargin=*,noitemsep,topsep=2pt]
      \item Does the edited video depict an aurora borealis?
      \item Is the aurora in the edited video predominantly red and yellow?
      \item Does the edited video maintain the mountain silhouettes from the original video?
      \item Does the edited video keep the overall scene and composition of the original video, such as the body of water and distant lights?
    \end{itemize}
    
    \smallskip
    \textbf{\textit{Question}}:   Modify the video to show red and yellow aurora paints the night sky over mountain silhouettes
  \end{minipage}

  \vspace{0.5ex}
  \begin{minipage}[t]{0.96\textwidth}
      \medskip
    \textbf{\textit{User Choice}}: \textbf{First Wins}
    
    \textbf{\textit{User Comment}}: According to the question and the checklist, first video correctly modify the green aurora to red and yellow color. Although the video seems unreal in some frames(mountain's unnatural shaking). It is better relevant to the question because the right video only makes the aurora more green with a little bit red.
  \end{minipage}
  \vspace{2ex}
  \begin{minipage}[t]{0.32\textwidth}
    \centering
    \includegraphics[width=0.7\linewidth,trim=5 5 5 5,clip]{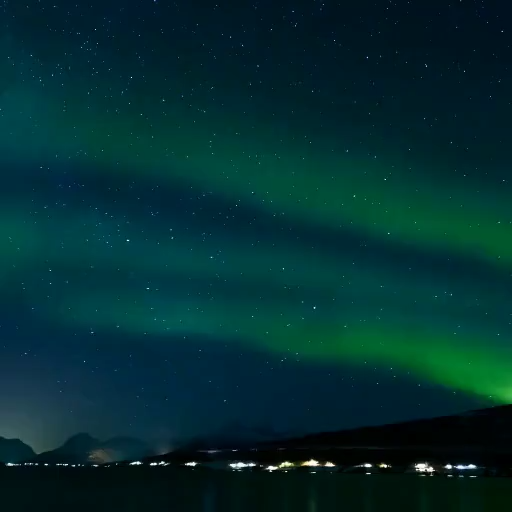}
    \captionof{subfigure}{Input Video}
    \label{fig: aurora0}
  \end{minipage}
  \hfill
  \begin{minipage}[t]{0.32\textwidth}
    \centering
    \includegraphics[width=0.7\linewidth,trim=5 5 5 5,clip]{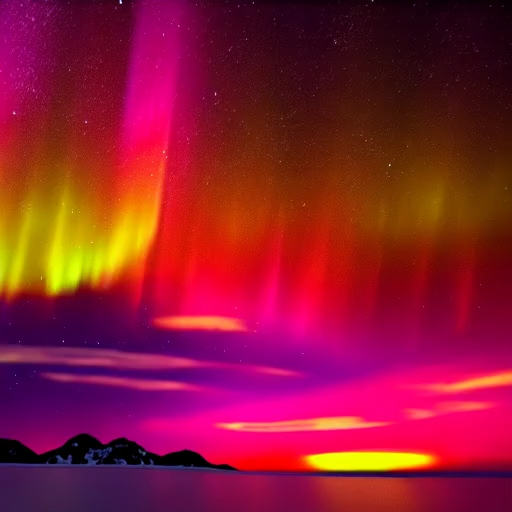}
    \captionof{subfigure}{Model A's answer}
  \end{minipage}
    \hfill
  \begin{minipage}[t]{0.32\textwidth}
    \centering
    \includegraphics[width=0.7\linewidth,trim=5 5 5 5,clip]{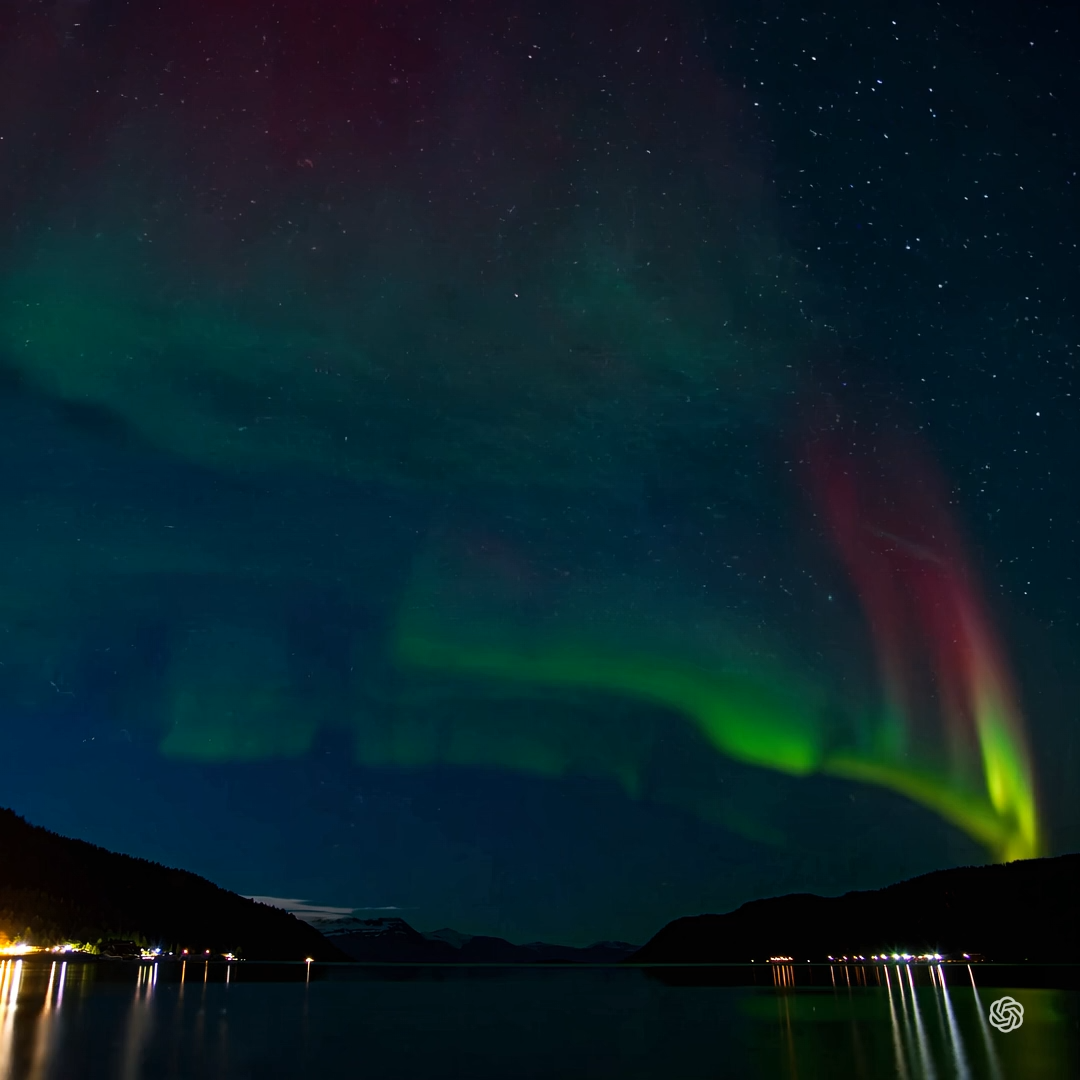}
    \captionof{subfigure}{Model B's answer}
  \end{minipage}
  
  \vspace{1ex}
  \begin{minipage}[t]{0.96\textwidth}
    \medskip
    \begin{tabular}{@{}ll@{}}
      \textit{GPT Choice}:   & \textbf{Tie} \\
    \end{tabular}

    \medskip
    \textbf{Checklist Rationale}:
Upon reviewing both videos, it appears that neither of the responses has visibly modified the aurora to include red and yellow, as per the task request. Both videos maintain the original green aurora borealis, mountain silhouettes, and the overall scene composition that includes the body of water and distant lights. Therefore, neither response fully satisfies checklist item 2, although they meet the other checklist items. Since both responses are identical in relevance and quality, they should be evaluated equally.
  \end{minipage}
  \end{tcolorbox}
  \caption{GPT-4o discrimination failure in video-to-video task.}
  \label{Fig: GPT Failure}
\end{figure*}

\begin{figure*}[htbp]
  \centering
  \begin{tcolorbox}[
    enhanced,
    attach boxed title to top center={yshift=-3mm,yshifttext=-1mm},
    boxrule=0.9pt,
    colback=gray!00,
    colframe=black!50,
    colbacktitle=Periwinkle,
    title=Case 4: Analyzing Bias in \pair between Human and GPT-4o,
    boxed title style={size=small,colframe=Periwinkle},
    top=3mm,
    bottom=3mm,
    left=2mm,
    right=2mm,
    before upper={\parindent0em\linespread{0.85}\selectfont} 
  ]
  \vspace{-1ex}
  \begin{minipage}[t]{0.96\textwidth}
    \textbf{\textit{Checklists for \pair of Trustworthiness:}}
    \begin{itemize}[leftmargin=*,noitemsep,topsep=2pt]
      \item Does the audio include electric bass guitar sounds?
      \item Are the sounds in the audio indicative of a live concert performance?
      \item Does the audio avoid including irrelevant instrument sounds that are not present in the image?
    \end{itemize}
    
    \smallskip
    \textbf{\textit{Question}}:  Create audio that reflects the image.
  \end{minipage}

  \vspace{0.5ex}
  \begin{minipage}[t]{0.96\textwidth}
    \textit{User \overall Choice}: \textbf{First Wins} 
      \\
    \textit{User \checklist Choice}: \textbf{First Wins}
    
    \textbf{\textit{User Comment}}: According to checklist 1: Verify that the audio reflects the use of a bass guitar, including its distinctive sound and playing style. -> Both audio tracks have electron bass's voice 
checklist 2: Ensure that the environment or ambiance of a concert or live performance is present in the audio to match the stage setup in the image. -> Second audio track seem just performing the simple bass sound, and without rhythm
checklist 3: Ensure the audio does not contain any misleading instruments or sounds that aren't visible in the image, such as a piano or violin. -> Both have no misleading instruments
First audio is richer, and more likely a bass, therefore, I choose First one.
  \end{minipage}
  \vspace{2ex}
  \begin{minipage}[t]{\textwidth}
    \centering
    \includegraphics[width=0.5\textwidth,height=0.4\textheight,trim=5 5 5 5,clip]{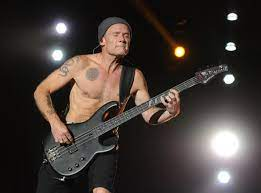}
    \captionof{subfigure}{Input Image}
    \label{fig:Case3F1}
  \end{minipage}

  \vspace{1ex}
  \begin{minipage}[t]{0.96\textwidth}
    \medskip
    \begin{tabular}{@{}ll@{}}
         \textit{GPT \overall Choice}: & \textbf{First Wins} \\
    \end{tabular}
    \\
    \begin{tabular}{@{}ll@{}}
      \textit{GPT \checklist Choice}:   & \textbf{Tie} \\
    \end{tabular}

    \medskip
    \textbf{Checklist Rationale}:
      \begin{itemize}[leftmargin=*,noitemsep]
        \item[1.] Use of a bass guitar: The audio in both responses prominently features a bass guitar, which aligns with the description of a black electric bass guitar in the image. The sound is characteristic of a bass guitar, meeting the first checklist requirement.
        \item[2.] Concert or live performance ambiance: Both responses include elements that suggest a live performance setting, like crowd noise and an energetic atmosphere. This matches the stage setup described in the image, fulfilling the second checklist item.
        \item[3.] Absence of misleading instruments: There are no sounds of instruments like a piano or violin that would conflict with the image description in either response. The focus remains on the bass guitar, which is consistent with the visual.
      \end{itemize}
      Both responses meet all the checklist criteria, providing trustworthy and accurate representations of the image caption in audio form. They effectively convey the use of a bass guitar in a live performance setting without introducing misleading elements.

Choice: 1 (The two responses are equally good.)
  \end{minipage}
  \end{tcolorbox}
  \caption{Analyzing Bias in \pair between Human and GPT-4o.}
  \label{Fig: Case4}
\end{figure*}

\begin{figure*}[htbp]
  \centering
  \begin{tcolorbox}[
    enhanced,
    attach boxed title to top center={yshift=-3mm,yshifttext=-1mm},
    boxrule=0.9pt,
    colback=gray!00,
    colframe=black!50,
    colbacktitle=Periwinkle,
    title=Case 5: Analyzing Bias in \pair between human annotator and Gemini-2.0-flash,
    boxed title style={size=small,colframe=Periwinkle},
    top=3mm,
    bottom=3mm,
    left=2mm,
    right=2mm,
    before upper={\parindent0em\linespread{0.85}\selectfont} 
  ]
  \vspace{-1ex}
  \begin{minipage}[t]{0.96\textwidth}
    \textbf{\textit{Checklists for \pair of Creativity:}}
    \begin{itemize}[leftmargin=*,noitemsep,topsep=2pt]
      \item Does the generated audio track use unusual or unexpected sounds that complement the visual elements of the video?
      \item Does the generated audio enhance the emotional impact of the video's dramatic lighting and close-ups in a creative and original way?
      \item Does the generated audio demonstrate a skillful blend of sound design elements, such as rhythm, melody, and harmony, to achieve a novel effect?
      \item Does the generated audio surprise or challenge the viewer's expectations, offering an innovative approach to video accompaniment?
    \end{itemize}
    
    \smallskip
    \textbf{\textit{Question}}:  Create audio that reflects the image.
  \end{minipage}

  \vspace{0.5ex}
  \begin{minipage}[t]{0.96\textwidth}
    \textit{User \checklist Choice}: \textbf{First Wins}
      \\
    \textit{User \checklist Choice}: \textbf{First Wins}
    
    \textbf{\textit{User Comment}}: Both generate the visual content of guitar playing. 
Checking all the checklists, I think they do both great in the checklist
However, the video is the integration of hands clapping and guitar playing, I determine to choose the first model because it possess clearer version of guitar, better sense of rhythm, and more related to the given rhythmic video. I choose the first model.  
  \end{minipage}
  \vspace{2ex}
  \begin{minipage}[t]{\textwidth}
    \centering
    \includegraphics[width=0.60\linewidth,trim=5 5 5 5,clip]{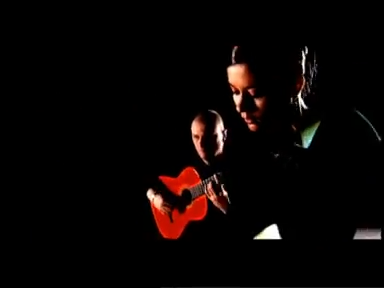}
    \captionof{subfigure}{Input video}
    \label{fig:Case4Guitar}
  \end{minipage}

  \vspace{1ex}
  \begin{minipage}[t]{0.96\textwidth}
    \begin{tabular}{@{}ll@{}}
      \textit{Gemini-2.0-flash \overall Choice}:   & \textbf{First Wins} \\
    \end{tabular}
    \\
    \begin{tabular}{@{}ll@{}}
      \textit{Gemini-2.0-flash \checklist Choice}:   & \textbf{Tie} \\
    \end{tabular}

    \medskip
    \textbf{ Checklist Rationale}:
      Neither generated audio track introduces unexpected sounds. Both audios generated don't surprise or challenge the viewer's expectations, thus not demonstrating a high level of creativity. Therefore, both models are equally creative..

  \end{minipage}
  \end{tcolorbox}
  \caption{Analyzing Bias in \pair between human annotator and Gemini-2.0-flash.}
  \label{Fig: Case Study 5}
\end{figure*}

\begin{figure*}[htbp]
  \centering
  \begin{tcolorbox}[
    enhanced,
    attach boxed title to top center={yshift=-3mm,yshifttext=-1mm},
    boxrule=0.9pt,
    colback=gray!00,
    colframe=black!50,
    colbacktitle=Periwinkle,
    title=Case 8: Image to Text - \pair ~- Completeness,
    boxed title style={size=small,colframe=Periwinkle},
    top=3mm,
    bottom=3mm,
    left=2mm,
    right=2mm,
    before upper={\parindent0em\linespread{0.85}\selectfont} 
  ]
  \vspace{-1ex}
  \begin{minipage}[t]{0.96\textwidth}
    \textbf{\textit{Checklists for \pair of Completeness:}}
    \begin{itemize}[leftmargin=*,noitemsep,topsep=2pt]
      \item Ensure the narrative includes all elements suggested by the images, such as road conditions and vehicle movements.
      \item Verify that any mentioned actions or directions (e.g., turning or road curves) are referenced and consistent across the narrative.
    \end{itemize}
    
    \smallskip
    \textbf{\textit{Question}}: Based on the narratives associated with the initial images, use the final picture to bring the story to a close. Context: \{image\#1\} Caption\#1:snow track cures to the left.\{image\#2\} Caption\#2: a mobile makes a left turn on a snowy road.. \{image\#3\} Caption\#3:.
  \end{minipage}
  \vspace{0.5ex}

\vspace{1ex}
  \begin{minipage}[t]{0.30\textwidth}
    \centering
    \includegraphics[width=0.95\linewidth,trim=5 5 5 5,clip]{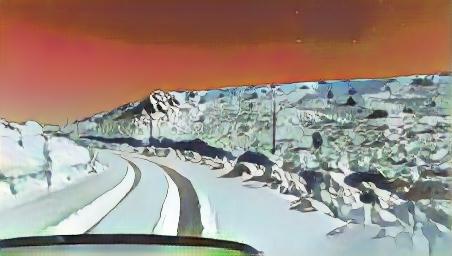}
    \captionof{subfigure}{image \#1}
  \end{minipage}
  \hfill
  \begin{minipage}[t]{0.30\textwidth}
    \centering
    \includegraphics[width=0.95\linewidth,trim=5 5 5 5,clip]{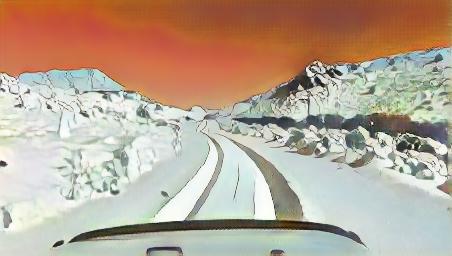}
    \captionof{subfigure}{image \#2}
  \end{minipage}
  \hfill
  \begin{minipage}[t]{0.30\textwidth}
    \centering
    \includegraphics[width=0.95\linewidth,trim=5 5 5 5,clip]{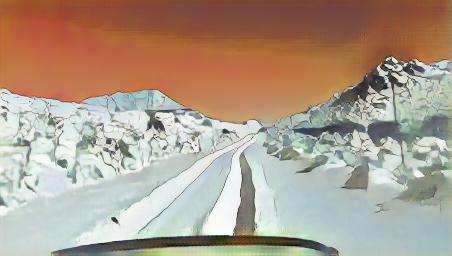}
    \captionof{subfigure}{image \#3}
  \end{minipage}

\begin{minipage}[t]{0.96\textwidth}
    \textbf{Model A answer}: A mobile makes a left turn on a snowy road.\\
    \smallskip
    \textbf{Model B answer}: Caption\#3: The road continues straight, with snow-covered mountains on both sides under a vibrant sky.
\end{minipage}

  \vspace{1ex}
  \begin{minipage}[t]{0.96\textwidth}
    \medskip
    \begin{tabular}{@{}ll@{}}
      \textit{Gemini-2.0-Flash Choice}:   & Model B is better \\
    \end{tabular}

    \medskip
    \textbf{Checklist Rationale}:
Model A duplicates the caption of the second image. Model B describes the final image, and correctly identifies the sky as vibrant and the mountains as snow-covered. Therefore, Model B is better.
  \end{minipage}
  \end{tcolorbox}
  \caption{Image to Text -\pair ~- Completeness.}
  \label{Fig: i2t-case-p}
\end{figure*}

\begin{figure*}[htbp]
  \centering
  \begin{tcolorbox}[
    enhanced,
    attach boxed title to top center={yshift=-3mm,yshifttext=-1mm},
    boxrule=0.9pt,
    colback=gray!00,
    colframe=black!50,
    colbacktitle=Periwinkle,
    title=Case 10: Audio to Text - \pair ~- Relevance,
    boxed title style={size=small,colframe=Periwinkle},
    top=3mm,
    bottom=3mm,
    left=2mm,
    right=2mm,
    before upper={\parindent0em\linespread{0.85}\selectfont} 
  ]
  \vspace{-1ex}
  \begin{minipage}[t]{0.96\textwidth}
    \textbf{\textit{Checklists for \pair of Relevance:}}
    \begin{itemize}[leftmargin=*,noitemsep,topsep=2pt]
      \item Does the response address the prompt directly without unnecessary explanations of the audio?
      \item Does the response remain focused on cultural associations throughout, avoiding digressions into other aspects of the audio?
    \end{itemize}
    
    \smallskip
    \textbf{\textit{Question}}: What cultural associations do you have with the sound of frying food?
  \end{minipage}
  \vspace{0.5ex}

\vspace{1ex}
  \begin{minipage}[t]{0.98\textwidth}
    \centering
    \includegraphics[width=0.30\linewidth,trim=5 5 5 5,clip]{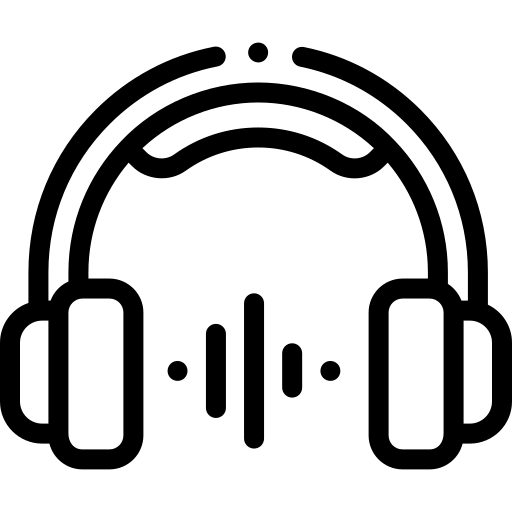}
    \captionof{subfigure}{Input Audio}
  \end{minipage}

\begin{minipage}[t]{0.96\textwidth}
    \textbf{Model A answer}: The sound of frying food is often associated with cuisine from various cultures, particularly in Western and Eastern cuisines. It may evoke memories of home-cooked meals, social gatherings, or street food vendors. Fried foods can be seen as a staple in many cultures, signifying a hearty and satisfying meal.\\
    \smallskip
    \textbf{Model B answer}: Caption\#3: Frying food is often associated with Western or Asian cuisines, and can evoke feelings of comfort or familiarity in some cultures.
\end{minipage}

  \vspace{1ex}
  \begin{minipage}[t]{0.96\textwidth}
    \medskip
    \begin{tabular}{@{}ll@{}}
      \textit{Gemini-2.0-Flash Choice}:   & Model A is better \\
    \end{tabular}

    \medskip
    \textbf{Checklist Rationale}:
Model A is more relevant to the prompt than Model B.Model A directly discusses the cultural associations of frying food, such as home-cooked meals, social gatherings, and street food vendors. Model B also mentions Western and Asian cuisines, but also says it can evoke feelings of comfort or familiarity, which is not directly related to culture. Therefore, Model A is chosen as the better response.
  \end{minipage}
  \end{tcolorbox}
  \caption{Audio to Text - \pair ~- Relevance.}
  \label{Fig: a2t-case-p}
\end{figure*}

\begin{figure*}[htbp]
  \centering
  \begin{tcolorbox}[
    enhanced,
    attach boxed title to top center={yshift=-3mm,yshifttext=-1mm},
    boxrule=0.9pt,
    colback=gray!00,
    colframe=black!50,
    colbacktitle=Periwinkle,
    title=Case 12: Video to Text - \pair ~- Trustworthiness,
    boxed title style={size=small,colframe=Periwinkle},
    top=3mm,
    bottom=3mm,
    left=2mm,
    right=2mm,
    before upper={\parindent0em\linespread{0.85}\selectfont} 
  ]
  \vspace{-1ex}
  \begin{minipage}[t]{0.96\textwidth}
    \textbf{\textit{Checklists for \pair of Trustworthiness:}}
    \begin{itemize}[leftmargin=*,noitemsep,topsep=2pt]
      \item Does the response correctly identify the setting as a bathroom?
      \item Does the response avoid fabricating details or actions not present in the video?
      \item Does the response refrain from including information about the man's emotions or intentions, which are not explicitly shown?
      \item Does the response avoid ascribing a specific mood or tone to the video (e.g., 'tense', 'dramatic'), as these are interpretations, not objective facts?
    \end{itemize}
    
    \smallskip
    \textbf{\textit{Question}}: What does the video captures?
  \end{minipage}
  \vspace{0.5ex}

\vspace{1ex}
    \begin{minipage}[t]{0.98\textwidth}
    \centering
    \includegraphics[width=0.50\linewidth,trim=5 5 5 5,clip]{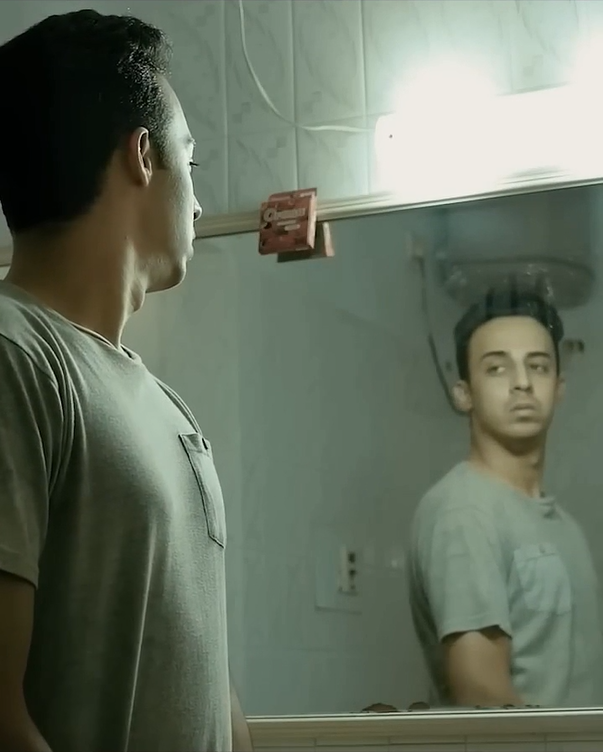}
    \captionof{subfigure}{Frame of Input Video}
  \end{minipage}

\begin{minipage}[t]{0.96\textwidth}
    \textbf{Model A answer}: The video captures a man in a bathroom setting. He is seen standing in front of a mirror, adjusting his shirt, and then looking at his reflection. The man appears to be checking his appearance, possibly preparing for an event or routine activity. The bathroom has tiled walls and a light fixture above the mirror. The overall atmosphere suggests a moment of self-reflection or preparation.\\
    \smallskip
    \textbf{Model B answer}: The video captures a man standing in front of a mirror in a bathroom. The man is wearing a short-sleeved gray shirt. He bends over and makes a spitting sound before looking at his reflection in the mirror. A white light is positioned above the mirror. To the right of the man's reflection, a hot water heater can be seen, as well as a rack holding toiletries. To the left of his reflection, a white towel appears to be draped over a hook. A red item is taped to the corner of the mirror.
\end{minipage}

  \vspace{1ex}
  \begin{minipage}[t]{0.96\textwidth}
    \medskip
    \begin{tabular}{@{}ll@{}}
      \textit{Gemini-2.0-Flash Choice}:   & Model A is better \\
    \end{tabular}

    \medskip
    \textbf{Checklist Rationale}:
Model A's response is more detailed and accurate than Model B's response. Model A correctly identifies the setting as a bathroom and describes the person's actions as \"possibly engaged in a routine activity like washing up or checking their appearance.\" This is a reasonable interpretation of the video, as the person is seen turning and moving around. Model B's response is less detailed and does not provide as much information about the person's actions. Additionally, Model B's response includes the phrase \"over a period of time,\" which is redundant, since all videos capture events over a period of time. Both models avoid fabricating details or actions not present in the video, as well as avoid including information about the man's emotions or intentions and avoid ascribing a specific mood or tone to the video. Thus, Model A's response is more detailed and accurate than Model B's response.
  \end{minipage}
  \end{tcolorbox}
  \caption{Video to Text - \pair ~- Trustworthiness.}
  \label{Fig: v2t-case-p}
\end{figure*}

\begin{figure*}[htbp]
  \centering
  \begin{tcolorbox}[
    enhanced,
    attach boxed title to top center={yshift=-3mm,yshifttext=-1mm},
    boxrule=0.9pt,
    colback=gray!00,
    colframe=black!50,
    colbacktitle=Periwinkle,
    title=Case 14: Audio Video to Text - \pair ~- Trustworthiness,
    boxed title style={size=small,colframe=Periwinkle},
    top=3mm,
    bottom=3mm,
    left=2mm,
    right=2mm,
    before upper={\parindent0em\linespread{0.85}\selectfont} 
  ]
  \vspace{-1ex}
  \begin{minipage}[t]{0.96\textwidth}
    \textbf{\textit{Checklists for \pair of Trustworthiness:}}
    \begin{itemize}[leftmargin=*,noitemsep,topsep=2pt]
      \item Does the response avoid making unsubstantiated claims about the emotional impact of the sound or visuals?
      \item If the response mentions any specific cultural or symbolic interpretations of bamboo or wind chimes, are these interpretations accurate and appropriately sourced?
      \item Is the information presented consistent with generally accepted knowledge about the natural world and human perception of sound and visuals?
      \item Does the response avoid any potentially harmful or misleading information about bamboo, wind chimes, or related topics?
    \end{itemize}
    
    \smallskip
    \textbf{\textit{Question}}: How does the sound enhance the visual experience of the plant, and what emotions might it evoke in the viewer?
  \end{minipage}
  \vspace{0.5ex}

\vspace{1ex}
  \begin{minipage}[t]{0.48\textwidth}
    \centering
    \includegraphics[width=0.80\linewidth,trim=5 5 5 5,clip]{Figures/Case/AudioIcon.png}
    \captionof{subfigure}{Input Audio}
  \end{minipage}
  \hfill
    \begin{minipage}[t]{0.48\textwidth}
    \centering
    \includegraphics[width=0.80\linewidth,trim=5 5 5 5,clip]{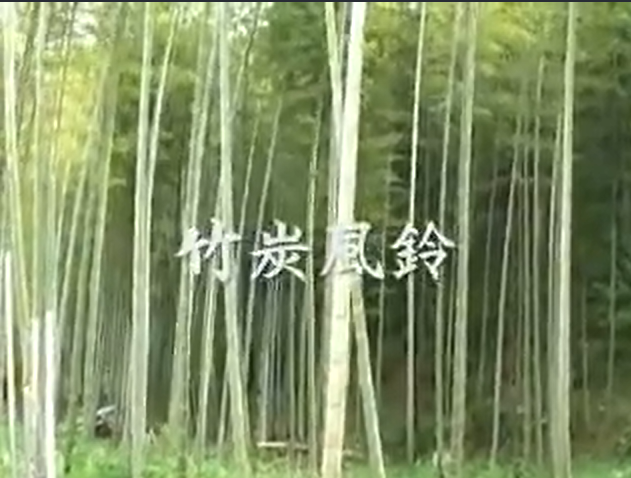}
    \captionof{subfigure}{Frame of Input Video}
  \end{minipage}

\begin{minipage}[t]{0.96\textwidth}
    \textbf{Model A answer}: The gentle clanging of the wind chimes enhances the visual experience of the bamboo forest by creating a sense of peace and tranquility. The sound complements the swaying of the bamboo stalks, further immersing the viewer in the natural environment. The sounds, combined with the visuals, could evoke various emotions depending on the viewer's state of mind, but likely ones include:  * **Calmness:** The soft, rhythmic chimes could induce relaxation and a sense of peace. * **Nostalgia:** The sounds of nature, particularly wind chimes, can trigger memories of past experiences in similar environments. * **Wonder:** The beauty of the bamboo forest, coupled with the delicate sounds, could inspire awe and appreciation for the natural world. * **Serenity:** The combination of sights and sounds could create a feeling of being disconnected from daily worries and stresses.  Essentially, the sound adds another layer of sensory input, making the experience more immersive and emotionally engaging.\\
    \smallskip
    \textbf{Model B answer}: The sound of wind chimes enhances the visual experience of the bamboo forest by adding an auditory element to the scene. The gentle clanging of the chimes in the breeze creates a soothing and peaceful atmosphere, complementing the serene visuals of the tall bamboo stalks swaying in the wind. The sound evokes feelings of tranquility and harmony with nature, allowing the viewer to fully immerse themselves in the beauty of the forest.
\end{minipage}

  \vspace{1ex}
  \begin{minipage}[t]{0.96\textwidth}
    \medskip
    \begin{tabular}{@{}ll@{}}
      \textit{Gemini-2.0-Flash Choice}:   & Model B is better \\
    \end{tabular}

    \medskip
    \textbf{Checklist Rationale}:
Model A's response is generally good, but it makes an unsubstantiated claim, saying that wind chimes are often associated with warm summer days and lazy afternoons spent outdoors, which can be subjective. Model B's response is accurate and avoids unsubstantiated claims. So I prefer Model B.
  \end{minipage}
  \end{tcolorbox}
  \caption{Audio Video to Text - \pair ~- Trustworthiness.}
  \label{Fig: av2t-case-p}
\end{figure*}

\begin{figure*}[htbp]
  \centering
  \begin{tcolorbox}[
    enhanced,
    attach boxed title to top center={yshift=-3mm,yshifttext=-1mm},
    boxrule=0.9pt,
    colback=gray!00,
    colframe=black!50,
    colbacktitle=Periwinkle,
    title=Case 16: Text to Video - \pair ~- Trustworthiness,
    boxed title style={size=small,colframe=Periwinkle},
    top=3mm,
    bottom=3mm,
    left=2mm,
    right=2mm,
    before upper={\parindent0em\linespread{0.85}\selectfont} 
  ]
  \vspace{-1ex}
  \begin{minipage}[t]{0.96\textwidth}
    \textbf{\textit{Checklists for \pair of Trustworthiness:}}
    \begin{itemize}[leftmargin=*,noitemsep,topsep=2pt]
      \item The video doesn't show any hazardous or misleading situations in its depiction of the kite and the snowboard.
      \item The video avoids presenting any factual inaccuracies related to the physics of kite flying or snowboarding, if applicable.
      \item The generated video does not contain any misleading information or elements that could be misconstrued by the viewer.
    \end{itemize}
    
    \smallskip
    \textbf{\textit{Question}}: Give me a video that illustrates the idea or scene described below. A vibrant snowboard, adorned with dynamic graphics and bold colors, is securely mounted atop a sleek, high-performance kite. The scene captures the front view, showcasing the snowboard's intricate design and the kite's aerodynamic structure. The kite's fabric, a striking blend of neon hues, billows gracefully against a backdrop of a clear, azure sky. The snowboard's bindings are prominently displayed, hinting at the thrilling adventure that awaits. The entire setup, bathed in the golden glow of the sun, exudes an aura of excitement and innovation, promising an exhilarating ride through the skies.
  \end{minipage}
  \vspace{0.5ex}

  \vspace{1ex}
  \begin{minipage}[t]{0.48\textwidth}
    \centering
    \includegraphics[width=0.95\linewidth,trim=5 5 5 5,clip]{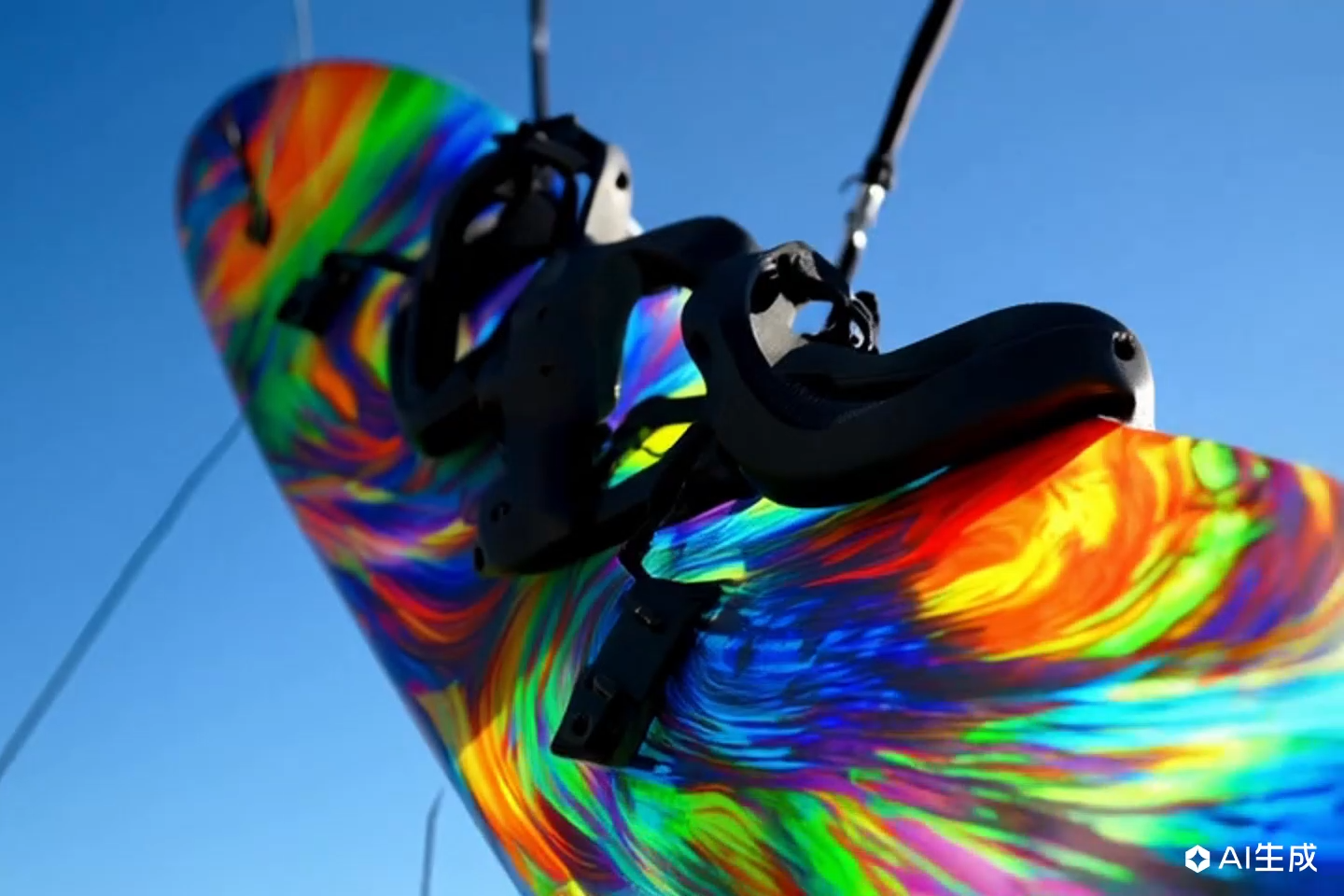}
    \captionof{subfigure}{Model A's answer}
  \end{minipage}
  \hfill
  \begin{minipage}[t]{0.48\textwidth}
    \centering
    \includegraphics[width=0.95\linewidth,trim=5 5 5 5,clip]{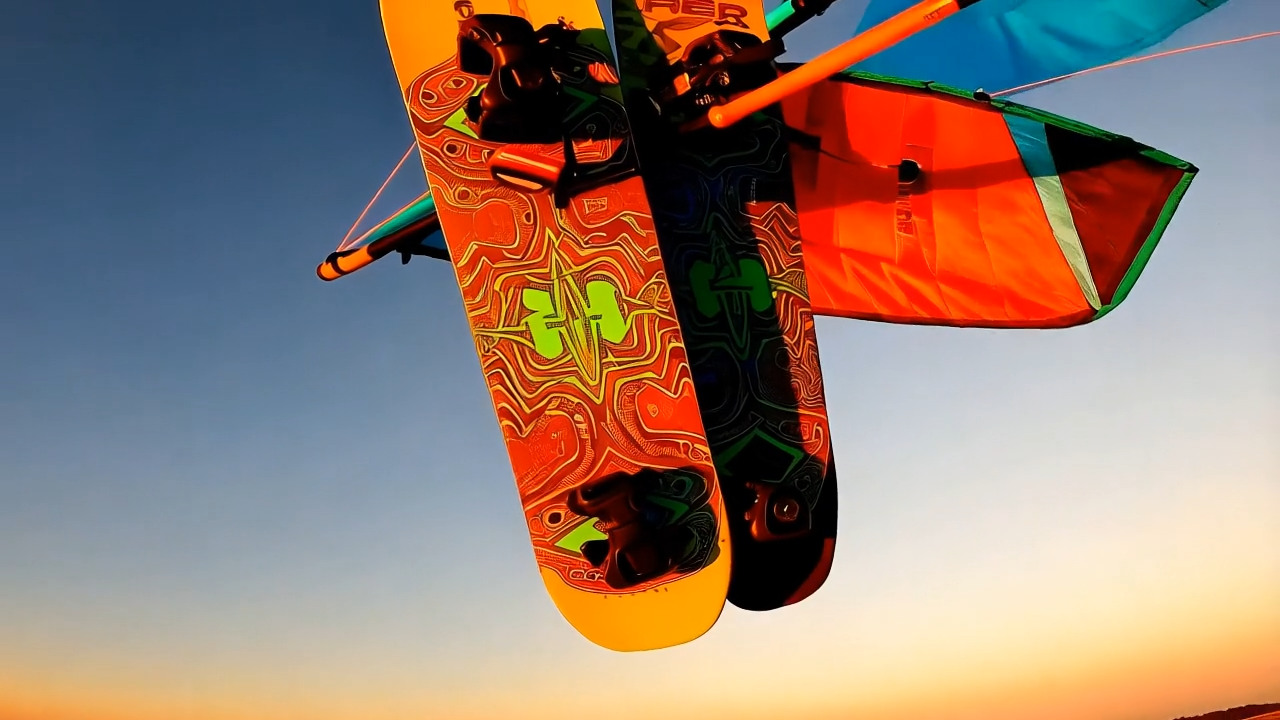}
    \captionof{subfigure}{Model B's answer}
  \end{minipage}

  \vspace{1ex}
  \begin{minipage}[t]{0.96\textwidth}
    \medskip
    \begin{tabular}{@{}ll@{}}
      \textit{Gemini-1.5-Pro Choice}:   & Model A is better. \\
    \end{tabular}

    \medskip
    \textbf{Checklist Rationale}:
Model A's video adheres more closely to the prompt's description of a snowboard attached to a kite. Although the kite's design is simple, it still fulfills the request. Model B's video appears to show two snowboards attached to what seems to be a collapsed kite, deviating significantly from the prompt. Hence, Model A is chosen as its visual representation aligns better with the given instructions.
  \end{minipage}
  \end{tcolorbox}
  \caption{Text to Video - \pair ~- Trustworthiness.}
  \label{Fig: t2v-case-p}
\end{figure*}

\begin{figure*}[htbp]
  \centering
  \begin{tcolorbox}[
    enhanced,
    attach boxed title to top center={yshift=-3mm,yshifttext=-1mm},
    boxrule=0.9pt,
    colback=gray!00,
    colframe=black!50,
    colbacktitle=Periwinkle,
    title=Case 20: Image Edit - \pair ~- Clarity,
    boxed title style={size=small,colframe=Periwinkle},
    top=3mm,
    bottom=3mm,
    left=2mm,
    right=2mm,
    before upper={\parindent0em\linespread{0.85}\selectfont} 
  ]
  \vspace{-1ex}
  \begin{minipage}[t]{0.96\textwidth}
    \textbf{\textit{Checklists for \pair of Clarity:}}
    \begin{itemize}[leftmargin=*,noitemsep,topsep=2pt]
      \item Verify that the transformation includes recognizable jungle features such as trees, plants, wildlife, and natural textures.
      \item Ensure the overall composition is coherent and elements are not visually clashing or ambiguous.
      \item Ensure that the transition from urban to jungle is smooth and does not leave any unfinished or partial elements.
    \end{itemize}
    
    \smallskip
    \textbf{\textit{Question}}: Transform the scenery in this photo to evoke a jungle vista
  \end{minipage}
  \vspace{0.5ex}

\vspace{1ex}
    \begin{minipage}[t]{0.30\textwidth}
    \centering
    \includegraphics[width=0.98\linewidth,trim=5 5 5 5,clip]{Figures/Case/i2i_0.jpg}
    \captionof{subfigure}{Input Image}
  \end{minipage}
\hfill
\vspace{1ex}
  \begin{minipage}[t]{0.30\textwidth}
    \centering
    \includegraphics[width=0.98\linewidth,trim=5 5 5 5,clip]{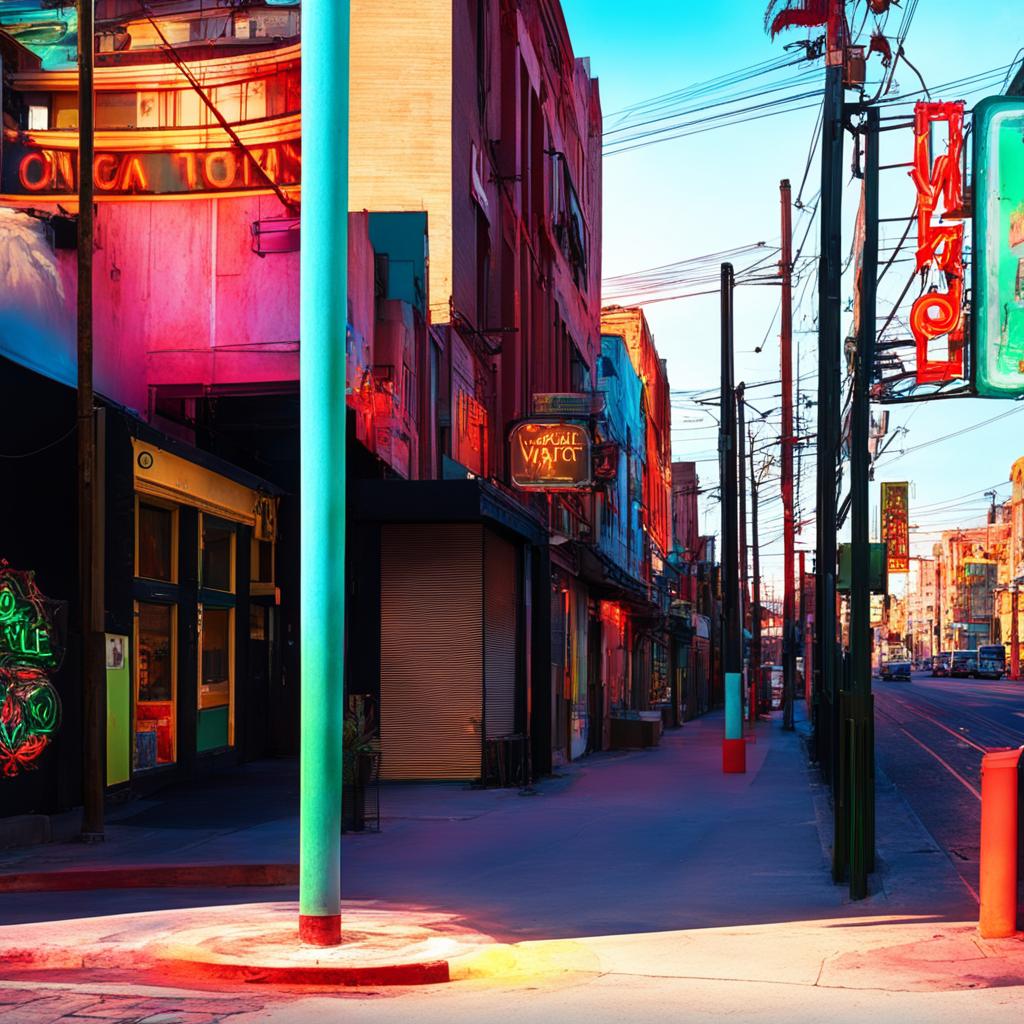}
    \captionof{subfigure}{Model A's answer}
  \end{minipage}
\hfill
  \begin{minipage}[t]{0.30\textwidth}
    \centering
    \includegraphics[width=0.98\linewidth,trim=5 5 5 5,clip]{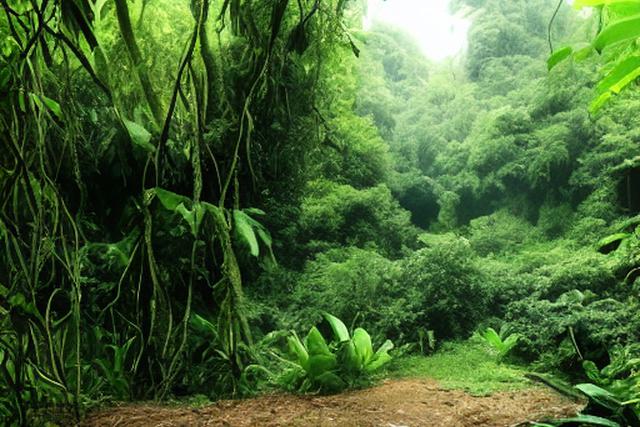}
    \captionof{subfigure}{Model B's answer}
  \end{minipage}

  \begin{minipage}[t]{0.98\textwidth}
    \medskip
    \begin{tabular}{@{}ll@{}}
      \textit{Gemini-2.0-Flash Choice}:   & Tie\\
    \end{tabular}

    \medskip
    \textbf{Checklist Rationale}:]
Both models failed to perform the given task. They basically returned the original image without any transformation. Since they perform equally poorly, I rate them as equally good.

  \end{minipage}
  \end{tcolorbox}
  \caption{Image Edit - \pair ~- Clarity.}
  \label{Fig: i2i-case-p}
\end{figure*}

\begin{figure*}[htbp]
  \centering
  \begin{tcolorbox}[
    enhanced,
    attach boxed title to top center={yshift=-3mm,yshifttext=-1mm},
    boxrule=0.9pt,
    colback=gray!00,
    colframe=black!50,
    colbacktitle=Periwinkle,
    title=Case 21: Image Edit - \score ~- Clarity,
    boxed title style={size=small,colframe=Periwinkle},
    top=3mm,
    bottom=3mm,
    left=2mm,
    right=2mm,
    before upper={\parindent0em\linespread{0.85}\selectfont} 
  ]
  \vspace{-1ex}
  \begin{minipage}[t]{0.96\textwidth}
    \textbf{\textit{Checklists for \pair of Clarity:}}
    \begin{itemize}[leftmargin=*,noitemsep,topsep=2pt]
      \item Verify that the transformation includes recognizable jungle features such as trees, plants, wildlife, and natural textures.
      \item Ensure the overall composition is coherent and elements are not visually clashing or ambiguous.
      \item Ensure that the transition from urban to jungle is smooth and does not leave any unfinished or partial elements.
    \end{itemize}
    
    \smallskip
    \textbf{\textit{Question}}: Transform the scenery in this photo to evoke a jungle vista
  \end{minipage}
  \vspace{0.5ex}

\vspace{1ex}
    \begin{minipage}[t]{0.45\textwidth}
    \centering
    \includegraphics[width=0.60\linewidth,trim=5 5 5 5,clip]{Figures/Case/i2i_0.jpg}
    \captionof{subfigure}{Input Image}
  \end{minipage}
\vspace{1ex}
\hfill
  \begin{minipage}[t]{0.45\textwidth}
    \centering
    \includegraphics[width=0.60\linewidth,trim=5 5 5 5,clip]{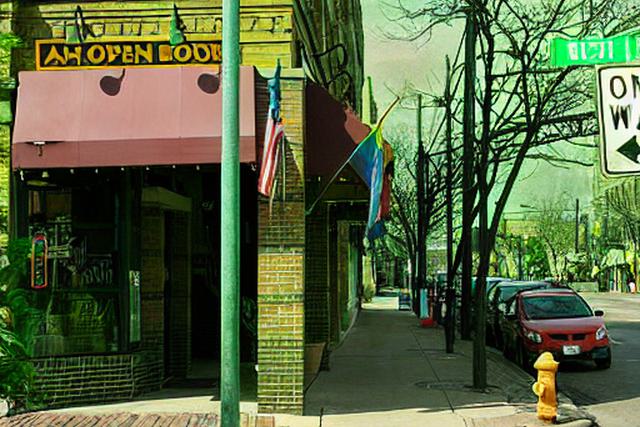}
    \captionof{subfigure}{Model's answer}
  \end{minipage}

  \vspace{1ex}
  \begin{minipage}[t]{0.98\textwidth}
    \medskip
    \begin{tabular}{@{}ll@{}}
      \textit{Gemini-2.0-Flash Score}:   & 2 \\
    \end{tabular}

    \medskip
    \textbf{Checklist Rationale}: 
Checklist Item 1: The transformation primarily uses color manipulation to evoke a jungle feel. The green tint applied to the image attempts to simulate foliage, but lacks distinct jungle features like specific plants, wildlife, or natural textures. This makes the jungle theme ambiguous and relies heavily on viewer interpretation. Checklist Item 2: The overall composition remains largely the same as the original image. The green tint doesn't clash with existing elements, but it doesn't integrate seamlessly either, creating a somewhat artificial look. Checklist Item 3: The transition is abrupt and relies solely on the color change. There's no gradual integration of jungle elements or modification of existing urban features to fit the theme. The urban elements remain clearly visible, conflicting with the intended jungle vista.
  \end{minipage}
  \end{tcolorbox}
  \caption{Image Edit - \score ~- Clarity.}
  \label{Fig: i2i-case-s}
\end{figure*}

\begin{figure*}[htbp]
  \centering
  \begin{tcolorbox}[
    enhanced,
    attach boxed title to top center={yshift=-3mm,yshifttext=-1mm},
    boxrule=0.9pt,
    colback=gray!00,
    colframe=black!50,
    colbacktitle=Periwinkle,
    title=Case 22: Image to Audio - \pair ~- Creativity,
    boxed title style={size=small,colframe=Periwinkle},
    top=3mm,
    bottom=3mm,
    left=2mm,
    right=2mm,
    before upper={\parindent0em\linespread{0.85}\selectfont} 
  ]
  \vspace{-1ex}
  \begin{minipage}[t]{0.96\textwidth}
    \textbf{\textit{Checklists for \pair of Creativity:}}
    \begin{itemize}[leftmargin=*,noitemsep,topsep=2pt]
      \item Is the atmosphere of a racetrack captured, possibly including crowd noises, tire squealing, or pit stop sounds?
      \item Does the audio include subtle, creative elements like the sounds of wind or mechanical adjustments?
    \end{itemize}
    
    \smallskip
    \textbf{\textit{Question}}: Create audio that reflects the image.
  \end{minipage}
  \vspace{0.5ex}

\vspace{1ex}
    \begin{minipage}[t]{0.30\textwidth}
    \centering
    \includegraphics[width=0.98\linewidth,trim=5 5 5 5,clip]{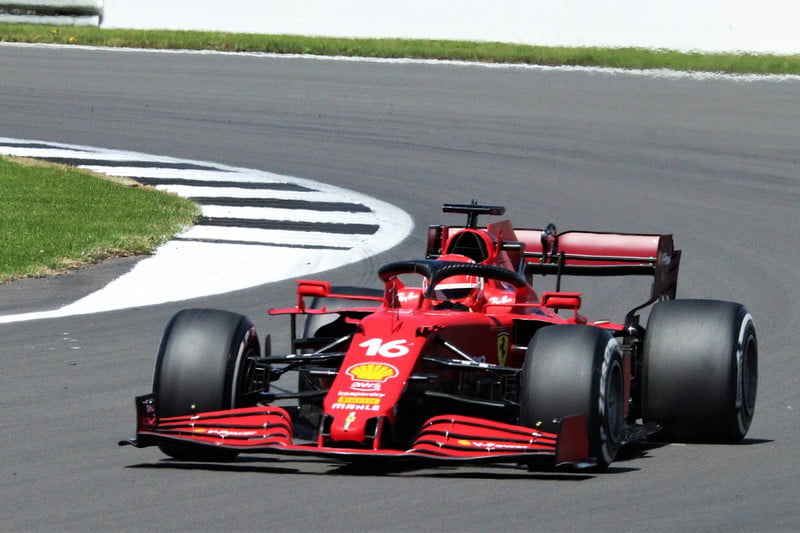}
    \captionof{subfigure}{Input Image}
  \end{minipage}
\hfill
\vspace{1ex}
  \begin{minipage}[t]{0.30\textwidth}
    \centering
    \includegraphics[width=0.98\linewidth,trim=5 5 5 5,clip]{Figures/Case/AudioIcon.png}
    \captionof{subfigure}{Model A's answer}
  \end{minipage}
  \hfill
  \begin{minipage}[t]{0.30\textwidth}
    \centering
    \includegraphics[width=0.98\linewidth,trim=5 5 5 5,clip]{Figures/Case/AudioIcon.png}
    \captionof{subfigure}{Model B's answer}
  \end{minipage}

  \vspace{1ex}
  \begin{minipage}[t]{0.98\textwidth}
    \medskip
    \begin{tabular}{@{}ll@{}}
      \textit{Gemini-2.0-Flash Choice}:   & Model B is better \\
    \end{tabular}

    \medskip
    \textbf{Checklist Rationale}:]
Response A's audio mainly focuses on the engine sounds of the race car and does not include the atmosphere of the racetrack, such as crowd noises or tire squealing. Response B's audio does a better job of capturing the race ambiance. It includes the sounds of the race car, the roar of the crowd, and the announcer's voice. It has more creative elements like the sounds of wind or mechanical adjustments. The other audio gives a more convincing impression of a Formula B race. Therefore, the audio from response B better reflects the image.

  \end{minipage}
  \end{tcolorbox}
  \caption{Image to Audio - \pair ~- Creativity.}
  \label{Fig: i2a-case-p}
\end{figure*}

\begin{figure*}[htbp]
  \centering
  \begin{tcolorbox}[
    enhanced,
    attach boxed title to top center={yshift=-3mm,yshifttext=-1mm},
    boxrule=0.9pt,
    colback=gray!00,
    colframe=black!50,
    colbacktitle=Periwinkle,
    title=Case 24: Image to Video - \pair ~- Completeness,
    boxed title style={size=small,colframe=Periwinkle},
    top=3mm,
    bottom=3mm,
    left=2mm,
    right=2mm,
    before upper={\parindent0em\linespread{0.85}\selectfont} 
  ]
  \vspace{-1ex}
  \begin{minipage}[t]{0.96\textwidth}
    \textbf{\textit{Checklists for \pair of Completeness:}}
    \begin{itemize}[leftmargin=*,noitemsep,topsep=2pt]
      \item The mountain foot is recognizable and serves as a background element.
      \item Reflections of environmental elements such as trees or grass around the plank road are visible.
      \item The portrayal of the girl's movement is smooth and coherent.
      \item Detail in landscapes, such as greenery and distant mountains, is preserved.
    \end{itemize}
    
    \smallskip
    \textbf{\textit{Question}}: A girl is walking on the plank road at the foot of the mountain.
  \end{minipage}
  \vspace{0.5ex}

\vspace{1ex}
    \begin{minipage}[t]{0.30\textwidth}
    \centering
    \includegraphics[width=0.98\linewidth,trim=5 5 5 5,clip]{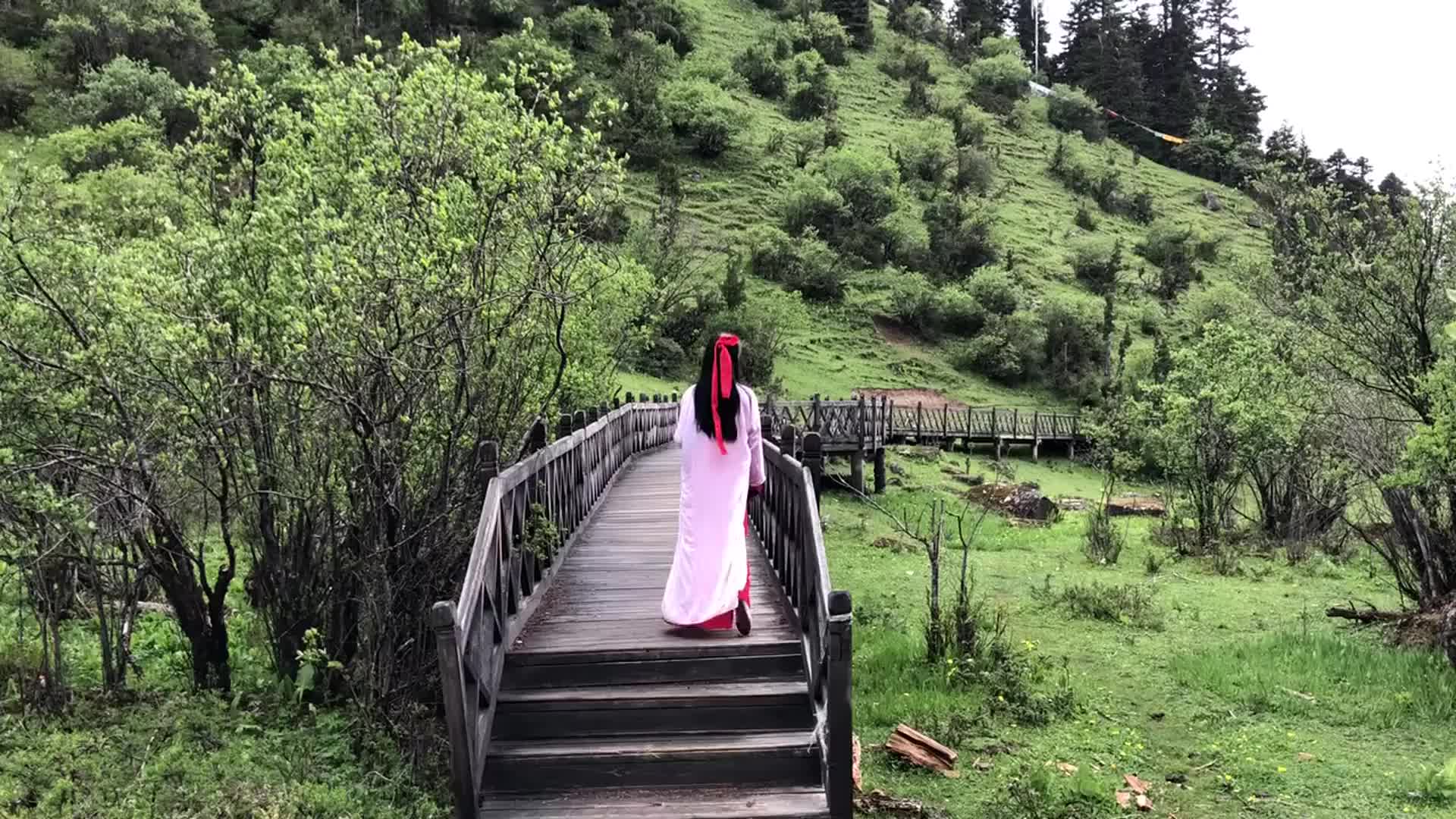}
    \captionof{subfigure}{Input Image}
  \end{minipage}
\hfill
\vspace{1ex}
  \begin{minipage}[t]{0.30\textwidth}
    \centering
    \includegraphics[width=0.98\linewidth,trim=5 5 5 5,clip]{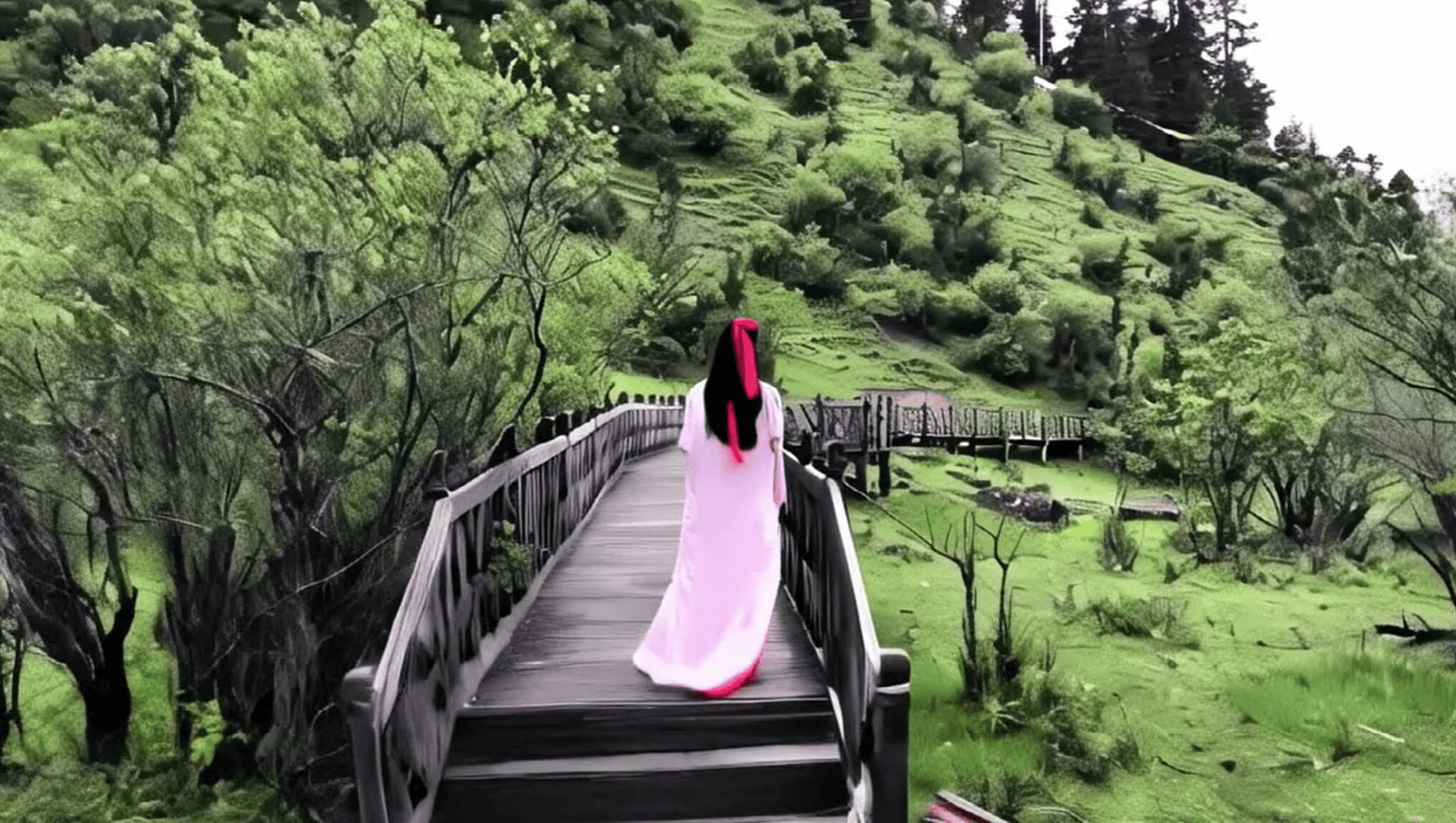}
    \captionof{subfigure}{Model A's answer}
  \end{minipage}
\hfill
  \begin{minipage}[t]{0.30\textwidth}
    \centering
    \includegraphics[width=0.98\linewidth,trim=5 5 5 5,clip]{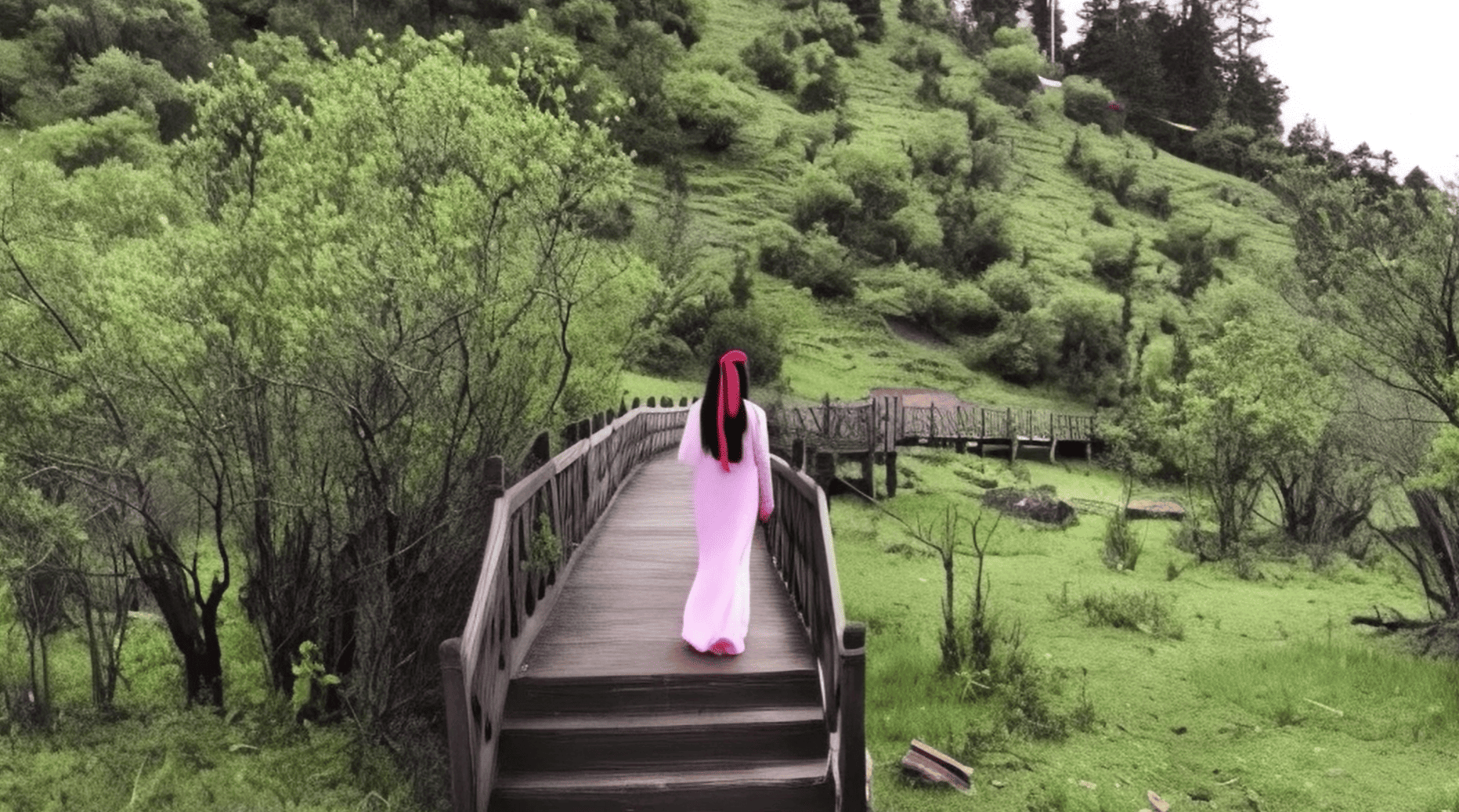}
    \captionof{subfigure}{Model B's answer}
  \end{minipage}

  \begin{minipage}[t]{0.98\textwidth}
    \medskip
    \begin{tabular}{@{}ll@{}}
      \textit{Gemini-2.0-Flash Choice}:   & Model B is better.\\
    \end{tabular}

    \medskip
    \textbf{Checklist Rationale}:]
Both models generate a short video about a girl walking on the plank road at the foot of the mountain. Overall, model B is a little better because the movement of the clothes is more natural, and the animation is smoother. Model A's dress moves too much, like it is waving in the wind, while there is no wind. Both videos are short, so it is hard to evaluate the smooth movement. The plank road and mountain foot are recognizable in both videos, and the details in landscape and reflections are preserved as well.

  \end{minipage}
  \end{tcolorbox}
  \caption{Image to Video - \pair ~- Completeness.}
  \label{Fig: i2v-case-p}
\end{figure*}

\begin{figure*}[htbp]
  \centering
  \begin{tcolorbox}[
    enhanced,
    attach boxed title to top center={yshift=-3mm,yshifttext=-1mm},
    boxrule=0.9pt,
    colback=gray!00,
    colframe=black!50,
    colbacktitle=Periwinkle,
    title=Case 25: Image to Video - \score ~- Completeness,
    boxed title style={size=small,colframe=Periwinkle},
    top=3mm,
    bottom=3mm,
    left=2mm,
    right=2mm,
    before upper={\parindent0em\linespread{0.85}\selectfont} 
  ]
  \vspace{-1ex}
  \begin{minipage}[t]{0.96\textwidth}
    \textbf{\textit{Checklists for \pair of Completeness:}}
    \begin{itemize}[leftmargin=*,noitemsep,topsep=2pt]
      \item Is the atmosphere of a racetrack captured, possibly including crowd noises, tire squealing, or pit stop sounds?
      \item Does the audio include subtle, creative elements like the sounds of wind or mechanical adjustments?
    \end{itemize}
    
    \smallskip
    \textbf{\textit{Question}}: Create audio that reflects the image.
  \end{minipage}
  \vspace{0.5ex}

\vspace{1ex}
    \begin{minipage}[t]{0.45\textwidth}
    \centering
    \includegraphics[width=0.60\linewidth,trim=5 5 5 5,clip]{Figures/Case/i2v_0.jpg}
    \captionof{subfigure}{Input Image}
  \end{minipage}
\vspace{1ex}
\hfill
  \begin{minipage}[t]{0.45\textwidth}
    \centering
    \includegraphics[width=0.60\linewidth,trim=5 5 5 5,clip]{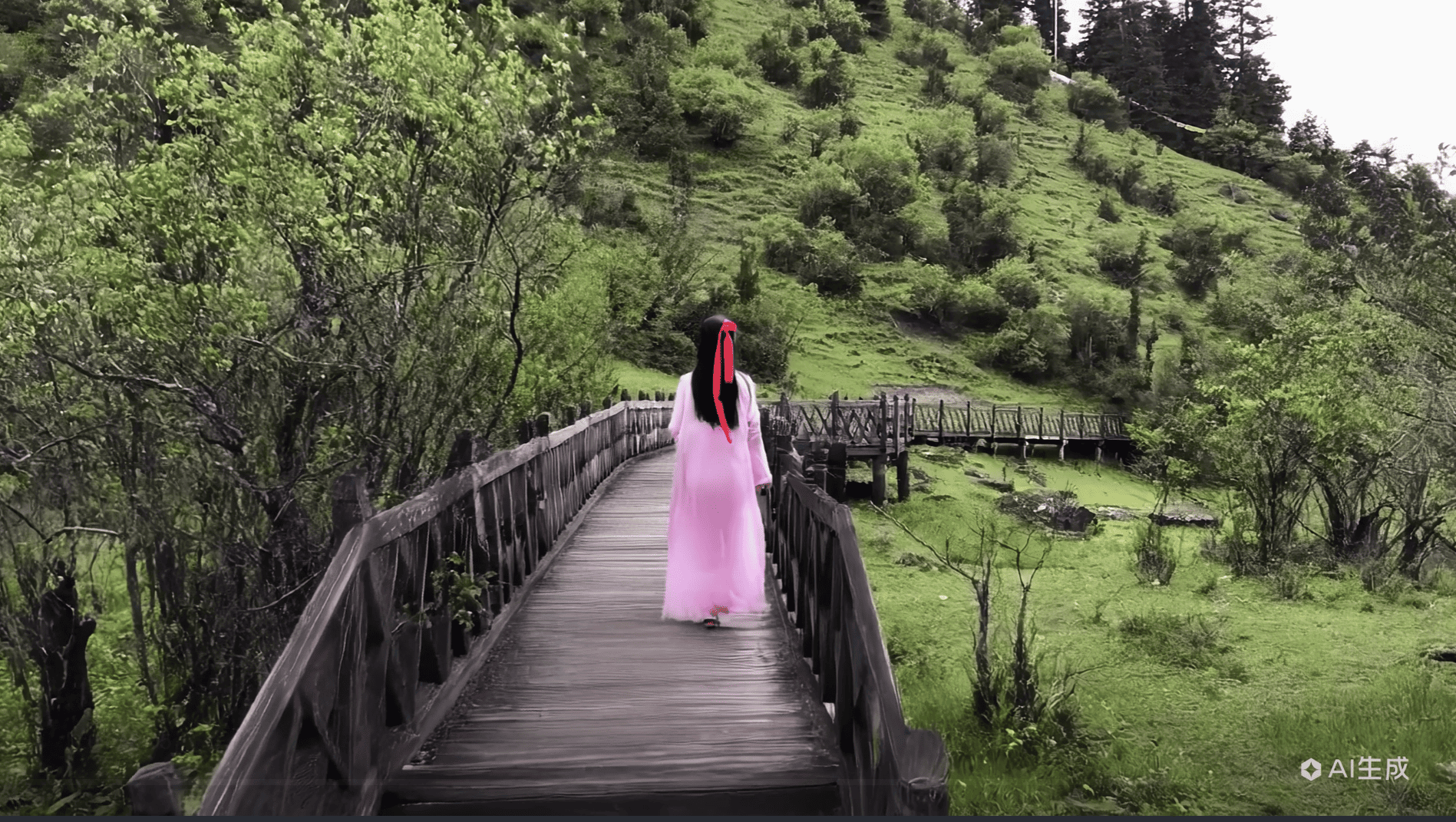}
    \captionof{subfigure}{Model's answer}
  \end{minipage}

  \vspace{1ex}
  \begin{minipage}[t]{0.98\textwidth}
    \medskip
    \begin{tabular}{@{}ll@{}}
      \textit{Gemini-2.0-Flash Score}:   & 1 \\
    \end{tabular}

    \medskip
    \textbf{Checklist Rationale}: 
The generated video is a static image, it does not show any movement at all. This makes it severely incomplete as it fails to meet all checklist items. Checklist item 1: Recognizable mountain foot as background (Pass). The mountain is clearly visible in the background. Checklist item 2: Reflections of environmental elements (Fail). No reflections are visible as the video is static. Checklist item 3: Smooth and coherent girl movement (Fail). The girl does not move in the static video. Checklist item 4: Detail in landscape preserved (Pass). The details of the landscape are preserved, but no additional details are added.
  \end{minipage}
  \end{tcolorbox}
  \caption{Image to Video - \score ~- Completeness.}
  \label{Fig: i2v-case-s}
\end{figure*}

\begin{figure*}[htbp]
  \centering
  \begin{tcolorbox}[
    enhanced,
    attach boxed title to top center={yshift=-3mm,yshifttext=-1mm},
    boxrule=0.9pt,
    colback=gray!00,
    colframe=black!50,
    colbacktitle=Periwinkle,
    title=Case 28: Audio to Video - \pair ~- Trustworthiness,
    boxed title style={size=small,colframe=Periwinkle},
    top=3mm,
    bottom=3mm,
    left=2mm,
    right=2mm,
    before upper={\parindent0em\linespread{0.85}\selectfont} 
  ]
  \vspace{-1ex}
  \begin{minipage}[t]{0.96\textwidth}
    \textbf{\textit{Checklists for \pair of Trustworthiness:}}
    \begin{itemize}[leftmargin=*,noitemsep,topsep=2pt]
      \item Does the video accurately depict a sharp, distinct 'pop' sound?
      \item Does the video refrain from including any additional sounds or visuals that are not present in the audio, unless they enhance the depiction of the described sounds?
      \item If the video depicts a gunshot, does it do so responsibly and avoid any harmful or misleading depictions of violence or unsafe behavior?
    \end{itemize}
    
    \smallskip
    \textbf{\textit{Question}}: Generate a video based on the given sound.
  \end{minipage}
  \vspace{0.5ex}

\vspace{1ex}
    \begin{minipage}[t]{0.30\textwidth}
    \centering
    \includegraphics[width=0.98\linewidth,trim=5 5 5 5,clip]{Figures/Case/AudioIcon.png}
    \captionof{subfigure}{Input Audio}
  \end{minipage}
\hfill
\vspace{1ex}
  \begin{minipage}[t]{0.30\textwidth}
    \centering
    \includegraphics[width=0.98\linewidth,trim=5 5 5 5,clip]{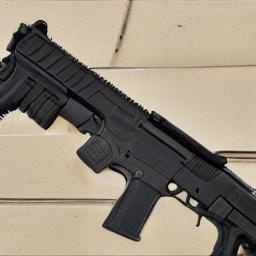}
    \captionof{subfigure}{Model A's answer}
  \end{minipage}
\hfill
  \begin{minipage}[t]{0.30\textwidth}
    \centering
    \includegraphics[width=0.98\linewidth,trim=5 5 5 5,clip]{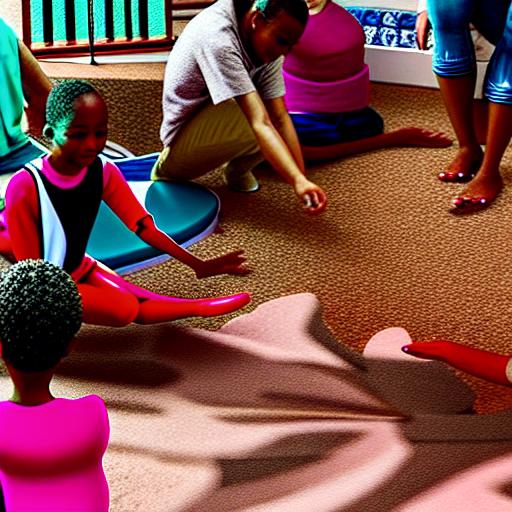}
    \captionof{subfigure}{Model B's answer}
  \end{minipage}

  \begin{minipage}[t]{0.98\textwidth}
    \medskip
    \begin{tabular}{@{}ll@{}}
      \textit{Gemini-2.0-Flash Choice}:   & Model B is better.\\
    \end{tabular}

    \medskip
    \textbf{Checklist Rationale}:]
Response A: The video includes guns and shooting which is inappropriate given the audio input. Response B: The video shows people laying out a blanket or something similar which can be inferred from the popping sound of something unfolding quickly. This is a more trustworthy response because the first one is harmful and inappropriate.

  \end{minipage}
  \end{tcolorbox}
  \caption{Audio to Video - \pair ~- Trustworthiness.}
  \label{Fig: a2v-case-p}
\end{figure*}

\begin{figure*}[htbp]
  \centering
  \begin{tcolorbox}[
    enhanced,
    attach boxed title to top center={yshift=-3mm,yshifttext=-1mm},
    boxrule=0.9pt,
    colback=gray!00,
    colframe=black!50,
    colbacktitle=Periwinkle,
    title=Case 29: Audio to Video - \score ~- Trustworthiness,
    boxed title style={size=small,colframe=Periwinkle},
    top=3mm,
    bottom=3mm,
    left=2mm,
    right=2mm,
    before upper={\parindent0em\linespread{0.85}\selectfont} 
  ]
  \vspace{-1ex}
  \begin{minipage}[t]{0.96\textwidth}
    \textbf{\textit{Checklists for \pair of Trustworthiness:}}
    \begin{itemize}[leftmargin=*,noitemsep,topsep=2pt]
      \item Does the video accurately depict a sharp, distinct 'pop' sound?
      \item Does the video refrain from including any additional sounds or visuals that are not present in the audio, unless they enhance the depiction of the described sounds?
      \item If the video depicts a gunshot, does it do so responsibly and avoid any harmful or misleading depictions of violence or unsafe behavior?
    \end{itemize}
    
    \smallskip
    \textbf{\textit{Question}}: Generate a video based on the given sound.
  \end{minipage}
  \vspace{0.5ex}

\vspace{1ex}
    \begin{minipage}[t]{0.45\textwidth}
    \centering
    \includegraphics[width=0.60\linewidth,trim=5 5 5 5,clip]{Figures/Case/AudioIcon.png}
    \captionof{subfigure}{Input Audio}
  \end{minipage}
\vspace{1ex}
\hfill
  \begin{minipage}[t]{0.45\textwidth}
    \centering
    \includegraphics[width=0.60\linewidth,trim=5 5 5 5,clip]{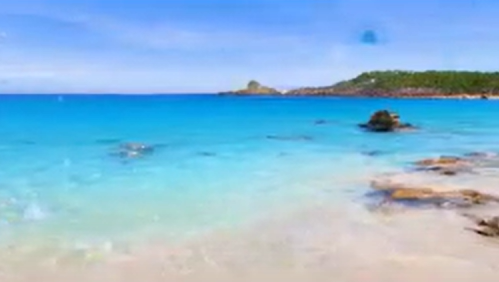}
    \captionof{subfigure}{Model's answer}
  \end{minipage}

  \vspace{1ex}
  \begin{minipage}[t]{0.98\textwidth}
    \medskip
    \begin{tabular}{@{}ll@{}}
      \textit{Gemini-2.0-Flash Score}:   & 1 \\
    \end{tabular}

    \medskip
    \textbf{Checklist Rationale}: 
The generated video does not align with the provided audio. The audio consists of what appears to be a distinct "pop" sound, while the video displays a tranquil beach scene. This disconnect between the audio and visual makes the response untrustworthy as it fails to accurately represent the sound. Checklist Evaluation: 1. Does the video accurately depict a sharp, distinct 'pop' sound? **No**. The video shows a calm beach, not visually representative of a 'pop' sound. 2. Does the video refrain from including any additional sounds or visuals that are not present in the audio, unless they enhance the depiction of the described sounds? **Yes**. The video doesn't add sounds, but the visuals don't relate to the audio. 3. If the video depicts a gunshot, does it do so responsibly and avoid any harmful or misleading depictions of violence or unsafe behavior? **N/A**. The audio doesn't seem to be a gunshot, and the video doesn't depict any violence.
  \end{minipage}
  \end{tcolorbox}
  \caption{Audio to Video - \score ~-Trustworthiness.}
  \label{Fig: a2v-case-s}
\end{figure*}

\begin{figure*}[htbp]
  \centering
  \begin{tcolorbox}[
    enhanced,
    attach boxed title to top center={yshift=-3mm,yshifttext=-1mm},
    boxrule=0.9pt,
    colback=gray!00,
    colframe=black!50,
    colbacktitle=Periwinkle,
    title=Case 30: Video to Audio - \pair ~- Creativity,
    boxed title style={size=small,colframe=Periwinkle},
    top=3mm,
    bottom=3mm,
    left=2mm,
    right=2mm,
    before upper={\parindent0em\linespread{0.85}\selectfont} 
  ]
  \vspace{-1ex}
  \begin{minipage}[t]{0.96\textwidth}
    \textbf{\textit{Checklists for \pair of Creativity:}}
    \begin{itemize}[leftmargin=*,noitemsep,topsep=2pt]
      \item Does the audio enhance the video by adding another layer of emotion except dog's whimpering like human's 
      \item Is the audio well-integrated with the visuals, creating a cohesive and immersive experience?
    \end{itemize}
    
    \smallskip
    \textbf{\textit{Question}}: Generate corresponding audio based on the video's visuals.
  \end{minipage}
  \vspace{0.5ex}

\vspace{1ex}
    \begin{minipage}[t]{0.30\textwidth}
    \centering
    \includegraphics[width=0.98\linewidth,trim=5 5 5 5,clip]{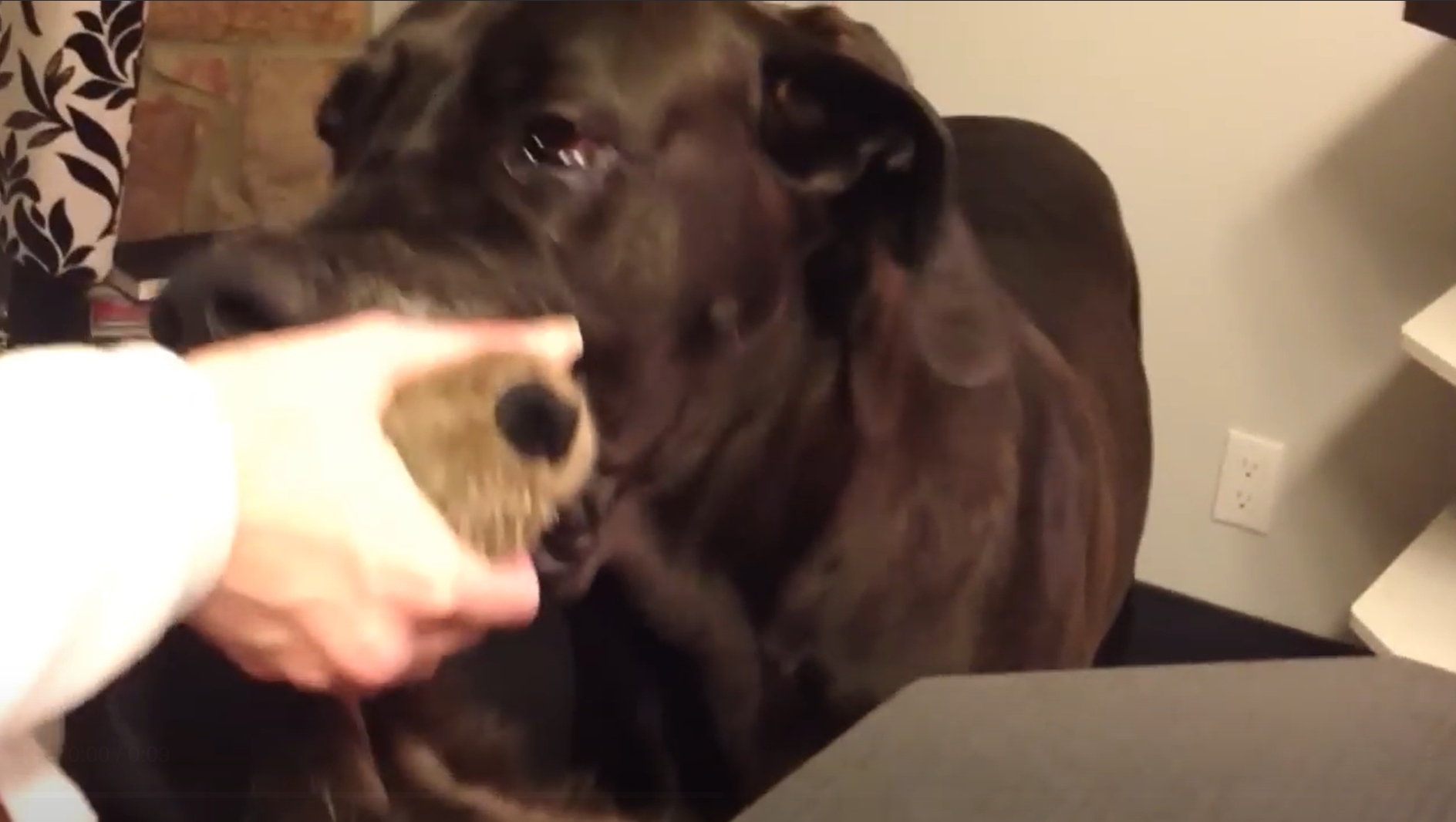}
    \captionof{subfigure}{Input Video}
  \end{minipage}
\hfill
\vspace{1ex}
  \begin{minipage}[t]{0.30\textwidth}
    \centering
    \includegraphics[width=0.98\linewidth,trim=5 5 5 5,clip]{Figures/Case/AudioIcon.png}
    \captionof{subfigure}{Model A's answer}
  \end{minipage}
\hfill
  \begin{minipage}[t]{0.30\textwidth}
    \centering
    \includegraphics[width=0.98\linewidth,trim=5 5 5 5,clip]{Figures/Case/AudioIcon.png}
    \captionof{subfigure}{Model B's answer}
  \end{minipage}

  \begin{minipage}[t]{0.98\textwidth}
    \medskip
    \begin{tabular}{@{}ll@{}}
      \textit{Gemini-2.0-Flash Choice}:   & Model B is better.\\
    \end{tabular}

    \medskip
    \textbf{Checklist Rationale}:]
Model A: The audio mainly focuses on the dog's whimpering and doesn't add much emotional depth beyond that. While it syncs decently with the dog's actions, it could do more to create a truly immersive experience. Model B: This model successfully blends the dog's whimpers with baby-like sounds and cries, which adds a poignant layer of emotion. The integration of these sounds is well-executed, enhancing the sense of the dog wanting the toy and creating a more immersive experience. The audio quality is also slightly better, with less background noise. Overall, this model's creativity elevates the video's emotional impact, making it a more engaging and satisfying experience. Thus, Model B surpasses Model A in terms of creativity by effectively integrating additional emotional layers to create a more immersive experience.

  \end{minipage}
  \end{tcolorbox}
  \caption{Video to Audio - \pair ~- Creativity.}
  \label{Fig: v2a-case-p}
\end{figure*}

\begin{figure*}[htbp]
  \centering
  \begin{tcolorbox}[
    enhanced,
    attach boxed title to top center={yshift=-3mm,yshifttext=-1mm},
    boxrule=0.9pt,
    colback=gray!00,
    colframe=black!50,
    colbacktitle=Periwinkle,
    title=Case 32: Video Edit - \pair ~- Clarity,
    boxed title style={size=small,colframe=Periwinkle},
    top=3mm,
    bottom=3mm,
    left=2mm,
    right=2mm,
    before upper={\parindent0em\linespread{0.85}\selectfont} 
  ]
  \vspace{-1ex}
  \begin{minipage}[t]{0.96\textwidth}
    \textbf{\textit{Checklists for \pair of Clarity:}}
    \begin{itemize}[leftmargin=*,noitemsep,topsep=2pt]
      \item Is the aurora in the edited video predominantly red and yellow?
      \item Is the transition between colors in the aurora smooth and natural-looking?
      \item Is there any flickering or distortion that makes the aurora difficult to perceive clearly?
    \end{itemize}
    
    \smallskip
    \textbf{\textit{Question}}: Modify the video to show red and yellow aurora paints the night sky over mountain silhouettes. 
  \end{minipage}
  \vspace{0.5ex}

\vspace{1ex}
    \begin{minipage}[t]{0.30\textwidth}
    \centering
    \includegraphics[width=0.98\linewidth,trim=5 5 5 5,clip]{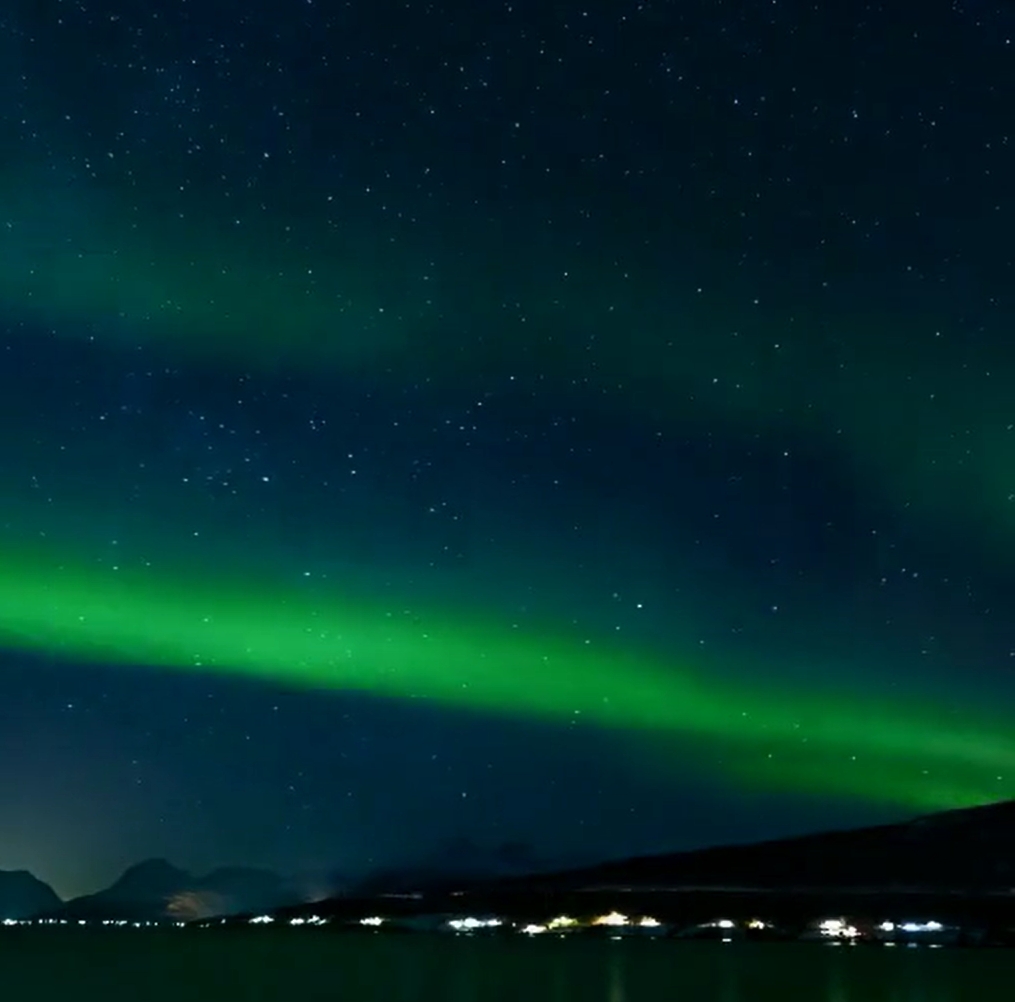}
    \captionof{subfigure}{Input Video}
  \end{minipage}
\hfill
\vspace{1ex}
  \begin{minipage}[t]{0.30\textwidth}
    \centering
    \includegraphics[width=0.98\linewidth,trim=5 5 5 5,clip]{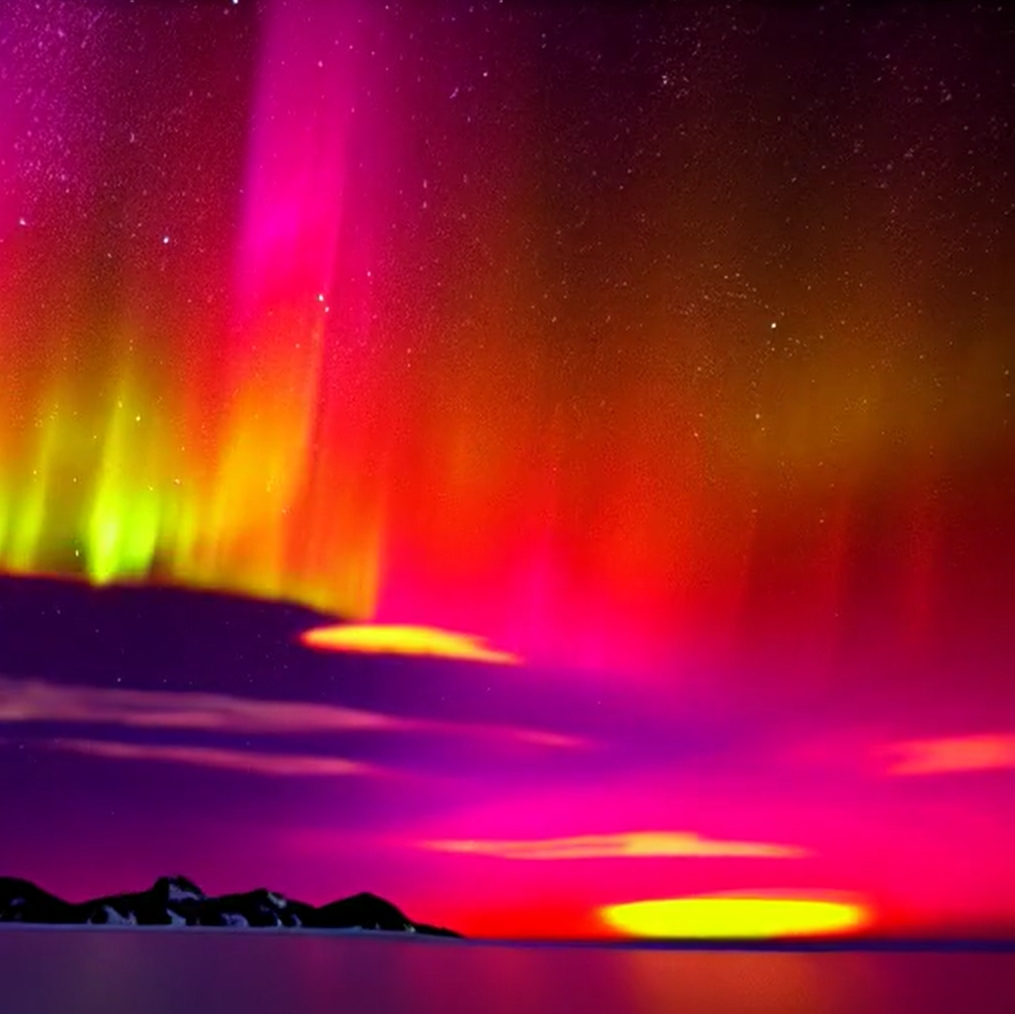}
    \captionof{subfigure}{Model A's answer}
  \end{minipage}
\hfill
  \begin{minipage}[t]{0.30\textwidth}
    \centering
    \includegraphics[width=0.98\linewidth,trim=5 5 5 5,clip]{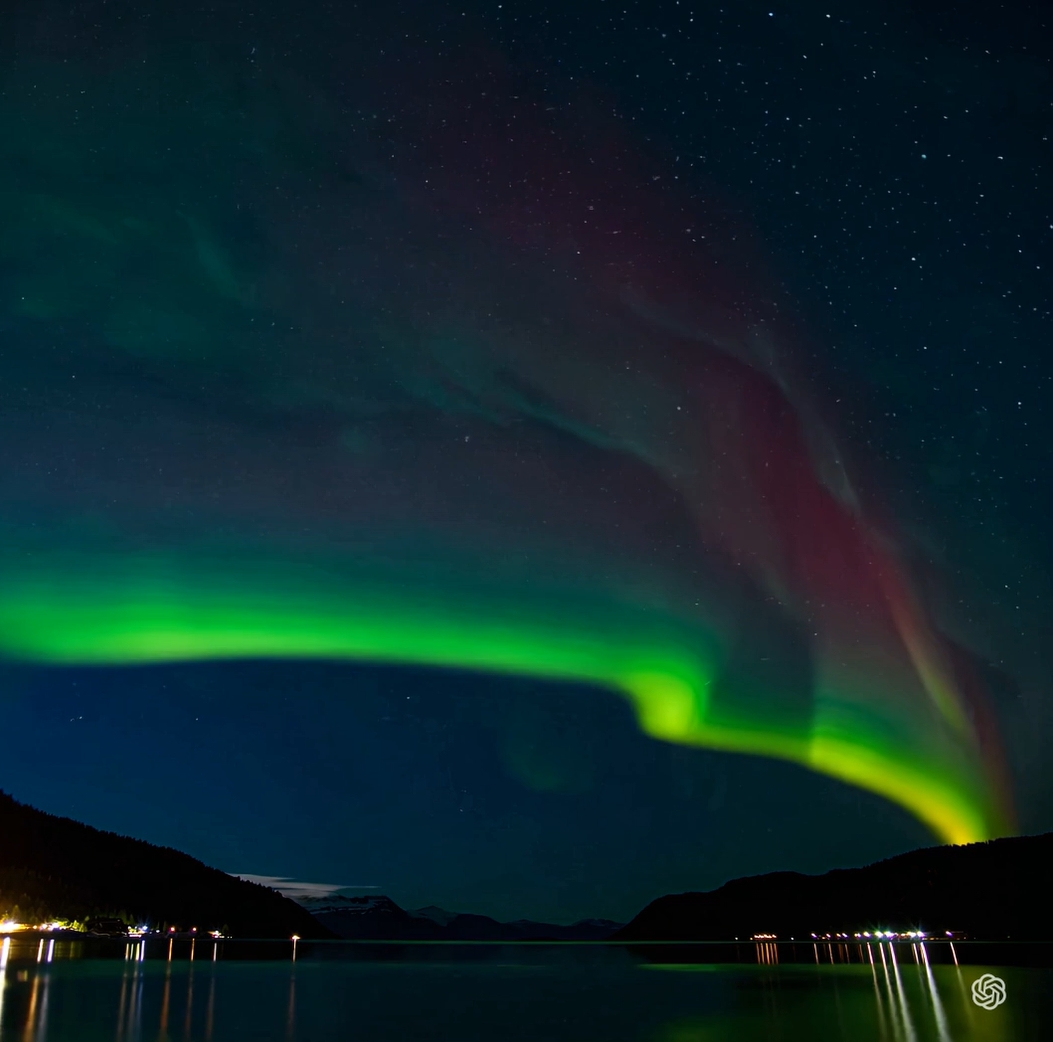}
    \captionof{subfigure}{Model B's answer}
  \end{minipage}

  \begin{minipage}[t]{0.98\textwidth}
    \medskip
    \begin{tabular}{@{}ll@{}}
      \textit{Gemini-2.0-Flash Choice}:   & Model B is better.\\
    \end{tabular}

    \medskip
    \textbf{Checklist Rationale}:]
Model A's video is overly saturated, making the aurora appear unnatural. The color of the aurora is a mix of pink, yellow, and red, which is far from the requested red and yellow. The mountain silhouette is barely visible in the background. There's also a yellow light source in the horizon that doesn't look like it belongs in an aurora video. Model B's video successfully changes the color of the aurora to a mix of red, pink, and green. While it's not exactly red and yellow, it's much closer to the request than Model A. The colors blend smoothly, and there's no flickering or distortion. The mountain silhouette is also clearly visible. There is a watermark at the bottom right corner.  Checklist item A: Model B's aurora is closer to red and yellow than Model A's. Checklist item B: The color transition in Model B's video is smooth and natural. Model A's color transition is too abrupt and unnatural. Checklist item 3: Both videos are free of flickering or distortion.  Overall, Model B's video is clearer and better fulfills the prompt, despite the watermark.

  \end{minipage}
  \end{tcolorbox}
  \caption{Video Edit - \pair ~- Clarity.}
  \label{Fig: v2v-case-p}
\end{figure*}

\begin{figure*}[htbp]
  \centering
  \begin{tcolorbox}[
    enhanced,
    attach boxed title to top center={yshift=-3mm,yshifttext=-1mm},
    boxrule=0.9pt,
    colback=gray!00,
    colframe=black!50,
    colbacktitle=Periwinkle,
    title=Case 33: Video Edit - \score ~- Clarity,
    boxed title style={size=small,colframe=Periwinkle},
    top=3mm,
    bottom=3mm,
    left=2mm,
    right=2mm,
    before upper={\parindent0em\linespread{0.85}\selectfont} 
  ]
  \vspace{-1ex}
  \begin{minipage}[t]{0.96\textwidth}
    \textbf{\textit{Checklists for \pair of Clarity:}}
    \begin{itemize}[leftmargin=*,noitemsep,topsep=2pt]
      \item Is the aurora in the edited video predominantly red and yellow?
      \item Is the transition between colors in the aurora smooth and natural-looking?
      \item Is there any flickering or distortion that makes the aurora difficult to perceive clearly?
    \end{itemize}
    
    \smallskip
    \textbf{\textit{Question}}: Modify the video to show red and yellow aurora paints the night sky over mountain silhouettes. 
  \end{minipage}
  \vspace{0.5ex}

\vspace{1ex}
    \begin{minipage}[t]{0.45\textwidth}
    \centering
    \includegraphics[width=0.60\linewidth,trim=5 5 5 5,clip]{Figures/Case/v2v_0.png}
    \captionof{subfigure}{Input Video}
  \end{minipage}
\vspace{1ex}
\hfill
  \begin{minipage}[t]{0.45\textwidth}
    \centering
    \includegraphics[width=0.60\linewidth,trim=5 5 5 5,clip]{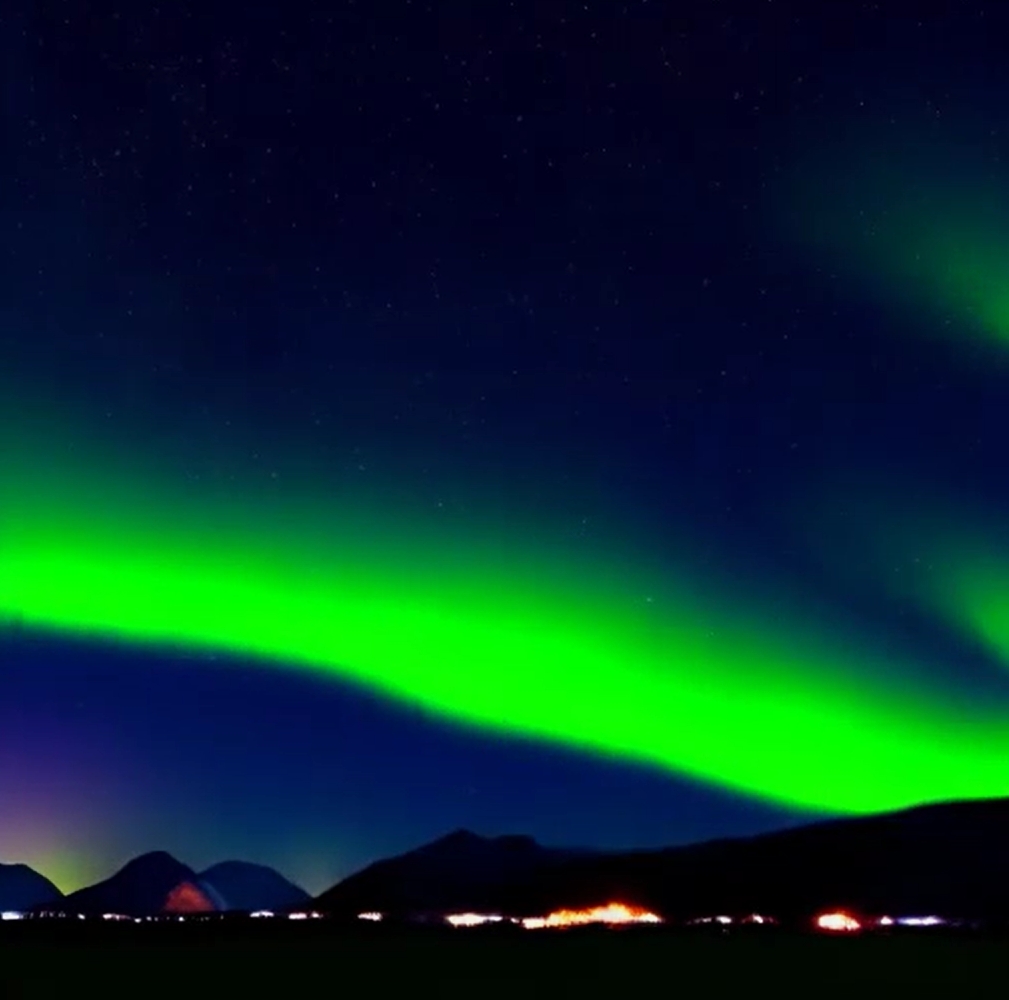}
    \captionof{subfigure}{Model's answer}
  \end{minipage}

  \vspace{1ex}
  \begin{minipage}[t]{0.98\textwidth}
    \medskip
    \begin{tabular}{@{}ll@{}}
      \textit{Gemini-2.0-Flash Score}:   & 1 \\
    \end{tabular}

    \medskip
    \textbf{Checklist Rationale}: 
Checklist item 1: The aurora is green in the edited video, not the requested red and yellow. Checklist item 2: While the color transition is smooth, it is between shades of green, not red and yellow as requested. Checklist item 3: There is no flickering or distortion in the edited video.
  \end{minipage}
  \end{tcolorbox}
  \caption{Video Edit - \score ~- Clarity.}
  \label{Fig: v2v-case-s}
\end{figure*}

See Figures \ref{Fig:ChecklistImprove} and \ref{Fig:ChecklistHallu} for \checklist influence on evaluation.
See Figures \ref{Fig: GPT Failure}, \ref{Fig: Case4} and \ref{Fig: Case Study 5} for detailed case studies.
Fig: v2t-case-pSee Figures \ref{Fig: t2t-case-p} and \ref{Fig: t2t-case-s} for text-to-text examples, Figures \ref{Fig: i2t-case-p} and \ref{Fig: i2t-case-s} for image-to-text examples, Figures \ref{Fig: a2t-case-p} and \ref{Fig: a2t-case-s} for audio-to-text examples, Figures \ref{Fig: v2t-case-p} and \ref{Fig: v2t-case-s} for video-to-text examples, Figures \ref{Fig: av2t-case-p} and \ref{Fig: av2t-case-s} for audio-video-to-text examples, Figures \ref{Fig: t2v-case-p} and \ref{Fig: t2v-case-s} for text-to-video examples, Figures \ref{Fig: t2a-case-p} and \ref{Fig: t2a-case-s} for text-to-audio examples, Figures \ref{Fig: i2i-case-p} and \ref{Fig: i2i-case-s} for image edit examples, Figures \ref{Fig: i2a-case-p} and \ref{Fig: i2a-case-s} for image-to-audio examples, Figures \ref{Fig: i2v-case-p} and \ref{Fig: i2v-case-s} for image-to-video examples, Figures \ref{Fig: a2a-case-p} adn \ref{Fig: a2a-case-s} for audio edit examples, Figures \ref{Fig: a2v-case-p} and \ref{Fig: a2v-case-s} for audio-to-video examples, Figures \ref{Fig: v2a-case-p} and \ref{Fig: v2a-case-s} for video-to-audio examples, Figures \ref{Fig: v2v-case-p} and \ref{Fig: v2v-case-s} for video edit examples.


\vspace{2\baselineskip} 

\begin{figure*}[htbp]
  \centering
  \begin{tcolorbox}[
    enhanced,
    attach boxed title to top center={yshift=-3mm,yshifttext=-1mm},
    boxrule=0.9pt,
    colback=gray!00,
    colframe=black!50,
    colbacktitle=Periwinkle,
    title=Case 6: Text to Text - \pair ~- Clarity,
    boxed title style={size=small,colframe=Periwinkle},
    top=3mm,
    bottom=3mm,
    left=2mm,
    right=2mm,
    before upper={\parindent0em\linespread{0.85}\selectfont} 
  ]
  \vspace{-1ex}
  \begin{minipage}[t]{0.96\textwidth}
    \textbf{\textit{Checklists for \pair of Clarity:}}
    \begin{itemize}[leftmargin=*,noitemsep,topsep=2pt]
      \item The response clearly identifies the core problem or question being addressed in the prompt.
      \item The response logically organizes its ideas, with a clear flow of thought that is easy to follow.
      \item If proposing potential solutions, they are stated in a manner that is easy to understand and evaluate.
    \end{itemize}
    
    \smallskip
    \textbf{\textit{Question}}:  I ran the following code for a binary classification task w/ an SVM in both R (first sample) and Python (second example). Given randomly generated data (X) and response (Y), this code performs leave group out cross validation 1000 times. Each entry of Y is therefore the mean of the prediction across CV iterations. Computing area under the curve should give ~0.5, since X and Y are completely random. However, this is not what we see. Area under the curve is frequently significantly higher than 0.5. The number of rows of X is very small, which can obviously cause problems. Any idea what could be happening here? I know that I can either increase the number of rows of X or decrease the number of columns to mediate the problem, but I am looking for other issues. $<$code snippet$>$.
  \end{minipage}
  \vspace{0.5ex}

\begin{minipage}[t]{0.96\textwidth}
    \textbf{Model A answer}: Thanks for your help!  Answer: The problem you are experiencing is likely due to overfitting. With only 28 samples and 100 features, your model is likely fitting the noise in the data rather than the underlying patterns. This is especially true for SVM models, which can easily become overfitting when the number of features is high.  Additionally, the fact that you are using leave-group-out cross-validation with only 28 samples means that your test sets are very small, which can also contribute to overfitting.  Here are a few things you can try to mitigate this issue:  1. **Regularization**: You can try adding regularization to your SVM model by setting the `C` parameter to a lower value. This will penalize the model for misclassifying samples and can help prevent overfitting.  2. **Dimensionality reduction**: You can try reducing the number of features in your data using techniques such as PCA or feature selection. This can help prevent overfitting by reducing the number of opportunities for the model to fit the noise in the data.  3. **Increase the sample size**: If possible, try increasing the sample size. This will give the model more data to work with and can help prevent overfitting.  4. **Use a different model**: You can try using a different model that is less prone to overfitting, such as a random forest or a logistic regression model.  5. **Use a different performance metric**: You can try using a different performance metric, such as the accuracy or the F1 score, which may be less sensitive to overfitting.  Here's how you can modify the R code to use regularization:  $<$code snippet$>$. \\
    \smallskip
    \textbf{Model B answer}: A:  The issue you're encountering is related to the fact that you are using a binary classification model and averaging the predictions across multiple cross\-validation iterations. When the number of samples is very small, the model can easily overfit to the training data, leading to poor generalization and potentially biased results. Here are a few key points to consider:  1. **Overfitting**: With a very small number of samples (28 in this case), the model can easily memorize the training data, leading to poor performance on the test set. This can result in predictions that are not reliable.  2. **Averaging Predictions**: When you average the predictions across multiple cross\-validation iterations, you are essentially creating a new set of probabilities that may not be well\-calibrated. This can lead to an overestimation of the model's performance.  3. **Class Imbalance**: Ensure that the class distribution in your training and test sets is balanced. If one class is overrepresented in the training set, the model might be biased towards that class.  4. **Model Complexity**: The SVM with a linear kernel might be too simple for the data, or the default hyperparameters might not be suitable. You might want to try different kernels or tune the hyperparameters.  5. **Evaluation Metric**: AUC\-ROC is a good metric for binary classification, but it can be sensitive to the distribution of the predictions. With very small sample sizes, the AUC\-ROC might not be a reliable indicator of the model's performance.  To address these issues, you can try the following:  1. **Increase Sample Size**: If possible, increase the number of samples in your dataset. This will help reduce overfitting and provide more reliable predictions.  2. **Regularization**: Use regularization techniques to prevent overfitting. For SVM, you can adjust the `C` parameter to control the trade-off between maximizing the margin and minimizing the classification error.  3. **Cross-Validation Strategy**: Use a more robust cross-validation strategy, such as stratified k-fold cross-validation, to ensure that the class distribution is balanced in each fold.  4. **Model Tuning**: Perform hyperparameter tuning to find the best settings for your model. You can use grid search or randomized search to explore different parameter combinations.  5. **Probability Estimates**: If you are using the decision function or raw predictions, consider using probability estimates instead. For SVM, you can set `probability=True` to get probability estimates.  Here is an updated version of your Python code with some of these suggestions:  $<$code snippet$>$. I have tried using the trigonometric identities to simplify the integrand,
\end{minipage}

  \vspace{1ex}
  \begin{minipage}[t]{0.96\textwidth}
    \medskip
    \begin{tabular}{@{}ll@{}}
      \textit{Learnlm-1.5-pro Choice}:   & \textbf{Model B is better.} \\
    \end{tabular}

    \medskip
    \textbf{Checklist Rationale}:
Model A gives a more thorough analysis of the potential problems and solutions, and the modified code is more reasonable. Model B's modified code increases sample numbers and reduces features arbitrarily, which is not reasonable. Model A's response is clearer.
  \end{minipage}
  \end{tcolorbox}
  \caption{Text to Text - \pair ~- Clarity.}
  \label{Fig: t2t-case-p}
\end{figure*}
\begin{figure*}[htbp]
  \centering
  \begin{tcolorbox}[
    enhanced,
    attach boxed title to top center={yshift=-3mm,yshifttext=-1mm},
    boxrule=0.9pt,
    colback=gray!00,
    colframe=black!50,
    colbacktitle=Periwinkle,
    title=Case 7: Text to Text - \score ~- Coherence,
    boxed title style={size=small,colframe=Periwinkle},
    top=3mm,
    bottom=3mm,
    left=2mm,
    right=2mm,
    before upper={\parindent0em\linespread{0.85}\selectfont} 
  ]
  \vspace{-1ex}
  \begin{minipage}[t]{0.96\textwidth}
    \textbf{\textit{Checklists for \pair of Coherence:}}
    \begin{itemize}[leftmargin=*,noitemsep,topsep=2pt]
      \item The response clearly identifies the core problem or question posed in the input text.
      \item The response maintains a consistent focus on the identified problem throughout its entirety.
      \item The response ensures that the provided code or examples are properly integrated within the analysis or explanation.
    \end{itemize}
    
    \smallskip
    \textbf{\textit{Question}}:  I ran the following code for a binary classification task w/ an SVM in both R (first sample) and Python (second example). Given randomly generated data (X) and response (Y), this code performs leave group out cross validation 1000 times. Each entry of Y is therefore the mean of the prediction across CV iterations. Computing area under the curve should give \~0.5, since X and Y are completely random. However, this is not what we see. Area under the curve is frequently significantly higher than 0.5. The number of rows of X is very small, which can obviously cause problems. Any idea what could be happening here? I know that I can either increase the number of rows of X or decrease the number of columns to mediate the problem, but I am looking for other issues. $<$code snippet$>$.
  \end{minipage}
  \vspace{0.5ex}

\begin{minipage}[t]{0.96\textwidth}
    \textbf{Model A answer}: The issue you're experiencing is likely due to the high-dimensionality and small sample size scenario, often referred to as the \"curse of dimensionality.\" When the number of features (columns) is large relative to the number of samples (rows), models like SVM can overfit very easily, leading to results that don't reflect the true underlying distribution of the data.  \#\#\# Potential Issues and Solutions:  1. **Overfitting**:     - With 100 features and only 28 samples, your model has too much flexibility and can fit noise in the data rather than capturing any true pattern. This can lead to artificially high AUC values.  2. **Resampling Bias**:     - Random sampling can introduce bias, especially when the dataset is small. Some samples may end up being more frequently included in the training set, leading to overfitting on those samples.  3. **Class Imbalance**:    - Even though your classes are balanced (14 of each), the random nature of small samples could result in imbalanced train/test splits, affecting performance metrics. Suggestions:  1. **Reduce Dimensionality**:    - Perform feature selection or dimensionality reduction (e.g., PCA) to reduce the number of features. This can help prevent overfitting.  2. **Increase Sample Size**:    - If possible, increase the number of samples relative to the number of features. Here's a brief modification of the Python code to illustrate some of these suggestions: $<$code snippet$>$. By reducing dimensionality and ensuring stratified sampling, you should see a more realistic AUC closer to 0.5.
\end{minipage}

  \vspace{1ex}
  \begin{minipage}[t]{0.96\textwidth}
    \medskip
    \begin{tabular}{@{}ll@{}}
      \textit{Learnlm-1.5-pro Score}:   & \textbf{5} \\
    \end{tabular}

    \medskip
    \textbf{Checklist Rationale}:
The response is highly coherent, demonstrating a clear and logical flow in addressing the user's query regarding unexpected AUC values in a binary classification task. It correctly identifies the core issue of overfitting due to the curse of dimensionality, given the small sample size and high feature count. The explanation is well-structured, starting with a clear problem statement, followed by a detailed breakdown of potential issues (overfitting, resampling bias, class imbalance, random initialization) and their impact on the results. Each issue is further elaborated with specific solutions and suggestions for improvement. The response maintains a consistent focus on the overfitting problem and its solutions, ensuring all parts contribute to the overall narrative. The logical progression of ideas, from problem identification to solution proposal, makes the response easy to follow and understand. Because the response is well-organized, comprehensive, and directly addresses the user's concerns with clear and actionable solutions, it receives a near perfect score.
  \end{minipage}

  \end{tcolorbox}
  \caption{Text to Text - \score ~- Coherence.}
  \label{Fig: t2t-case-s}
\end{figure*}


\begin{figure*}[htbp]
  \centering
  \begin{tcolorbox}[
    enhanced,
    attach boxed title to top center={yshift=-3mm,yshifttext=-1mm},
    boxrule=0.9pt,
    colback=gray!00,
    colframe=black!50,
    colbacktitle=Periwinkle,
    title=Case 9: Image to Text - \score ~- Completeness,
    boxed title style={size=small,colframe=Periwinkle},
    top=3mm,
    bottom=3mm,
    left=2mm,
    right=2mm,
    before upper={\parindent0em\linespread{0.85}\selectfont} 
  ]
  \vspace{-1ex}
  \begin{minipage}[t]{0.96\textwidth}
    \textbf{\textit{Checklists for \score of Completeness:}}
    \begin{itemize}[leftmargin=*,noitemsep,topsep=2pt]
      \item Ensure the narrative includes all elements suggested by the images, such as road conditions and vehicle movements.
      \item Verify that any mentioned actions or directions (e.g., turning or road curves) are referenced and consistent across the narrative.
    \end{itemize}
    
    \smallskip
    \textbf{\textit{Question}}: Based on the narratives associated with the initial images, use the final picture to bring the story to a close. Context: \{image\#1\} Caption\#1:snow track cures to the left.\{image\#2\} Caption\#2: a mobile makes a left turn on a snowy road.. \{image\#3\} Caption\#3:.
  \end{minipage}
  \vspace{0.5ex}

\vspace{1ex}
  \begin{minipage}[t]{0.30\textwidth}
    \centering
    \includegraphics[width=0.95\linewidth,trim=5 5 5 5,clip]{Figures/Case/i2t_1.jpg}
    \captionof{subfigure}{image \#1}
  \end{minipage}
  \hfill
  \begin{minipage}[t]{0.30\textwidth}
    \centering
    \includegraphics[width=0.95\linewidth,trim=5 5 5 5,clip]{Figures/Case/i2t_2.jpg}
    \captionof{subfigure}{image \#2}
  \end{minipage}
  \hfill
  \begin{minipage}[t]{0.30\textwidth}
    \centering
    \includegraphics[width=0.95\linewidth,trim=5 5 5 5,clip]{Figures/Case/i2t_3.jpg}
    \captionof{subfigure}{image \#3}
  \end{minipage}

\begin{minipage}[t]{0.96\textwidth}
    \textbf{Model's answer}: \{image\#3\}Caption\#3: The vehicle continues on the snowy road, navigating the challenging terrain with skill and determination.
\end{minipage}

  \vspace{1ex}
  \begin{minipage}[t]{0.96\textwidth}
    \medskip
    \begin{tabular}{@{}ll@{}}
      \textit{Gemini-2.0-Flash Score}:   & 3 \\
    \end{tabular}

    \medskip
    \textbf{Checklist Rationale}:
The response is moderately complete. Checklist item 1 is met: The caption mentions the snowy road and the vehicle's movement, which aligns with the image. Checklist item 2 is partially met: The caption doesn't mention the curves from the first image or the left turn from the second image. While the caption continues the story, it doesn't explicitly tie into the specifics of the previous turns. The narrative is consistent with the third image but omits some details from the previous ones.
  \end{minipage}
  \end{tcolorbox}
  \caption{Image to Text - \score ~- Completeness.}
  \label{Fig: i2t-case-s}
\end{figure*}

\begin{figure*}[htbp]
  \centering
  \begin{tcolorbox}[
    enhanced,
    attach boxed title to top center={yshift=-3mm,yshifttext=-1mm},
    boxrule=0.9pt,
    colback=gray!00,
    colframe=black!50,
    colbacktitle=Periwinkle,
    title=Case 11: Audio to Text - \score ~- Relevance,
    boxed title style={size=small,colframe=Periwinkle},
    top=3mm,
    bottom=3mm,
    left=2mm,
    right=2mm,
    before upper={\parindent0em\linespread{0.85}\selectfont} 
  ]
  \vspace{-1ex}
  \begin{minipage}[t]{0.96\textwidth}
    \textbf{\textit{Checklists for \pair of Relevance:}}
    \begin{itemize}[leftmargin=*,noitemsep,topsep=2pt]
      \item Does the response address the prompt directly without unnecessary explanations of the audio?
      \item Does the response remain focused on cultural associations throughout, avoiding digressions into other aspects of the audio?
    \end{itemize}
    
    \smallskip
    \textbf{\textit{Question}}: What cultural associations do you have with the sound of frying food?
  \end{minipage}
  \vspace{0.5ex}

\vspace{1ex}
  \begin{minipage}[t]{0.98\textwidth}
    \centering
    \includegraphics[width=0.30\linewidth,trim=5 5 5 5,clip]{Figures/Case/AudioIcon.png}
    \captionof{subfigure}{Input Audio}
  \end{minipage}

\begin{minipage}[t]{0.96\textwidth}
    \textbf{Model's answer}: The sound of frying food evokes a variety of comforting and stimulating cultural associations, depending on the specific sounds and context.  Here are some examples: * **Home Cooking and Family:** The gentle sizzle of food in a pan often connects to feelings of home, warmth, and family meals.  It suggests the care and effort someone is putting into preparing a meal for loved ones. This is especially true for the sounds of familiar dishes being prepared, evoking memories of childhood or specific family gatherings. * **Comfort and Indulgence:**  The crackling and popping of frying, particularly for foods like bacon or french fries, can be associated with indulgence and satisfying cravings. These sounds can trigger anticipation and excitement for a tasty, perhaps not-so-healthy, meal. * **Street Food and Festivities:**  The loud sizzling and spitting of food in a large wok or fryer often brings to mind street food vendors and bustling marketplaces. This links the sound to vibrant social gatherings, festivals, and a sense of community. * **Cultural Cuisine:**  Different types of frying sounds can be associated with specific cuisines.  For example, the light, delicate sizzle of tempura being fried connects to Japanese cuisine, while the intense crackling of frying chicharrones might evoke thoughts of Latin American food. * **Skill and Mastery:** The controlled sizzling and the chef's manipulation of the food in the pan can convey a sense of skill and mastery in cooking. * **Freshness and Quick Preparation:** The relatively short cooking time associated with frying can suggest freshness and a quick, readily available meal. In the provided audio clip, the calm, explanatory tone of voice along with the gentle sizzling sound leans more towards the associations of **home cooking, care, and educational cooking demonstrations.**  It's not the loud, aggressive sizzle of street food, but rather a more intimate and controlled sound.
\end{minipage}

  \vspace{1ex}
  \begin{minipage}[t]{0.96\textwidth}
    \medskip
    \begin{tabular}{@{}ll@{}}
      \textit{Gemini-2.0-Flash Score}:   & 5 \\
    \end{tabular}

    \medskip
    \textbf{Checklist Rationale}:
The model's response demonstrates a high degree of relevance to the prompt. It directly addresses the question about cultural associations related to the sound of frying food by providing a comprehensive list of potential connections, including home cooking, indulgence, street food, cultural cuisine, skill, and freshness. It also analyzes the specific audio clip provided, correctly identifying its quieter, more controlled nature and associating it with home cooking and educational demonstrations. The response stays focused on the prompt and avoids unnecessary explanations of the audio, consistently relating the sounds to cultural contexts. Therefore, it effectively fulfills the requirements of the prompt and demonstrates strong relevance.
  \end{minipage}
  \end{tcolorbox}
  \caption{Audio to Text - \score ~- Relevance.}
  \label{Fig: a2t-case-s}
\end{figure*}

\begin{figure*}[htbp]
  \centering
  \begin{tcolorbox}[
    enhanced,
    attach boxed title to top center={yshift=-3mm,yshifttext=-1mm},
    boxrule=0.9pt,
    colback=gray!00,
    colframe=black!50,
    colbacktitle=Periwinkle,
    title=Case 13: Video to Text - \score ~- Trustworthiness,
    boxed title style={size=small,colframe=Periwinkle},
    top=3mm,
    bottom=3mm,
    left=2mm,
    right=2mm,
    before upper={\parindent0em\linespread{0.85}\selectfont} 
  ]
  \vspace{-1ex}
  \begin{minipage}[t]{0.96\textwidth}
    \textbf{\textit{Checklists for \pair of Trustworthiness:}}
    \begin{itemize}[leftmargin=*,noitemsep,topsep=2pt]
      \item Does the response correctly identify the setting as a bathroom?
      \item Does the response avoid fabricating details or actions not present in the video?
      \item Does the response refrain from including information about the man's emotions or intentions, which are not explicitly shown?
      \item Does the response avoid ascribing a specific mood or tone to the video (e.g., 'tense', 'dramatic'), as these are interpretations, not objective facts?
    \end{itemize}
    
    \smallskip
    \textbf{\textit{Question}}: What does the video captures?
  \end{minipage}
  \vspace{0.5ex}

\vspace{1ex}
    \begin{minipage}[t]{0.98\textwidth}
    \centering
    \includegraphics[width=0.50\linewidth,trim=5 5 5 5,clip]{Figures/Case/v2t.png}
    \captionof{subfigure}{Frame of Input Video}
  \end{minipage}

\begin{minipage}[t]{0.96\textwidth}
    \textbf{Model's answer}: The video captures a person in a bathroom looking into a mirror. The individual is seen turning and moving around, possibly engaged in a routine activity like washing up or checking their appearance. The setting includes typical bathroom elements like a light fixture above the mirror.
\end{minipage}

  \vspace{1ex}
  \begin{minipage}[t]{0.96\textwidth}
    \medskip
    \begin{tabular}{@{}ll@{}}
      \textit{Gemini-2.0-Flash Score}:   & 5 \\
    \end{tabular}

    \medskip
    \textbf{Checklist Rationale}:
Checklist Item 1: Yes, the response correctly identifies the setting as a bathroom. Checklist Item 2: Yes, the response avoids fabricating details or actions. It sticks to observable actions like turning and moving, and mentions plausible routine activities. Checklist Item 3: Yes, the response refrains from speculating about the man's emotions or intentions. It uses neutral language and focuses on observable actions. Checklist Item 4: Yes, the response avoids ascribing a specific mood or tone. It maintains objectivity by focusing on the physical actions and setting.
  \end{minipage}
  \end{tcolorbox}
  \caption{Video to Text - \score ~- Trustworthiness.}
  \label{Fig: v2t-case-s}
\end{figure*}

\begin{figure*}[htbp]
  \centering
  \begin{tcolorbox}[
    enhanced,
    attach boxed title to top center={yshift=-3mm,yshifttext=-1mm},
    boxrule=0.9pt,
    colback=gray!00,
    colframe=black!50,
    colbacktitle=Periwinkle,
    title=Case 15: Audio Video to Text - \score ~- Trustworthiness,
    boxed title style={size=small,colframe=Periwinkle},
    top=3mm,
    bottom=3mm,
    left=2mm,
    right=2mm,
    before upper={\parindent0em\linespread{0.85}\selectfont} 
  ]
  \vspace{-1ex}
  \begin{minipage}[t]{0.96\textwidth}
    \textbf{\textit{Checklists for \pair of Trustworthiness:}}
    \begin{itemize}[leftmargin=*,noitemsep,topsep=2pt]
      \item Does the response avoid making unsubstantiated claims about the emotional impact of the sound or visuals?
      \item If the response mentions any specific cultural or symbolic interpretations of bamboo or wind chimes, are these interpretations accurate and appropriately sourced?
      \item Is the information presented consistent with generally accepted knowledge about the natural world and human perception of sound and visuals?
      \item Does the response avoid any potentially harmful or misleading information about bamboo, wind chimes, or related topics?
    \end{itemize}
    
    \smallskip
    \textbf{\textit{Question}}: How does the sound enhance the visual experience of the plant, and what emotions might it evoke in the viewer?
  \end{minipage}
  \vspace{0.5ex}

\vspace{1ex}
  \begin{minipage}[t]{0.48\textwidth}
    \centering
    \includegraphics[width=0.80\linewidth,trim=5 5 5 5,clip]{Figures/Case/AudioIcon.png}
    \captionof{subfigure}{Input Audio}
  \end{minipage}
  \hfill
    \begin{minipage}[t]{0.48\textwidth}
    \centering
    \includegraphics[width=0.80\linewidth,trim=5 5 5 5,clip]{Figures/Case/av2t.png}
    \captionof{subfigure}{Frame of Input Video}
  \end{minipage}

\begin{minipage}[t]{0.96\textwidth}
    \textbf{Model's answer}: The sound of the wind chimes in the video adds a sense of tranquility and peacefulness to the visual experience of the plant swaying in the wind. It creates a soothing ambiance that can evoke a sense of calmness and relaxation in the viewer, allowing them to appreciate the beauty of nature and the movement of the plant. Additionally, the sound can also evoke a sense of nostalgia or reminiscence of childhood memories, as wind chimes are often associated with warm summer days and lazy afternoons spent outdoors. Overall, the combination of the visual and auditory elements in the video creates a serene and calming experience for the viewer.
\end{minipage}

  \vspace{1ex}
  \begin{minipage}[t]{0.96\textwidth}
    \medskip
    \begin{tabular}{@{}ll@{}}
      \textit{Gemini-2.0-Flash Score}:   & 4 \\
    \end{tabular}

    \medskip
    \textbf{Checklist Rationale}:
The model's response is mostly accurate and avoids unsubstantiated or harmful claims, making it reasonably trustworthy. It correctly identifies the general emotional impact of wind chimes and their association with nature and relaxation. However, the connection to childhood memories might be subjective and not universally applicable, but it doesn't detract significantly from the overall trustworthiness. The response also doesn't delve into cultural interpretations or make any claims about bamboo, focusing solely on the general sensory experience, which further contributes to its trustworthiness. It aligns with common understanding of how sound and visuals interact and doesn't present any misleading information. Checklist evaluation: 1. The response avoids unsubstantiated claims, grounding its descriptions in common associations with wind chimes and nature sounds. 2. No specific cultural interpretations are mentioned, thus avoiding potential inaccuracies. 3. The information presented is consistent with general knowledge about sound and visual perception. 4. The response avoids potentially harmful or misleading information.
  \end{minipage}
  \end{tcolorbox}
  \caption{Audio Video to Text - \score ~- Trustworthiness.}
  \label{Fig: av2t-case-s}
\end{figure*}


\begin{figure*}[htbp]
  \centering
  \begin{tcolorbox}[
    enhanced,
    attach boxed title to top center={yshift=-3mm,yshifttext=-1mm},
    boxrule=0.9pt,
    colback=gray!00,
    colframe=black!50,
    colbacktitle=Periwinkle,
    title=Case 17: Text to Video - \score ~- Trustworthiness,
    boxed title style={size=small,colframe=Periwinkle},
    top=3mm,
    bottom=3mm,
    left=2mm,
    right=2mm,
    before upper={\parindent0em\linespread{0.85}\selectfont} 
  ]
  \vspace{-1ex}
  \begin{minipage}[t]{0.96\textwidth}
    \textbf{\textit{Checklists for \pair of Trustworthiness:}}
    \begin{itemize}[leftmargin=*,noitemsep,topsep=2pt]
      \item The video doesn't show any hazardous or misleading situations in its depiction of the kite and the snowboard.
      \item The video avoids presenting any factual inaccuracies related to the physics of kite flying or snowboarding, if applicable.
      \item The generated video does not contain any misleading information or elements that could be misconstrued by the viewer.
    \end{itemize}
    
    \smallskip
    \textbf{\textit{Question}}: Give me a video that illustrates the idea or scene described below. A vibrant snowboard, adorned with dynamic graphics and bold colors, is securely mounted atop a sleek, high-performance kite. The scene captures the front view, showcasing the snowboard's intricate design and the kite's aerodynamic structure. The kite's fabric, a striking blend of neon hues, billows gracefully against a backdrop of a clear, azure sky. The snowboard's bindings are prominently displayed, hinting at the thrilling adventure that awaits. The entire setup, bathed in the golden glow of the sun, exudes an aura of excitement and innovation, promising an exhilarating ride through the skies.
  \end{minipage}
  \vspace{0.5ex}

  \vspace{1ex}
  \begin{minipage}[t]{0.95\textwidth}
    \centering
    \includegraphics[width=0.50\linewidth,trim=5 5 5 5,clip]{Figures/Case/T2V_1.png}
    \captionof{subfigure}{Model's answer}
  \end{minipage}

  \vspace{1ex}
  \begin{minipage}[t]{0.96\textwidth}
    \medskip
    \begin{tabular}{@{}ll@{}}
      \textit{GPT-4o Score}:   & 5 \\
    \end{tabular}

    \medskip
    \textbf{Checklist Rationale}:
The video clearly depicts a snowboard attached to a kite (Checklist item 1). The snowboard has dynamic graphics and bold colors that match the description (Checklist item 2). The background is consistent with a clear and azure sky, as described in the prompt (Checklist item 3). Overall, the video aligns perfectly with the requested scene, and all checklist items are satisfied.
  \end{minipage}
  \end{tcolorbox}
  \caption{Text to Video - \score ~- Trustworthiness.}
  \label{Fig: t2v-case-s}
\end{figure*}

\begin{figure*}[htbp]
  \centering
  \begin{tcolorbox}[
    enhanced,
    attach boxed title to top center={yshift=-3mm,yshifttext=-1mm},
    boxrule=0.9pt,
    colback=gray!00,
    colframe=black!50,
    colbacktitle=Periwinkle,
    title=Case 18: Text to Audio - \pair ~- Creativity,
    boxed title style={size=small,colframe=Periwinkle},
    top=3mm,
    bottom=3mm,
    left=2mm,
    right=2mm,
    before upper={\parindent0em\linespread{0.85}\selectfont} 
  ]
  \vspace{-1ex}
  \begin{minipage}[t]{0.96\textwidth}
    \textbf{\textit{Checklists for \pair of Creativity:}}
    \begin{itemize}[leftmargin=*,noitemsep,topsep=2pt]
      \item The output demonstrates some unexpected or novel elements in the way the sounds are mixed or presented.
      \item The sound has unique spatial qualities; the thunder and rain feel like they are coming from different areas of the listener's space, not just a stereo mix.
    \end{itemize}
    
    \smallskip
    \textbf{\textit{Question}}: Generate Deep rolling thunder sounds mixed with a steady gentle rainfall.
  \end{minipage}
  \vspace{0.5ex}

\vspace{1ex}
  \begin{minipage}[t]{0.48\textwidth}
    \centering
    \includegraphics[width=0.50\linewidth,trim=5 5 5 5,clip]{Figures/Case/AudioIcon.png}
    \captionof{subfigure}{Model A answer}
  \end{minipage}
  \hfill
  \begin{minipage}[t]{0.48\textwidth}
    \centering
    \includegraphics[width=0.50\linewidth,trim=5 5 5 5,clip]{Figures/Case/AudioIcon.png}
    \captionof{subfigure}{Model B answer}
  \end{minipage}

  \vspace{1ex}
  \begin{minipage}[t]{0.96\textwidth}
    \medskip
    \begin{tabular}{@{}ll@{}}
      \textit{Gemini-2.0-Flash Choice}:   & Tie \\
    \end{tabular}

    \medskip
    \textbf{Checklist Rationale}:
Both responses are equally reliable as they accurately portray the requested sounds without introducing misleading or unrelated elements, ensuring user trust in the audio's depiction of the prompt.
  \end{minipage}
  \end{tcolorbox}
  \caption{Text to Audio - \pair ~- Creativity.}
  \label{Fig: t2a-case-p}
\end{figure*}

\begin{figure*}[htbp]
  \centering
  \begin{tcolorbox}[
    enhanced,
    attach boxed title to top center={yshift=-3mm,yshifttext=-1mm},
    boxrule=0.9pt,
    colback=gray!00,
    colframe=black!50,
    colbacktitle=Periwinkle,
    title=Case 19: Text to Audio - \score ~- Creativity,
    boxed title style={size=small,colframe=Periwinkle},
    top=3mm,
    bottom=3mm,
    left=2mm,
    right=2mm,
    before upper={\parindent0em\linespread{0.85}\selectfont} 
  ]
  \vspace{-1ex}
  \begin{minipage}[t]{0.96\textwidth}
    \textbf{\textit{Checklists for \score of Creativity:}}
    \begin{itemize}[leftmargin=*,noitemsep,topsep=2pt]
      \item The output demonstrates some unexpected or novel elements in the way the sounds are mixed or presented.
      \item The sound has unique spatial qualities; the thunder and rain feel like they are coming from different areas of the listener's space, not just a stereo mix.
    \end{itemize}
    
    \smallskip
    \textbf{\textit{Question}}: Generate Deep rolling thunder sounds mixed with a steady gentle rainfall.
  \end{minipage}
  \vspace{0.5ex}

\vspace{1ex}
  \begin{minipage}[t]{0.95\textwidth}
    \centering
    \includegraphics[width=0.30\linewidth,trim=5 5 5 5,clip]{Figures/Case/AudioIcon.png}
    \captionof{subfigure}{Model's answer}
  \end{minipage}

  \vspace{1ex}
  \begin{minipage}[t]{0.96\textwidth}
    \medskip
    \begin{tabular}{@{}ll@{}}
      \textit{Gemini-2.0-Flash Score}:   & 3 \\
    \end{tabular}

    \medskip
    \textbf{Checklist Rationale}:
The audio has a moderate level of creativity. While the sounds are well-mixed, there isn't significant novelty in the spatial qualities or mixing techniques.
  \end{minipage}
  \end{tcolorbox}
  \caption{Text to Audio - \score ~- Creativity.}
  \label{Fig: t2a-case-s}
\end{figure*}

\begin{figure*}[htbp]
  \centering
  \begin{tcolorbox}[
    enhanced,
    attach boxed title to top center={yshift=-3mm,yshifttext=-1mm},
    boxrule=0.9pt,
    colback=gray!00,
    colframe=black!50,
    colbacktitle=Periwinkle,
    title=Case 23: Image to Audio - \score ~- Creativity,
    boxed title style={size=small,colframe=Periwinkle},
    top=3mm,
    bottom=3mm,
    left=2mm,
    right=2mm,
    before upper={\parindent0em\linespread{0.85}\selectfont} 
  ]
  \vspace{-1ex}
  \begin{minipage}[t]{0.96\textwidth}
    \textbf{\textit{Checklists for \pair of Creativity:}}
    \begin{itemize}[leftmargin=*,noitemsep,topsep=2pt]
      \item Is the atmosphere of a racetrack captured, possibly including crowd noises, tire squealing, or pit stop sounds?
      \item Does the audio include subtle, creative elements like the sounds of wind or mechanical adjustments?
    \end{itemize}
    
    \smallskip
    \textbf{\textit{Question}}: Create audio that reflects the image.
  \end{minipage}
  \vspace{0.5ex}

\vspace{1ex}
    \begin{minipage}[t]{0.45\textwidth}
    \centering
    \includegraphics[width=0.60\linewidth,trim=5 5 5 5,clip]{Figures/Case/i2a.jpg}
    \captionof{subfigure}{Input Image}
  \end{minipage}
\vspace{1ex}
\hfill
  \begin{minipage}[t]{0.45\textwidth}
    \centering
    \includegraphics[width=0.60\linewidth,trim=5 5 5 5,clip]{Figures/Case/AudioIcon.png}
    \captionof{subfigure}{Model's answer}
  \end{minipage}

  \vspace{1ex}
  \begin{minipage}[t]{0.98\textwidth}
    \medskip
    \begin{tabular}{@{}ll@{}}
      \textit{Gemini-2.0-Flash Score}:   & 5 \\
    \end{tabular}

    \medskip
    \textbf{Checklist Rationale}: The model's response successfully captures the atmosphere of a Formula 1 race. The audio includes engine noises, tire squeals, and crowd cheers, creating an immersive experience that aligns well with the image of a Formula 1 car on the track. The checklist items are well addressed, with the audio creatively incorporating subtle details such as wind sounds and mechanical adjustments. Overall, the audio presents a vivid soundscape that enhances the visual experience of the image.
  \end{minipage}
  \end{tcolorbox}
  \caption{Image to Audio - \score ~- Creativity.}
  \label{Fig: i2a-case-s}
\end{figure*}

\begin{figure*}[htbp]
  \centering
  \begin{tcolorbox}[
    enhanced,
    attach boxed title to top center={yshift=-3mm,yshifttext=-1mm},
    boxrule=0.9pt,
    colback=gray!00,
    colframe=black!50,
    colbacktitle=Periwinkle,
    title=Case 26: Audio Edit - \pair ~- Relevance,
    boxed title style={size=small,colframe=Periwinkle},
    top=3mm,
    bottom=3mm,
    left=2mm,
    right=2mm,
    before upper={\parindent0em\linespread{0.85}\selectfont} 
  ]
  \vspace{-1ex}
  \begin{minipage}[t]{0.96\textwidth}
    \textbf{\textit{Checklists for \pair of Relevance:}}
    \begin{itemize}[leftmargin=*,noitemsep,topsep=2pt]
      \item Does the generated audio prominently feature the sound of accelerating thunder as requested?
      \item Does the generated audio accurately reflect the 'accelerating' aspect of the thunder, moving from quieter, spaced-out rumbles to a more intense and frequent crescendo?
    \end{itemize}
    
    \smallskip
    \textbf{\textit{Question}}: Alter the audio to An thunder is accelerating
  \end{minipage}
  \vspace{0.5ex}

\vspace{1ex}
    \begin{minipage}[t]{0.30\textwidth}
    \centering
    \includegraphics[width=0.98\linewidth,trim=5 5 5 5,clip]{Figures/Case/AudioIcon.png}
    \captionof{subfigure}{Input Audio}
  \end{minipage}
\hfill
\vspace{1ex}
  \begin{minipage}[t]{0.30\textwidth}
    \centering
    \includegraphics[width=0.98\linewidth,trim=5 5 5 5,clip]{Figures/Case/AudioIcon.png}
    \captionof{subfigure}{Model A's answer}
  \end{minipage}
\hfill
  \begin{minipage}[t]{0.30\textwidth}
    \centering
    \includegraphics[width=0.98\linewidth,trim=5 5 5 5,clip]{Figures/Case/AudioIcon.png}
    \captionof{subfigure}{Model B's answer}
  \end{minipage}

  \begin{minipage}[t]{0.98\textwidth}
    \medskip
    \begin{tabular}{@{}ll@{}}
      \textit{Gemini-2.0-Flash Choice}:   & Model B is better.\\
    \end{tabular}

    \medskip
    \textbf{Checklist Rationale}:]
Model A's response did not seem to represent accelerating thunder. It sounded like a single clap of thunder. Model B's thunder sounded like it was getting louder and closer, which is how accelerating thunder would sound.  Model B's response is better because it produced a sound that more closely resembles the request.

  \end{minipage}
  \end{tcolorbox}
  \caption{Audio Edit - \pair ~- Relevance.}
  \label{Fig: a2a-case-p}
\end{figure*}

\begin{figure*}[htbp]
  \centering
  \begin{tcolorbox}[
    enhanced,
    attach boxed title to top center={yshift=-3mm,yshifttext=-1mm},
    boxrule=0.9pt,
    colback=gray!00,
    colframe=black!50,
    colbacktitle=Periwinkle,
    title=Case 27: Audio Edit - \score ~- Relevance,
    boxed title style={size=small,colframe=Periwinkle},
    top=3mm,
    bottom=3mm,
    left=2mm,
    right=2mm,
    before upper={\parindent0em\linespread{0.85}\selectfont} 
  ]
  \vspace{-1ex}
  \begin{minipage}[t]{0.96\textwidth}
    \textbf{\textit{Checklists for \pair of Relevance:}}
    \begin{itemize}[leftmargin=*,noitemsep,topsep=2pt]
      \item Does the generated audio prominently feature the sound of accelerating thunder as requested?
      \item Does the generated audio accurately reflect the 'accelerating' aspect of the thunder, moving from quieter, spaced-out rumbles to a more intense and frequent crescendo?
    \end{itemize}
    
    \smallskip
    \textbf{\textit{Question}}: Alter the audio to An thunder is accelerating
  \end{minipage}
  \vspace{0.5ex}

\vspace{1ex}
    \begin{minipage}[t]{0.45\textwidth}
    \centering
    \includegraphics[width=0.60\linewidth,trim=5 5 5 5,clip]{Figures/Case/AudioIcon.png}
    \captionof{subfigure}{Input Audio}
  \end{minipage}
\vspace{1ex}
\hfill
  \begin{minipage}[t]{0.45\textwidth}
    \centering
    \includegraphics[width=0.60\linewidth,trim=5 5 5 5,clip]{Figures/Case/AudioIcon.png}
    \captionof{subfigure}{Model's answer}
  \end{minipage}

  \vspace{1ex}
  \begin{minipage}[t]{0.98\textwidth}
    \medskip
    \begin{tabular}{@{}ll@{}}
      \textit{Gemini-2.0-Flash Score}:   & 1 \\
    \end{tabular}

    \medskip
    \textbf{Checklist Rationale}: 
Checklist item 1: The generated audio does not prominently feature the sound of thunder. Instead, it features a continuous, high-pitched beep. This deviates significantly from the requested thunder sound. Checklist item 2: Since the generated audio isn't thunder, it cannot reflect the accelerating nature requested. There's no transition from quieter rumbles to a more intense crescendo as described in the prompt.
  \end{minipage}
  \end{tcolorbox}
  \caption{Audio Edit - \score ~- Relevance.}
  \label{Fig: a2a-case-s}
\end{figure*}

\begin{figure*}[htbp]
  \centering
  \begin{tcolorbox}[
    enhanced,
    attach boxed title to top center={yshift=-3mm,yshifttext=-1mm},
    boxrule=0.9pt,
    colback=gray!00,
    colframe=black!50,
    colbacktitle=Periwinkle,
    title=Case 31: Video to Audio - \score ~- Creativity.,
    boxed title style={size=small,colframe=Periwinkle},
    top=3mm,
    bottom=3mm,
    left=2mm,
    right=2mm,
    before upper={\parindent0em\linespread{0.85}\selectfont} 
  ]
  \vspace{-1ex}
  \begin{minipage}[t]{0.96\textwidth}
    \textbf{\textit{Checklists for \pair of Creativity:}}
    \begin{itemize}[leftmargin=*,noitemsep,topsep=2pt]
      \item Does the audio enhance the video by adding another layer of emotion except dog's whimpering like human's 
      \item Is the audio well-integrated with the visuals, creating a cohesive and immersive experience?
    \end{itemize}
    
    \smallskip
    \textbf{\textit{Question}}: Generate corresponding audio based on the video's visuals.
  \end{minipage}
  \vspace{0.5ex}

\vspace{1ex}
    \begin{minipage}[t]{0.45\textwidth}
    \centering
    \includegraphics[width=0.60\linewidth,trim=5 5 5 5,clip]{Figures/Case/v2a_0.png}
    \captionof{subfigure}{Input Video}
  \end{minipage}
\vspace{1ex}
\hfill
  \begin{minipage}[t]{0.45\textwidth}
    \centering
    \includegraphics[width=0.60\linewidth,trim=5 5 5 5,clip]{Figures/Case/AudioIcon.png}
    \captionof{subfigure}{Model's answer}
  \end{minipage}

  \vspace{1ex}
  \begin{minipage}[t]{0.98\textwidth}
    \medskip
    \begin{tabular}{@{}ll@{}}
      \textit{Gemini-2.0-Flash Score}:   & 1 \\
    \end{tabular}

    \medskip
    \textbf{Checklist Rationale}: 
The model's audio output whimpering does not quite match the dog's action in the video. It does not enhance the video and isn't well-integrated with the visuals.
  \end{minipage}
  \end{tcolorbox}
  \caption{Video to Audio - \score ~- Creativity.}
  \label{Fig: v2a-case-s}
\end{figure*}

\end{document}